\pdfoutput=1

\documentclass[11pt]{article}

\PassOptionsToPackage{dvipsnames,table}{xcolor}

\usepackage[preprint]{coling}

\usepackage{caption}
\usepackage{capt-of}
\usepackage{times}
\usepackage{latexsym}
\usepackage[T1]{fontenc}
\usepackage[utf8]{inputenc}
\usepackage{microtype}
\usepackage{inconsolata}
\usepackage{graphicx}
\usepackage{courier}

\usepackage{amsmath}
\usepackage{amssymb}
\usepackage{dsfont}
\usepackage{booktabs}
\usepackage{multirow}
\usepackage[most]{tcolorbox}
\usepackage[dvipsnames]{xcolor}
\usepackage{enumitem}
\usepackage[table]{xcolor}

\definecolor{gain}{HTML}{0173B3}
\definecolor{drop}{HTML}{8F4A55}

\usepackage{pdfcomment}
\usepackage{arydshln}

\usepackage{listings}
\lstdefinestyle{promptbox}{basicstyle=\ttfamily\scriptsize, breaklines=true, breakatwhitespace=false, columns=fullflexible, keepspaces=true, showstringspaces=false, upquote=true, frame=none, aboveskip=2pt, belowskip=2pt}

\newcommand{\uzh}{\includegraphics[width=1.5em]{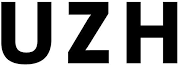}}

\newtcbox{\rulebadge}{
  on line,
  colback=white,
  colframe=BrickRed,
  boxrule=0.8pt,
  arc=2pt,
  left=2pt,
  right=2pt,
  top=1pt,
  bottom=1pt,
  boxsep=0pt
}

\newtcolorbox{promptbox}[1]{
  breakable,
  enhanced jigsaw,
  colback=gray!5, colframe=black!65,
  coltitle=white, fonttitle=\bfseries\footnotesize,
  fontupper=\footnotesize,
  title={#1},
  boxrule=1pt, arc=1.5pt,
  left=3pt, right=3pt, top=3pt, bottom=3pt,
  before skip=6pt, after skip=6pt
}

\newcommand{\yiq}[1]
{{\color{RoyalBlue}\bgroup\textbf{[yingqiang:}}~#1\textbf{]}\egroup}

\title{\textsc{AD4Edu}: Benchmarking University Lecture Audio Description across Accessibility Standards, Metrics, and Learner Preferences}

\author{
  \hspace{0.5em}
  \textbf{Stylianos Psychias}$^{\,\uzh}$ \hspace{0.2em}
  \textbf{Yingqiang Gao\textsuperscript{*}}$^{\uzh}$
  \\ \\
  \hspace{1em}
  $^{\uzh\,}$Department of Computational Linguistics, University of Zurich, Switzerland \\ \\
  \hspace{0.6em}
  \texttt{\{stylianos.psychias, yingqiang.gao\}@uzh.ch} \\
}

\begin{document}
\maketitle

\begingroup
\renewcommand{\thefootnote}{\fnsymbol{footnote}}
\footnotetext[1]{Corresponding author.}
\endgroup

\begin{abstract}

Audio description (AD) aims to enhance the accessibility of visual media for blind and low-vision (BLV) persons. However, in higher education where slide-based lectures are a default mode of teaching, lecture AD remains largely underexplored. The main reason is that existing production guidelines, evaluation metrics, and benchmark datasets are oriented toward entertainment content and do not transfer to lectures, where ADs must complement the lecturer's speech, resolve deictic references, and fit within narrow pauses. In this work, we propose 
\textsc{AD4Edu}, consisting of a AD benchmark \textsc{AD4Edu-Bench} and a rule-diagnosable evaluation metric \textsc{AD4Edu-Eval}, for generating and evaluating ADs in higher education. Specifically, we consolidated established accessibility guidance, including WCAG 2.2, the DCMP Description Key, and NCAM's STEM description practices. We further collected silver reference ADs, judge-ordered preference pairs, and pairwise preferences from four BLV raters, and post-trained models with direct preference optimization (DPO) for detailed analysis. Our findings indicate that while the metric reliably detects violations of the standard, ADs that comply with it do not necessarily fulfill the subjective preferences of BLV persons, most notably on description length, underscoring the necessity of active participation of BLV persons in co-designing inclusive AI solutions.

\begin{minipage}{\columnwidth}
\centering
\raisebox{-0.1cm}{\hspace{-3em}\includegraphics[height=0.5cm]{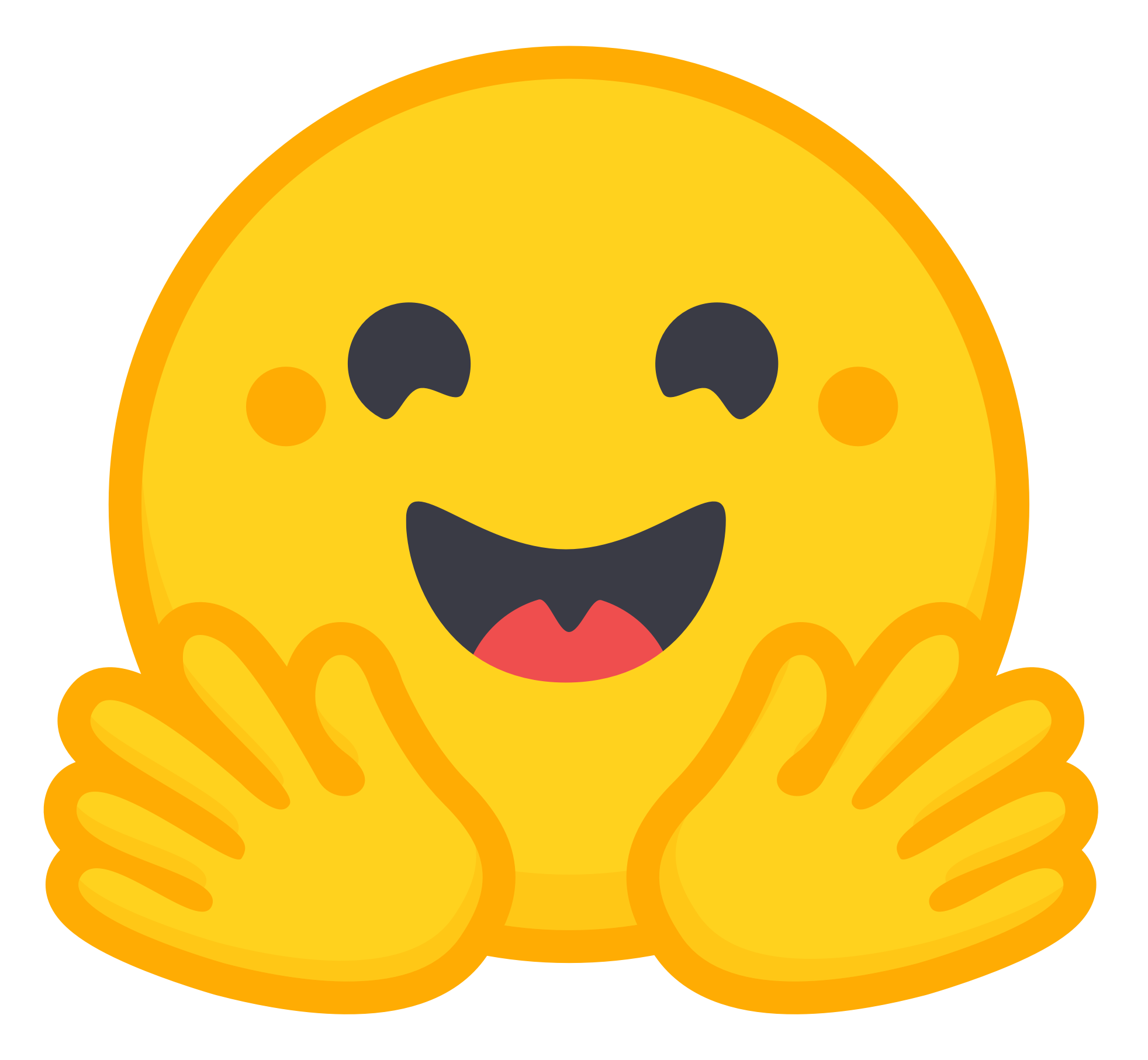}}~
\href{https://huggingface.co/datasets/Hermeneia/AD4Edu}{\textcolor{gray}{Data}} \quad
\raisebox{-0.15cm}{\includegraphics[height=0.5cm]{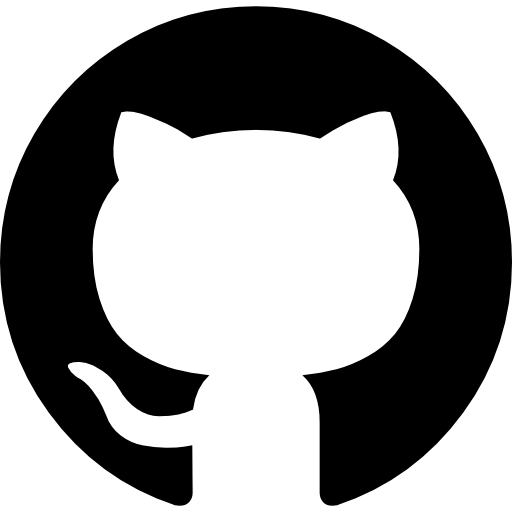}}~
\href{https://github.com/psychias/ad_for_edu}{\textcolor{gray}{Code}}
\quad
\raisebox{-0.15cm}
{\includegraphics[height=0.5cm]{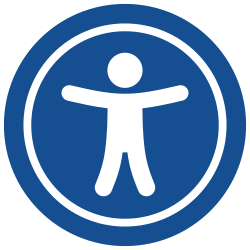}}~
\href{https://github.com/psychias/blv_annotation_tool}{\textcolor{gray}{Tool}}
\quad
\raisebox{-0.15cm}
{\includegraphics[height=0.5cm]{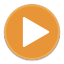}}~
\href{https://github.com/psychias/blv_mediaplayer}{\textcolor{gray}{Player}}

\end{minipage}

\end{abstract}

\section{Introduction}
\label{sec:intro}

Audio description (AD) renders visual video content as spoken narration for blind and low-vision (BLV) persons, conveying actions, scene changes, and on-screen information during pauses in the soundtrack \citep{w3c2015maur,dcmp2026descriptionkey}. At least 2.2 billion people worldwide have a vision impairment \citep{who2019vision}, meanwhile 15\% of students in European higher education report a sensory impairment (including vision) that limits their studies \citep{eurostudent2021}. Equal access to education is recognized in the UN Convention on the Rights of Persons with Disabilities \citep{uncrpd2006} and Sustainable Development Goal 4 \citep{un2015sdg}. Recorded slide-based lectures are challenging for BLV students because teaching relies on diagrams, equations, tables, and pointing gestures that are referenced but not verbalized. Effective AD therefore must complement speech, resolve deictic references, and fit within narrow pauses \citep{Jiang_2024}. Yet automatic AD research has focused primarily on entertainment content such as film \citep{han2023autoad,han2024autoad3, fischer2025swissadt, narrad2025}, leaving educational settings largely unexplored.

Recent work has improved automatic AD through contextual conditioning, guideline-conditioned prompting, on-demand description, and interactive question answering \citep{han2023autoad2,autoadzero2024,distinctad2025,videoa11y2025,cheema2025nowuserdrivenaudiodescription,anderer2025lecture}. Evaluation has likewise adopted reference overlap, image--text alignment, and LLM-based judgments \citep{popovic2015chrf,zhang2020bertscore,hessel2021clipscore,zheng2023judging,gu2024survey}. BLV requirements, however, remain weakly integrated into system development. Professional accessibility guidelines \citep{w3c2024wcag22,netflix2024adstyle,ncam2026stem} are typically used as prompt text rather than as verifiable constraints, and existing metrics reduce evaluation to aggregate scores that do not say which requirements are satisfied or violated. BLV feedback thus enters only at the final evaluation stage.

We address this gap with a new benchmark and a rule-diagnosable metric for automatic AD of slide-based university lectures. We consolidate professional and regulatory guidance into a 45-rule standard, investigate direct preference optimization (DPO; \citealp{rafailov2023dpo}) as a trial preference-learning method, and involve BLV raters in reviewing the standard and annotating preferences. 

The main contributions are: (1) a provenance-tracked, 45-rule standard for educational AD; (2) \textsc{AD4Edu-Bench}, a corpus of pedagogically meaningful lecture moments with silver reference ADs, judge-ordered preferences, and 400 AD pairs judged by four BLV raters; (3) \textsc{AD4Edu-Eval}, which operationalizes the standard as a rule-diagnosable metric in three separately reported modes; and (4) an empirical analysis of how rule compliance, automatic metrics, machine judgments, and BLV preferences relate under supervised fine-tuning (SFT) and DPO post-training.


\section{Related Work}
\label{sec:related}

Most work on automatic AD generation has focused on film \citep{han2023autoad,han2023autoad2,han2024autoad3,autoadzero2024,narrad2025,distinctad2025,chu2024llmadlargelanguagemodel}; recent studies cover guideline-conditioned and user-driven description of web video \citep{videoa11y2025,cheema2025nowuserdrivenaudiodescription,do2026scalableaudiodescriptionquality}, accessible interaction with lecture recordings \citep{anderer2025lecture,sechayk2025veasyguide,Jiang_2024}, multimodal lecture corpora \citep{lee2023lpm,chen2024m3av}, professional AD standards \citep{dcmp2026descriptionkey,ncam2026stem,netflix2024adstyle,frankel2016clearspeak}, and authoring tools for timing-constrained description \citep{pavel2020rescribe}. Closest to our work, \citet{videoa11y2025} condition a multimodal model on curated AD guidelines and evaluate it with BLV raters, while \citet{vidscribe2026} provide BLV persons with customization and conversational question answering. These approaches, however, use guidelines to steer generation rather than formalize them as verifiable conditions, so a failure cannot be attributed to a specific requirement \citep{gao2024adsurvey}.

\begin{figure*}
    \centering
    \pdftooltip{\includegraphics[width=\textwidth]{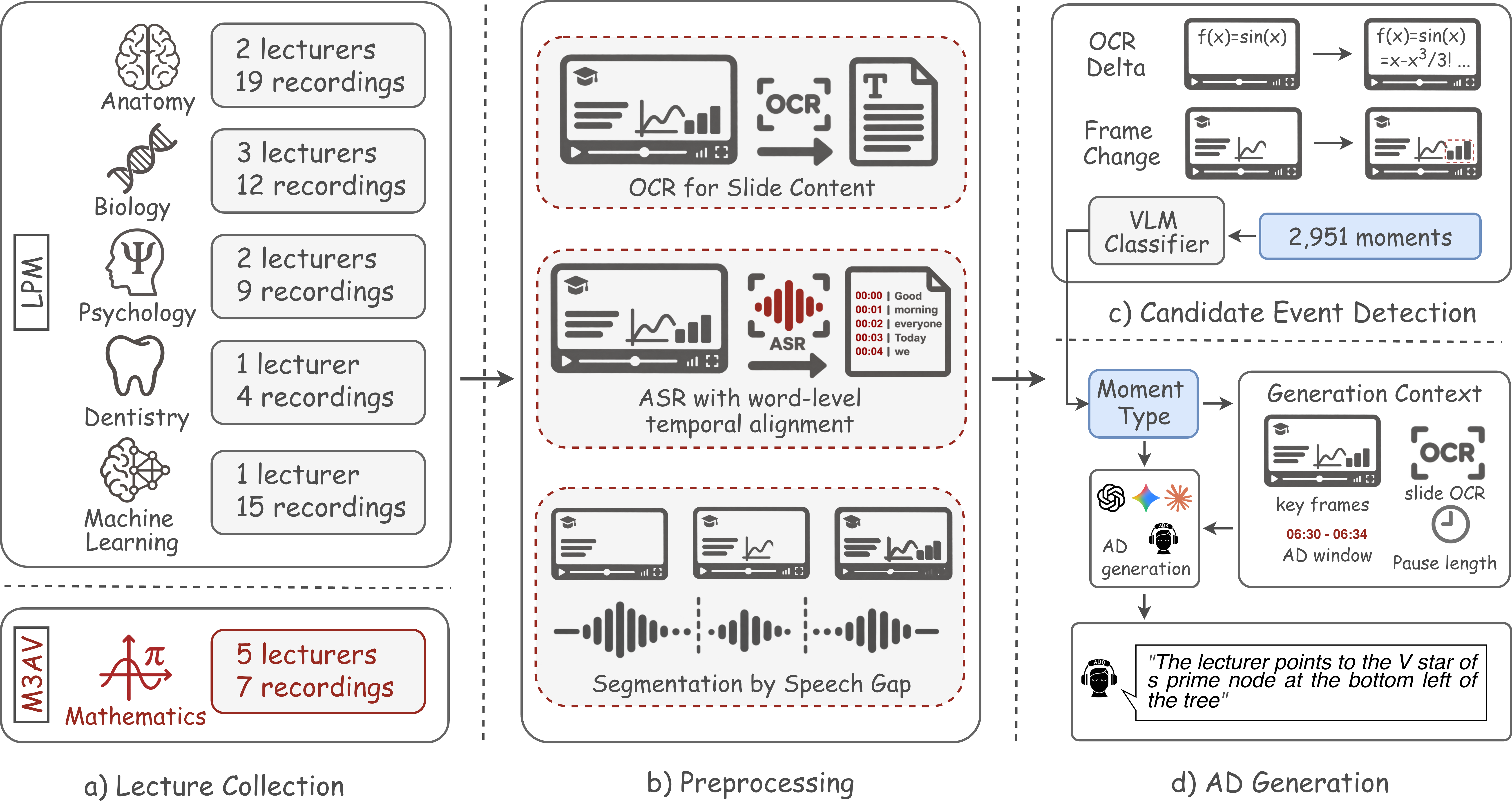}}{Four-panel pipeline diagram. Panel a: lecture recordings are collected from the LPM and M3AV datasets across six subjects. Panel b: each recording is processed with slide OCR, word-level ASR, and speech-gap segmentation. Panel c: candidate visual events from OCR deltas, frame changes, and speech gaps are classified by a VLM into pedagogical moment types or rejected. Panel d: for each retained moment, the moment type, keyframes, slide OCR, transcript window, and pause length are passed to multiple VLMs, which write an audio description or stay silent.}
    \caption{\textbf{Corpus construction and AD generation pipeline}. (a) Lecture recordings are collected from LPM and M3AV datasets, covering six academic subjects. (b) Each recording is processed with slide OCR, word-level ASR, and speech-gap segmentation to obtain temporally aligned multimodal information. (c) Candidate visual events are proposed from OCR deltas, frame changes, and speech gaps, and a VLM classifier assigns each candidate a pedagogically meaningful moment type or rejects it. (d) For each retained moment, the moment type, keyframes, slide OCR, transcript window, and pause length are provided as generation context to multiple VLMs, which either produce an audio description or decide to remain silent.}
    \label{fig:corpus-construct}
\end{figure*}

Evaluation of machine-generated ADs commonly relies on reference-based metrics such as chrF, BERTScore, CIDEr, and SPICE \citep{popovic2015chrf,post2018sacrebleu,zhang2020bertscore,vedantam2015cider,anderson2016spice}, image--text alignment and hallucination detection metrics \citep{hessel2021clipscore,hu2023tifa,li2023pope,rohrbach2018chair,kala2025wysiwa}, or LLM-as-a-Judge and rubric-based protocols \citep{zheng2023judging,gu2024survey,li2024judgesurvey,gunjal2025rubrics,liu2025openrubrics}, which may exhibit self-preference bias \citep{panickssery2024selfpreference}. Preference learning methods such as DPO \citep{rafailov2023dpo} offer a complementary way to optimize properties that are difficult to capture with scalar metrics. 
Because DPO learns from pairwise preferences without an explicit reward model, standard-aware pairs can steer generation toward compliant ADs, which suits accessibility settings, where professional requirements are well documented but rarely enter training in a testable form.

In this work, we study university lecture ADs against an explicit, provenance-tracked standard and address three research questions:

\begin{itemize}[left=0pt, topsep=2pt, itemsep=0pt]
    \item \textbf{RQ1.} Can a rubric judge applying the standard localize rule violations more reliably than reference-based metrics, and can its offline modes rank systems by compliance?
    \item \textbf{RQ2.} Does DPO on standard-ordered preference pairs improve over SFT in rule compliance and automatic metrics?
    \item \textbf{RQ3.} How well do rule compliance, automatic metrics, and machine judgments align with BLV preferences?
\end{itemize}


\section{Data, Model, and Method}
\label{sec:method}

We introduce \textsc{AD4Edu-Bench}, a new benchmark designed to make ADs of slide-based university lectures assessable and trainable against an explicit accessibility standard (Figure~\ref{fig:corpus-construct}) and, to the best of our knowledge, it is the first benchmark for lecture AD with preference data from BLV raters. \textsc{AD4Edu-Bench} consists of four parts: (1) a 45-rule standard for lecture ADs; (2) a corpus of lecture moments with silver reference ADs, together with the SFT checkpoints and preference data derived from them; (3) \textsc{AD4Edu-Eval}, a rule-diagnosable AD evaluation metric operationalized from the 45-rule standard; and (4) a user study where four BLV raters judged 400 AD preference pairs.

\subsection{The \textsc{AD4Edu-Bench} Dataset}
\label{sec:standard}

\textbf{Standard Curation.} We curated a 45-rule standard from five sources of professional AD guidelines: WCAG 2.2 \citep{w3c2024wcag22}, the DCMP Description Key \citep{dcmp2026descriptionkey}, NCAM's STEM description practices \citep{ncam2026stem}, the Netflix style guide \citep{netflix2024adstyle}, and ClearSpeak \citep{frankel2016clearspeak}.
These guidelines define what visual information should be conveyed, but not the lecture-specific constraints: AD must complement the lecturer's speech, resolve deictic references such as ``\textit{this term}'', and fit within short pauses.

We therefore retained all rules applicable to lecture AD and recorded their provenance, yielding 24 documented rules. We added 21 provisional rules for education-specific gaps, including deixis and delivery timing, each with a recorded derivation basis. Thirteen rules were further reviewed by a BLV university staff member with professional experience in digital accessibility, with the verdicts documented in Appendix~\ref{app:rules}. The final standard contains 45 rules: 14 are mechanically checkable, 17 are evaluated by a judge, 12 are unscored for documented reasons, and two are enforced at the sequence level.

\textbf{Corpus Construction.} We selected lecture recordings from the LPM \citep{lee2023lpm} and M3AV collections \citep{chen2024m3av}: 66 recordings from ten lecture series covering six subjects, namely anatomy (two lecturers, 19 recordings), biology (three, 12), psychology (two, 9), dentistry (one, 4), and machine learning (one, 15) from LPM, and the Oxford Mathematics series of M3AV (five, 7). We transcribed each recording with Whisper \citep{radford2023whisper} at word level, detected speech gaps of at least 0.5\,s with the Silero voice activity detector \citep{silero2024vad}, extracted keyframes with FFmpeg \citep{ffmpeg}, and obtained the slide text with EasyOCR \citep{jaided2020easyocr}. A deterministic detector proposed candidate visual events from scene changes, OCR deltas, frame differences, and speech gaps, and a VLM (Gemini 3.1 Pro; \citealp{google2026gemini31}) kept the genuine pedagogical moments, assigning each one of twelve moment types (slide, transition, reveal, pointing, ink, figure, chart, table, math, code, animation, other), and rejected the rest.

For each moment, three VLMs (Gemini 3.1 Pro, \citealp{google2026gemini31}; GPT-5.5, \citealp{openai2026gpt55}; and Claude Sonnet 4.6, \citealp{anthropic2026sonnet46}) received the 45-rule standard together with multimodal context and either generated an AD or chose to remain silent. The context included three or five timestamped stills sampled around the event, a transcript window opening 5\,s before the moment and closing either 8\,s after it or at the next pedagogical moment, OCR from the nearest keyframe, and the longest pause starting within the following 12\,s (Appendix~\ref{app:training}; Appendix~\ref{app:moments} shows six moments). Each (moment, model) call forms one \emph{reference row}, storing the decision, generated AD if any, rationale, and input context. This yielded 6,868 rows over 2,951 moments, of which 1,956 were processed by all three models, 5 by two and 990 by one, including 3,070 ADs across 1,574 moments. Available pauses were short (median 0.5\,s, 90th percentile 1.6\,s).

\textbf{Preference Pairs.} We constructed two types of AD pairs for preference learning:
\begin{itemize}[left=0pt, topsep=2pt, itemsep=0pt]
\item \textbf{natural}: two candidate ADs for the same moment, generated under five focus directives by four VLMs (Gemini 3.1 Pro, GPT-5.5, Claude Sonnet 4.6, and Qwen3-VL-235B \citep{bai2025qwen3vl}) and ranked by a video-watching judge (Gemini 3.1 Pro) viewing a 16\,s clip in both presentation orders;
\item \textbf{controlled}: two factually equivalent ADs differing in exactly one of six rule categories (style, terminology, length, deixis, faithfulness, or non-redundancy).
\end{itemize}
Among 2,961 natural pairs pooled across the two judging runs, 78.9\% received the same preference in both presentation orders, 14.7\% flipped with position, and 6.4\% were ties or unparsed replies; disagreements were retained and flagged. 3,540 pairs (2,501 natural and 1,039 controlled) across 808 moments remained eligible after quality checks. We held out a lecture-disjoint development set of 488 pairs, leaving 3,052 for post-training.

\textbf{BLV Preference Annotation.} For evaluation, we sampled 400 disjoint pairs from 271 moments across 27 lectures: 250 controlled pairs, 100 consistently judged natural pairs, and 50 natural pairs with order-sensitive preferences. Four BLV raters, all Swiss residents, independently evaluated every pair (1,600 judgments) using an audio-first, screen-reader-native interface developed with a Swiss university's equality, diversity, and inclusion office (see Appendix~\ref{app:tool}). Each item presented the lecture context and both ADs using a fixed speech synthesizer; each AD was played on its own, not inserted into the lecture audio. Raters selected A, B, or equal and provided a brief rationale. Presentation order was randomized and AD identity remained masked during the annotation process.

\subsection{The \textsc{AD4Edu-Eval} Metric}
\label{sec:metric}

\textsc{AD4Edu-Eval} operationalizes the standard introduced above, providing three complementary evaluation modes over the same standard. Their outputs are reported separately rather than combined into a single score:
\begin{itemize}[left=0pt, topsep=2pt, itemsep=0pt]
\item \textbf{mechanical}: deterministically evaluates the 14 rules that can be checked without learned models and abstains on the remainder;
\item \textbf{local}: extends the mechanical checks with inference- and embedding-based tests for faithfulness and non-redundancy;
\item \textbf{rubric}: supplies an LLM judge with the moment context, references, and a compressed form of the 45-rule standard, producing six category-level scores and one overall score.
\end{itemize}

\begin{table}[t]
\centering
\caption{\textbf{Offline checks of \textsc{AD4Edu-Eval}.} Anchor: the rule each category's controlled pairs are built on (Appendices~\ref{app:rules} and~\ref{app:examples}). Slide text is the OCR plus the on-screen textual content. 
}
\footnotesize
\setlength{\tabcolsep}{2.5pt}
\resizebox{\columnwidth}{!}{
\begin{tabular}{@{}l l p{4.8cm}@{}}
\toprule
Category & Anchor & Offline check (mode) \\
\midrule
style          & \rulebadge{\textbf{Rule \#2.}}  & forbidden meta-language (``the slide shows'') and a property-based style detector (mech.) \\
terminology    & \rulebadge{\textbf{Rule \#4.}}  & share of content terms found on the slide text (mech.) \\
length         & \rulebadge{\textbf{Rule \#12.}} & word count against the budget of the delivery rung reachable in the pause (mech.) \\
deixis         & \rulebadge{\textbf{Rule \#35.}} & pointing-act phrasing without a named referent (mech.) \\
faithfulness   & \rulebadge{\textbf{Rule \#5.}}  & NLI contradiction test against the slide text, with \texttt{nli-deberta-v3-base} (local) \\
non-redund.    & \rulebadge{\textbf{Rule \#1.}}  & embedding similarity to the transcript window, with \texttt{all-MiniLM-L6-v2} (local) \\
\bottomrule
\end{tabular}
}

\label{tab:metric-map}
\end{table}

The mechanical and local modes share one scoring rule. For the $i$-th description $a_i$ of a lecture and its moment $m$,
\begin{equation}
\small
\begin{aligned}
\mathrm{E}(a_i,m)&=\bar{c}(a_i,m)\cdot\min\!\Big(1,\frac{\mathrm{nov}(a_i)}{\kappa}\Big),\\
\bar{c}(a_i,m)&=\frac{1}{|\mathcal{C}(m)|}\sum_{c\in\mathcal{C}(m)} c(a_i,m),\\
\mathrm{nov}(a_i)&=1-\max_{j<i}\,\mathrm{Jac}\big(\mathrm{tok}(a_i),\mathrm{tok}(a_j)\big),
\end{aligned}
\label{eq:ad4edu}
\end{equation}
where $\bar{c}(a_i,m)$ is the mean over $\mathcal{C}(m)$, the components of Table~\ref{tab:metric-map} applicable to $m$ in the given mode. A component whose rules do not apply to $m$ abstains rather than scores zero, so the mean renormalizes over the components that remain. The novelty term $\mathrm{nov}(a_i)$ is one minus the largest token-set Jaccard overlap ($\mathrm{Jac}$) between $a_i$ and any earlier description $a_j$ of the same lecture; with $\kappa=0.4$, a description whose Jaccard overlap with every earlier description is at most 0.6 receives its full compliance mean, whereas one that repeats itself across the moments of a lecture, such as a slide-title readout, is scaled down (Appendix~\ref{app:robustness} reports the sensitivity of the system ranking to $\kappa$). The novelty term enforces \rulebadge{\textbf{Rule \#6.}} and the once-only clause of \rulebadge{\textbf{Rule \#44.}}; no other check scores \rulebadge{\textbf{Rule \#44.}}. These two modes measure compliance with the standard, not listener preference.

The rubric mode scores the same six categories by judgment. The judge (GPT-5.5; Prompt~\ref{app:prompt-rubric}) grades them against the compressed standard of Prompt~\ref{app:prompt-system}, given the on-screen content, OCR, transcript window, and the moment's references, of which any identical to the candidate is withheld. To compare two descriptions of one moment, the rubric is applied as a cascade: prefer the side with (i)~the higher overall score, else (ii)~the higher sum of its six category scores, else (iii)~the higher chrF to the references; stage (iii) decisions are not attributable to a rule category.

\subsection{Models and Training}
\label{sec:models}

\textbf{SFT Training.} We selected seven instruction-tuned, open-weight VLMs with pinnable revisions to ensure reproducible training: SmolVLM2-2.2B \citep{marafioti2025smolvlm}, Qwen3-VL-2B/4B/8B \citep{bai2025qwen3vl}, InternVL3.5-4B/8B \citep{wang2025internvl35}, and Gemma-3-4B \citep{gemma2025gemma3}. Each model was fine-tuned in multimodal (keyframes plus context) and text-only settings; the text-only setting replaces the keyframes with an annotated description of the screen (Prompt~B.37), so it is caption-conditioned rather than vision-free. We used LoRA \citep{hu2022lora} with rank 16, scaling 32, dropout 0.05, three epochs, learning rate $10^{-4}$, and effective batch size 8. Checkpoints were selected on a moment-disjoint 10\% development slice. Training used one NVIDIA RTX 3090 (24\,GB) per run on an eight-GPU node (see Appendix~\ref{app:training}). We trained with three seeds (0, 1, 2) and report seed-averaged results.
 
\textbf{DPO Post-Training.} Each post-training instance is a triple $(x,y_w,y_l)$, where $x$ is the moment context, $y_w$ the preferred AD, and $y_l$ the alternative. For natural pairs, preference follows the video-watching judge; for controlled pairs, the rule-compliant AD is preferred by construction. We applied DPO \citep{rafailov2023dpo}, initializing the policy $\pi_\theta$ from the SFT checkpoint and keeping the reference policy $\pi_{\mathrm{ref}}$ frozen. Each SFT model was post-trained on 3,052 preference pairs; holding out an in-loop, moment-disjoint development slice of 310 pairs (roughly 10\%) left 2,742 for optimization, for two epochs with learning rate $1.5\times10^{-5}$ and $\beta=0.1$, selected on the 488-pair development set (Appendix~\ref{app:training}). DPO was run from each seed's SFT checkpoint, for all seven text-only models and the two multimodal models that fit on one GPU, and results were averaged across seeds as for SFT.


\section{Evaluation}
\label{sec:evaluation}
\subsection{Automatic Evaluation}
\label{sec:auto-eval}
To evaluate the SFT and DPO checkpoints, we used forced emission to generate ADs for all 516 described test moments, drawn from 17 lectures held out from both SFT and DPO training, and calculated the reference-based metrics chrF \citep{popovic2015chrf} and BERTScore \citep{zhang2020bertscore}, reported after baseline rescaling, the reference-free metric CLIPScore \citep{hessel2021clipscore}, the LLM judge LLM-AD-Eval \citep{han2024autoad3}, and our \textsc{AD4Edu-Eval} (Eq.~\ref{eq:ad4edu}), which ranks systems by compliance with the standard. To validate \textsc{AD4Edu-Eval} as a ranker, we measured on the 250 controlled pairs whether each metric prefers the compliant side, and compared metrics with an exact McNemar test \citep{mcnemar1947} on the pairs both decide.

We then computed win rates \citep{rafailov2023dpo} of the DPO checkpoints against their SFT precursors on a draw of 150 test moments, judged by two machine annotators in both presentation orders. Given moments $\{x_i\}_{i=1}^{N}$ described by both systems, $W_{A\succ B}$ is the proportion of decided moments that $A$ wins in both orders:
\begin{equation}
W_{A\succ B}=\frac{\sum_{i=1}^{N}\mathds{1}[A \text{ wins } x_i \text{ in both orders}]}{\sum_{i=1}^{N}\mathds{1}[x_i \text{ decided}]},
\label{eq:winrate}
\end{equation}
where a moment is \emph{decided} if either system wins it in both orders; we report $W_{\mathrm{SFT}\succ\mathrm{DPO}}$, so values below 0.50 indicate an improvement from post-training.

\subsection{Human Evaluation}
\label{sec:human-eval}

To assess whether the metrics track BLV preferences, we measured agreement with each rater's non-tie labels on the 400 pairs. Per-rater agreement is our primary analysis, with metrics compared using Holm-corrected \citep{holm1979simple} exact McNemar tests and their minimum detectable effects. A preference-aligned metric should exceed chance for every rater. As reference points, we include two machine annotators (GPT-5.6-Terra-Pro and Claude Opus 5) and a fitted preference head, a logistic model over six pairwise metric differences trained on three raters and evaluated once on R3 (Appendix~\ref{app:training}).
\section{Results and Discussion}
\label{sec:results}

\begin{table}[!t]
\centering
\caption{\textbf{Defect localization on the 250 controlled preference pairs}. Cov. is the fraction of pairs decided, and Acc. the accuracy on decided pairs. McNemar tests compare each scorer with chrF; Disc. gives discordant counts (scorer right, chrF right). The rubric-alone test uses the 178 pairs decided by both scorers, whereas the full cascade uses all 250 pairs. Best accuracies are in bold.}
\small
\setlength{\tabcolsep}{3pt}
\resizebox{\columnwidth}{!}{%
\begin{tabular}{@{}lrrcrr@{}}
\toprule
 & & & \multicolumn{3}{c}{McNemar vs.\ chrF} \\
\cmidrule(l){4-6}
Scorer & Cov. & Acc. & Disc. & $p$ & $n$ \\
\midrule
chrF-to-references & 98\% & 61.2 & - & - & -  \\
Mechanical mode & 60\% & 72.2 & - & - & - \\
Local mode & 100\% & 63.6 & (66, 59) & $0.59$ & 245 \\
Rubric alone & 72\% & \textbf{88.3} & (58, 8) & $<10^{-4}$ & 178 \\
\ +~subscore tie-break & 88\% & \textbf{89.1} & - & - & -  \\
\begin{tabular}[t]{@{}l@{}}\ +~chrF residue\\\ \phantom{+}~(full cascade)\end{tabular} & 99\% & 85.4 & (71, 10) & $<10^{-11}$ & 250 \\
\bottomrule
\end{tabular}}

\label{tab:validity}
\vspace{10pt}
\centering
\pdftooltip{\includegraphics[width=0.9\columnwidth]{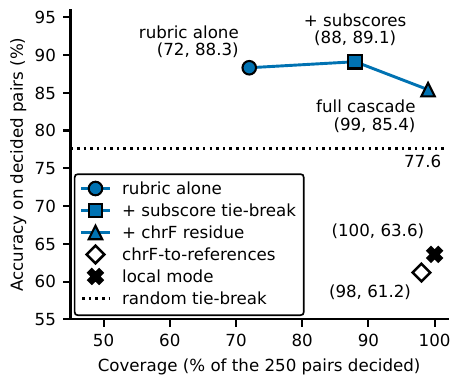}}{Scatter plot of coverage, the percentage of the 250 controlled pairs a scorer decides, on the horizontal axis from 45 to 102, against accuracy on decided pairs on the vertical axis from 55 to 96. A blue line connects three cascade stages: a circle for the rubric alone at coverage 72 and accuracy 88.3, a square for the subscore tie-break at 88 and 89.1, and a triangle for the full cascade with chrF residue at 99 and 85.4, so the line rises slightly then dips while moving right. Two black baselines sit near the bottom right: an open diamond for chrF-to-references at 98 and 61.2 and a filled cross for the local mode at 100 and 63.6. A dotted horizontal line at 77.6 marks random tie-breaking of the rubric's ties at 100 coverage; all three cascade stages lie above it and both baselines below it.}


\captionof{figure}{Coverage versus accuracy on the 250 controlled pairs. Connected points show the three cascade stages, increasing coverage from 72\% to 99\% while retaining higher accuracy than the chrF and local-mode baselines. The dotted line indicates random tie-breaking of rubric ties (77.6\%). Values correspond to Table~\ref{tab:validity}.}
\label{fig:cascade}
\end{table}

\subsection{Metric Validity}
\label{sec:res-metric}

\begin{table*}[t]
\centering
\caption{\textbf{Automatic evaluation of the nine SFT checkpoints with a DPO counterpart and their DPO checkpoints on the 516 held-out moments}. Scores are seed-averaged; DPO values above their SFT precursor are shown in bold blue and below it in plain orange; BERTScore is baseline-rescaled with the raw F1 in parentheses; best SFT value per metric in bold. $^{\dagger}$The two \textsc{AD4Edu-Eval} columns rank systems by compliance with the standard: mechanical is the deterministic kernel alone and local adds the inference- and embedding-based checks; both rank the slide-title readout last.}
\footnotesize
\setlength{\tabcolsep}{3pt}
\resizebox{\textwidth}{!}{
\begin{tabular}{llcccccc}
\toprule
VLM & Setting & chrF & \begin{tabular}[b]{@{}c@{}}BERTScore\\(raw)\end{tabular} & CLIPScore & \begin{tabular}[b]{@{}c@{}}LLM-AD-Eval\\(GPT-3.5 / Llama-2)\end{tabular} & \begin{tabular}[b]{@{}c@{}}\textsc{AD4Edu-Eval}\\mech$^{\dagger}$\end{tabular} & \begin{tabular}[b]{@{}c@{}}\textsc{AD4Edu-Eval}\\local$^{\dagger}$\end{tabular} \\
\specialrule{\lightrulewidth}{\aboverulesep}{0pt}
\rowcolor{gray!20}[2pt][2pt]
\multicolumn{8}{c}{\textbf{SFT checkpoints}} \\
Qwen3-VL-8B & text-only & \textbf{0.344} & \textbf{0.526} (0.885) & 0.656 & 0.635 / \textbf{0.740} & 0.839 & 0.828 \\
Qwen3-VL-4B & text-only & 0.326 & 0.503 (0.879) & 0.651 & 0.617 / 0.734 & 0.837 & 0.821 \\
Qwen3-VL-2B & text-only & 0.316 & 0.505 (0.880) & 0.658 & 0.602 / 0.739 & 0.837 & 0.828 \\
Qwen3-VL-2B & multimodal & 0.328 & 0.514 (0.882) & 0.663 & 0.615 / 0.737 & 0.778 & 0.768 \\
InternVL3.5-8B & text-only & 0.333 & 0.509 (0.881) & 0.661 & \textbf{0.638} / 0.739 & 0.847 & 0.832 \\
InternVL3.5-4B & text-only & 0.321 & 0.511 (0.881) & 0.650 & 0.609 / 0.732 & 0.813 & 0.812 \\
SmolVLM2-2.2B & text-only & 0.298 & 0.481 (0.874) & 0.651 & 0.560 / 0.734 & \textbf{0.858} & \textbf{0.838} \\
SmolVLM2-2.2B & multimodal & 0.271 & 0.462 (0.869) & \textbf{0.668} & 0.522 / 0.728 & 0.794 & 0.781 \\
Gemma-3-4B & text-only & 0.303 & 0.504 (0.880) & 0.660 & 0.567 / 0.735 & 0.850 & 0.837 \\
\specialrule{\lightrulewidth}{\aboverulesep}{0pt}
\rowcolor{gray!20}[2pt][2pt]
\multicolumn{8}{c}{\textbf{DPO checkpoints}} \\
Qwen3-VL-8B & text-only & \textcolor{drop}{0.321} & \textcolor{drop}{0.473} & \textcolor{gain}{\textbf{0.683}} & \textcolor{gain}{\textbf{0.687}} / \textcolor{gain}{\textbf{0.745}} & \textcolor{drop}{0.745} & \textcolor{drop}{0.785} \\
Qwen3-VL-4B & text-only & \textcolor{drop}{0.305} & \textcolor{drop}{0.470} & \textcolor{gain}{\textbf{0.686}} & \textcolor{gain}{\textbf{0.687}} / \textcolor{gain}{\textbf{0.745}} & \textcolor{drop}{0.777} & \textcolor{drop}{0.806} \\
Qwen3-VL-2B & text-only & \textcolor{drop}{0.301} & \textcolor{drop}{0.464} & \textcolor{gain}{\textbf{0.677}} & \textcolor{gain}{\textbf{0.655}} / \textcolor{gain}{\textbf{0.740}} & \textcolor{drop}{0.756} & \textcolor{drop}{0.791} \\
Qwen3-VL-2B & multimodal & \textcolor{drop}{0.314} & \textcolor{drop}{0.414} & \textcolor{gain}{\textbf{0.775}} & \textcolor{gain}{\textbf{0.722}} / \textcolor{gain}{\textbf{0.742}} & \textcolor{drop}{0.630} & \textcolor{drop}{0.664} \\
InternVL3.5-8B & text-only & \textcolor{drop}{0.299} & \textcolor{drop}{0.468} & \textcolor{gain}{\textbf{0.676}} & \textcolor{gain}{\textbf{0.661}} / \textcolor{gain}{\textbf{0.740}} & \textcolor{drop}{0.813} & \textcolor{gain}{\textbf{0.834}} \\
InternVL3.5-4B & text-only & \textcolor{drop}{0.292} & \textcolor{drop}{0.460} & \textcolor{gain}{\textbf{0.689}} & \textcolor{gain}{\textbf{0.656}} / \textcolor{gain}{\textbf{0.740}} & \textcolor{drop}{0.810} & \textcolor{gain}{\textbf{0.832}} \\
SmolVLM2-2.2B & text-only & \textcolor{drop}{0.215} & \textcolor{drop}{0.374} & \textcolor{gain}{\textbf{0.679}} & \textcolor{drop}{0.476} / \textcolor{drop}{0.727} & \textcolor{drop}{0.835} & \textcolor{drop}{0.816} \\
SmolVLM2-2.2B & multimodal & \textcolor{drop}{0.204} & \textcolor{drop}{0.333} & \textcolor{gain}{\textbf{0.695}} & \textcolor{drop}{0.344} / \textcolor{drop}{0.718} & \textcolor{drop}{0.718} & \textcolor{drop}{0.695} \\
Gemma-3-4B & text-only & \textcolor{drop}{0.288} & \textcolor{drop}{0.458} & \textcolor{gain}{\textbf{0.673}} & \textcolor{gain}{\textbf{0.624}} / \textcolor{gain}{\textbf{0.741}} & \textcolor{drop}{0.751} & \textcolor{drop}{0.794} \\
\midrule
Slide-title readout & -- & 0.152 & 0.340 (0.840) & 0.735 & 0.220 / 0.688 & 0.532 & 0.501 \\
\bottomrule
\end{tabular}
}

\label{tab:systems}
\end{table*}

To answer RQ1, Table~\ref{tab:validity} reports whether each scorer prefers the compliant side of the controlled pairs. The rubric judge decides 72\% of pairs and is right on 88.3\% of them; its subscore tie-break raises accuracy to 89.1\% at 88\% coverage, and the full cascade reaches 85.4\% at 99\% coverage (Figure~\ref{fig:cascade}). It is the only scorer that significantly outperforms chrF (58 discordant pairs against 8, $p<10^{-4}$), whereas the local mode is indistinguishable from it ($p=0.59$). Without knowing the differing category, the rubric still finds the defect on over 85\% of pairs in five categories but on 62.1\% for non-redundancy (Table~\ref{tab:localization}). These accuracies are an upper bound, since the controlled defects follow the rubric's own categories; a judge that wrote no reference (GPT-3.5 Turbo) agrees with GPT-5.5 on 93.5\% of the pairs both decide (Appendix~\ref{app:robustness}). The mechanical mode abstains on 40\% of the pairs and prefers the compliant side on 72.2\% of the rest, as expected of a compliance index rather than a pairwise judge. At the system level, the mechanical and local columns of Table~\ref{tab:systems} order the systems consistently across $\kappa$ (Appendix~\ref{app:robustness}), place the slide-title readout last, score the three reference writers at 0.80--0.82 (Table~\ref{tab:refwriters}), and are uncorrelated with chrF ($\rho=0.06$ and $-0.16$) and with LLM-AD-Eval ($-0.40$ and $-0.13$, intervals covering zero; Appendix~\ref{app:robustness}): compliance orders systems along an axis the reference-based metrics do not measure. Table~\ref{tab:properties} summarizes five validity properties per metric: only \textsc{AD4Edu-Eval} is rule-diagnosable, and no metric has all five.

\subsection{Post-training from SFT}
\label{sec:res-training}

\begin{table}[!htb]
\centering
\caption{\textbf{Head-to-head win rate on the 150-moment draw} for the seed-0 text-only Qwen3-VL-8B checkpoints, reported as $W_{\mathrm{SFT}\succ\mathrm{DPO}}$ (Eq.~\ref{eq:winrate}), so values above 0.50 favor SFT; $n$ is the number of decided moments and CIs are lecture-clustered. The judges differ in tie frequency, not in direction.}
\footnotesize
\setlength{\tabcolsep}{4pt}
\begin{tabular}{llrrr}
\toprule
Comparison & Judge & $W$ & 95\% CI & $n$ \\
\midrule
SFT vs.\ DPO & GPT & .545 & [.387, .711] & 101 \\
SFT vs.\ DPO & Claude & .625 & [.467, .781] & 112 \\
\bottomrule
\end{tabular}
\label{tab:winrates}
\end{table}

To answer RQ2, Tables~\ref{tab:systems} and~\ref{tab:winrates} compare SFT with DPO post-training. Under SFT alone, no configuration leads reliably: text-only Qwen3-VL-8B attains the highest chrF and BERTScore and multimodal SmolVLM2 the highest CLIPScore, whereas text-only models attain the highest compliance (0.813--0.858 against 0.778--0.794 for the two multimodal checkpoints).

\begin{table*}[t]
\centering
\caption{\textbf{Validity properties of the seven metrics.} Each cell gives the verdict and its supporting figure. \emph{Rule-diagnosable}: rule categories (of six) to which a failure can be attributed. \emph{Prefers compliant side}: accuracy and pairs decided on the 250 controlled pairs (Table~\ref{tab:validity}); partial if a metric decides a minority of pairs or falls to chance when both presentation orders must agree (Appendix~\ref{app:robustness}). \emph{Ranks readout below systems}: rank of the slide-title readout among the 19 rows of Table~\ref{tab:systems} (three systems of Table~\ref{tab:rubric} for judge-based metrics) and its margin to the best trained row. \emph{Above chance, every rater}: raters (of four) with agreement significant at $p<0.05$, and the lowest per-rater agreement (Table~\ref{tab:human}). The rubric row uses the cascade on controlled pairs and the rubric score elsewhere.}

\footnotesize
\setlength{\tabcolsep}{3pt}
\resizebox{\textwidth}{!}{%
\begin{tabular}{@{}lccccc@{}}
\toprule
Metric & \begin{tabular}[b]{@{}c@{}}Rule-\\diagnosable\\(categories of 6)\end{tabular} & \begin{tabular}[b]{@{}c@{}}Prefers\\compliant side\\(\%, $n$ decided)\end{tabular} & \begin{tabular}[b]{@{}c@{}}Ranks readout\\below systems\\(rank, margin)\end{tabular} & \begin{tabular}[b]{@{}c@{}}Above chance,\\every rater\\(raters of 4, min.\ \%)\end{tabular} & \begin{tabular}[b]{@{}c@{}}Offline,\\no API\end{tabular} \\
\midrule
chrF-to-references & no (0) & yes (61.2, 245) & yes (19/19, $-$0.192) & yes (4, 57.7) & yes \\
BERTScore & no (0) & yes (71.0, 248) & partial (18/19, $-$0.186) & partial (1, 51.2) & yes \\
CLIPScore & no (0) & no (48.8, 250) & no (2/19, $-$0.040) & partial (1, 47.6) & yes \\
LLM-AD-Eval & no (0) & partial (84.3, 83) & yes (3/3, $-$0.443 / $-$0.063) & yes (4, 67.8) & no \\
\hdashline
\rowcolor{gray!20}[2pt][2pt]
\multicolumn{6}{c}{\textbf{Ours}} \\
\textsc{AD4Edu-Eval}-mechanical & yes (4) & partial (72.2, 151) & yes (19/19, $-$0.326) & no (0, 42.8) & yes \\
\textsc{AD4Edu-Eval}-local & yes (6) & yes (63.6, 250) & yes (19/19, $-$0.337) & no (0, 41.5) & yes \\
\textsc{AD4Edu-Eval}-rubric (cascade) & yes (6) & yes (85.4, 247) & yes (3/3, $-$4.35) & partial (3, 53.7) & no \\
\bottomrule
\end{tabular}}

\label{tab:properties}
\end{table*}

DPO post-training does not improve on this baseline. It lowers chrF and BERTScore and raises CLIPScore on every row; LLM-AD-Eval rises on seven of nine rows; and \textsc{AD4Edu-Eval} compliance falls below the SFT precursor on every row in the mechanical mode and on seven of nine in the local mode. Head-to-head (Table~\ref{tab:winrates}), DPO does not beat its SFT precursor at a win rate whose lecture-clustered interval excludes 0.50. Only the rubric mode ranks the DPO checkpoint first (6.32 against 5.37; Table~\ref{tab:rubric}), disagreeing with every other signal in the suite. 
The answer to RQ2 is thus negative except on the judge-based scores, a split that RQ3 examines.

\begin{table}[t]
\centering

\caption{\textbf{Rubric-mode \textsc{AD4Edu-Eval} on the 150-moment sample from Table~\ref{tab:winrates}}. Scores are means on a 1--10 scale with lecture-clustered 95\% CIs. \(n\) denotes scored moments; Undesc. counts moments left undescribed under best-effort forced emission and excluded from scoring. DPO and SFT are seed-0 text-only Qwen3-VL-8B checkpoints.}
\footnotesize
\setlength{\tabcolsep}{4pt}
\begin{tabular}{lrrrr}
\toprule
System & Rubric & 95\% CI & $n$ & Undesc. \\
\midrule
DPO & 6.32 & [5.96, 6.63] & 145 & 5 \\
SFT & 5.37 & [4.93, 5.81] & 150 & 0 \\
Readout & 1.97 & [1.75, 2.27] & 150 & 0 \\
\bottomrule
\end{tabular}
\label{tab:rubric}
\end{table}

Recent studies have shown limits of DPO that bear on this outcome: its implicit rewards can overfit and grow unbounded \citep{fisch2024robust}; it forgets earlier behavior across data regimes \citep{qi2024onlinedpo}; and it regularizes responses of different quality unevenly \citep{hu2024desiderata}. Here, the training pairs were ordered by a judge whose verdict flipped with presentation order on 14.7\% of pairs, and preference accuracy peaks at step 400 of 686 and recedes thereafter. Lower-burden objectives such as KTO \citep{ethayarajh2024kto}, or objectives that oblige the AD to add unspoken content, may suit this setting better.

\subsection{Transfer to BLV Preferences}
\label{sec:res-human}

To answer RQ3, Table~\ref{tab:human} shows how well each metric reproduces the raters' choices. The judge-based metrics and the fitted preference head agree with individual raters on 54--77\%, chrF on 58--66\%, and the mechanical and local modes not significantly above chance for any rater, an ordering that replicates on the held-out rater R3. The rubric tracks preference only moderately; its subscores agree with the raters on 89.1\% of decided deixis labels ($n=129$) but on only 16.5\% of decided length labels ($n=97$), where raters prefer the longer side the standard penalizes (all six categories in Table~\ref{tab:percat-raters}). The raters themselves agree above chance on controlled pairs (70.4\%) but near chance on natural ones (52.8\%), so varying preference, close pairs, and a demanding listening task cannot be separated. The disagreement lies between raters rather than between machines and raters: scored like a metric, the other three raters agree with a given rater on 64.1\% of pairs on average, the two machine annotators on 65.6\% and 67.4\%, and on the 57 pairs all four raters agree, every judge-based metric exceeds 87\%.

Compliance and BLV preference thus appear to be distinct constructs, although raters heard each AD in isolation, so a longer description never cost them lecture audio. This motivates a layered reading of the standard, in which deixis, on which rubric and raters agree, is a candidate hard constraint; concision, terminology, and delivery conventions are professional defaults open to revision; and length and tolerated redundancy, where the raters contradict the standard, are user-dependent preferences better routed to user-level adaptation than to global post-training. This layering is a hypothesis generated by four raters on 400 pairs.

\begin{table}[!tb]
\centering
\caption{\textbf{Agreement of each metric with each BLV rater's non-tie preferences on the 400 pairs} (\%). Judge-based metrics are evaluated on the pairs they decide. Inter-rater agreement is Krippendorff's $\alpha=0.14$ \citep{krippendorff2018content} over three labels. The fitted preference head is trained on three raters and evaluated out-of-fold, with R3 tested once using frozen weights. LLM-AD-Eval uses GPT-5.5.
}
\small
\begin{tabular}{lrrrr}
\toprule
Metric & R1 & R2 & R3 & R4 \\
\midrule
chrF-to-references & 62.3 & 65.6 & 57.7 & 58.6 \\
BERTScore & 55.4 & 53.1 & 58.5 & 51.2 \\
CLIPScore & 47.6 & 56.0 & 54.7 & 55.5 \\
LLM-AD-Eval & 74.8 & \textbf{67.8} & 69.9 & \textbf{70.6} \\
Machine annotator (Claude) & 69.5 & 56.3 & \textbf{74.5} & 62.1 \\
Machine annotator (GPT) & 73.5 & 57.1 & 73.7 & 65.0 \\
Fitted preference head & \textbf{77.2} & 63.3 & 70.3 & 70.2 \\
\textsc{AD4Edu-Eval}-mechanical & 45.3 & 45.5 & 51.4 & 42.8 \\
\textsc{AD4Edu-Eval}-local & 46.1 & 44.9 & 52.5 & 41.5 \\
\textsc{AD4Edu-Eval}-rubric & 72.6 & 53.7 & 67.2 & 60.8 \\
\bottomrule
\end{tabular}
\label{tab:human}
\end{table}

\section{Conclusion and Future Work}
\label{sec:conclusion}

In this work, we studied whether AD of slide-based university lectures can be evaluated and trained against an explicit accessibility standard. We consolidated professional guidance into a 45-rule standard, operationalized it as \textsc{AD4Edu-Eval}, a rule-diagnosable metric that ranks systems by compliance, and introduced \textsc{AD4Edu-Bench}, to the best of our knowledge, the first benchmark for lecture AD with preference data from BLV raters. Our findings expose a limitation of treating compliance as quality: the rubric localizes controlled rule violations reliably and DPO over judge-ordered preferences did not improve on SFT, yet compliance aligned with BLV judgments on deixis and diverged on length and redundancy, suggesting that the two are complementary constructs.

In future work, we will partition the standard into validated constraints, professional defaults, and user-dependent preferences with BLV raters, collect preferences from more raters, and explore alignment objectives that learn from lower-burden feedback. More broadly, we advocate for accessibility research that involves BLV persons not merely as evaluators but as co-creators of the standards systems are trained against.

\section*{Limitations}
Our study has several limitations. First, the human study rests on four BLV raters whose agreement is low ($\alpha=0.14$) and near chance on natural pairs, so individual taste, close natural pairs, and the demands of judging by ear cannot be separated. Second, with 93 to 343 decided pairs per rater and metric, the study is powered only for coarse differences; eleven of the twelve per-rater gaps we tested lie below their minimum detectable effects (Appendix~\ref{app:training}). Third, the corpus covers lectures with OCR-readable slides and a pointing-dominated moment distribution, and the system evaluation and the rating study share only one subject (Appendix~\ref{app:training}), so the findings may not transfer to other subjects or to lecture styles with few on-slide visuals or different pacing. Fourth, the training pairs were ordered by a machine judge rather than by BLV raters, so the post-training arms never learned from the preferences they were scored against. Fifth, GPT-5.5 wrote roughly a third of the reference ADs and is also the rubric judge, which is given those references; a judge that wrote no reference agrees with it on 93.5\% of the controlled pairs both decide (Appendix~\ref{app:robustness}), which bounds but does not remove the concern. Finally, the references are silver, written by rule-conditioned VLMs and not checked by professional describers, which bounds every reference-based score.

\section*{Ethical Considerations}

This study involved the collection of pairwise preference judgments from BLV students. The study was registered with the ethics committee of the faculty at the authors' institution. Participation was voluntary and without compensation; each rater read the participant information sheet inside the rating tool and gave consent before any item was shown, and could stop at any time by closing the tool.

Raters provided a name or initials for session continuity, which we replaced with pseudonymous identifiers R1--R4 in every analysis reported here and do not release; for each item we recorded the preference choice, a one-line free-text reason, and the time spent, and at consent we recorded age and field of study. The rating tool ran offline on the rater's own device, raters returned their answer files to the authors, and no audio or video of participants was captured. 

The lecture recordings are third-party academic lectures whose copyright remains with their authors; we do not redistribute them. Stills from the recordings are reproduced only in the appendices: inside the tool screenshot of Figure~\ref{fig:tool-item}, in the player screenshots of Figures~\ref{fig:player-dna-replication}--\ref{fig:player-anaphase}, and in the six example moments of Appendix~\ref{app:moments}; their copyright remains with their publishers and they are reproduced under fair dealing for criticism and review. No recording is released. 

Thirteen of the standard's rules were reviewed by a BLV collaborator before the standard was used to score or train any system, and the rating tool was built to be operated by screen reader from the outset rather than adapted afterwards. Machine-generated descriptions can misdescribe visual content; the benchmark measures faithfulness against the slide text and references but does not certify any system for deployment, and we caution against using the trained models with students without human review.

\section*{Acknowledgement}

This study was supported by the Department of Computational Linguistics (CL) and the Office for Equality, Diversity and Inclusion (EDI) at the University of Zurich. We sincerely thank Daniele Corciulo for his expertise in digital inclusion and valuable feedback on accessible software implementation, as well as our BLV participants for their precious contributions to the co-design of this research.

\bibliography{custom}

@inproceedings{han2023autoad,
  title={{AutoAD: Movie Description in Context}},
  author={Han, Tengda and Bain, Max and Nagrani, Arsha and Varol, G{\"u}l and Xie, Weidi and Zisserman, Andrew},
  booktitle={2023 IEEE/CVF Conference on Computer Vision and Pattern Recognition (CVPR)},
  pages={18930--18940},
  year={2023},
  url={https://ieeexplore.ieee.org/abstract/document/10204048}
}

@inproceedings{han2023autoad2,
  title={{AutoAD II: The Sequel--Who, When, and What in Movie Audio Description}},
  author={Han, Tengda and Bain, Max and Nagrani, Arsha and Varol, G{\"u}l and Xie, Weidi and Zisserman, Andrew},
  booktitle={2023 IEEE/CVF International Conference on Computer Vision (ICCV)},
  pages={13599--13609},
  year={2023},
  url={https://ieeexplore.ieee.org/abstract/document/10378433}
}

@inproceedings{han2024autoad3,
  title={{AutoAD III: The Prequel-Back to the Pixels}},
  author={Han, Tengda and Bain, Max and Nagrani, Arsha and Varol, G{\"u}l and Xie, Weidi and Zisserman, Andrew},
  booktitle={2024 IEEE/CVF Conference on Computer Vision and Pattern Recognition (CVPR)},
  pages={18164--18174},
  year={2024},
  url={https://ieeexplore.ieee.org/abstract/document/10656902}
}

@inproceedings{narrad2025,
  title={{NarrAD: Automatic Generation of Audio Descriptions for Movies with Rich Narrative Context}},
  author={Park, Jaehyeong and Ye, Junchel and Lee, Seungkook and Ka, Hyun W. and Han, Dongsu},
  booktitle={Proceedings of the Winter Conference on Applications of Computer Vision (WACV)},
  pages={409--419},
  year={2025},
  url={https://openaccess.thecvf.com/content/WACV2025/html/Park_NarrAD_Automatic_Generation_of_Audio_Descriptions_for_Movies_with_Rich_WACV_2025_paper.html}
}

@inproceedings{distinctad2025,
  title={{DistinctAD: Distinctive Audio Description generation in contexts}},
  author={Fang, Bo and Wu, Wenhao and Wu, Qiangqiang and Song, Yuxin and Chan, Antoni B},
  booktitle={2025 IEEE/CVF Conference on Computer Vision and Pattern Recognition (CVPR)},
  pages={13571--13581},
  year={2025},
  url={https://ieeexplore.ieee.org/abstract/document/11093911}
}

@inproceedings{gao2024adsurvey,
  title={{Audio Description Generation in the Era of LLMs and VLMs: A Review of Transferable Generative AI Technologies}},
  author={Gao, Yingqiang and Fischer, Lukas and Lintner, Alexa and Ebling, Sarah},
  booktitle={Findings of the Association for Computational Linguistics: NAACL 2025},
  pages={471--490},
  year={2025},
  url={https://aclanthology.org/2025.findings-naacl.29/}
}

@article{chu2024llmadlargelanguagemodel,
  title={{LLM-AD: Large Language Model based Audio Description System}},
  author={Chu, Peng and Wang, Jiang and Abrantes, Andre},
  journal={arXiv preprint arXiv:2405.00983},
  year={2024},
  url={https://arxiv.org/abs/2405.00983}, 
}

@inproceedings{autoadzero2024,
    title={{AutoAD-Zero: A Training-Free Framework for Zero-Shot Audio Description}},
    author={Xie, Junyu and Han, Tengda and Bain, Max and Nagrani, Arsha and Varol, G\"ul and Xie, Weidi and Zisserman, Andrew},
    booktitle={Proceedings of the Asian Conference on Computer Vision (ACCV)},
    month={December},
    year={2024},
    pages={2265-2281},
    url={https://openaccess.thecvf.com/content/ACCV2024/html/Xie_AutoAD-Zero_A_Training-Free_Framework_for_Zero-Shot_Audio_Description_ACCV_2024_paper.html}
}

@inproceedings{videoa11y2025,
  title={{VideoA11y: Method and Dataset for Accessible Video Description}}, 
  author={Li, Chaoyu and Padmanabhuni, Sid and Cheema, Maryam S and Seifi, Hasti and Fazli, Pooyan},
  year={2025},
  pages={1-29},
  booktitle= {Proceedings of the 2025 Conference on Human Factors in Computing Systems (CHI)},
  url={https://dl.acm.org/doi/full/10.1145/3706598.3714096}, 
}

@inproceedings{cheema2025nowuserdrivenaudiodescription,
  title={{Describe Now: User-driven Audio Description for Blind and Low Vision Individuals}},
  author={Cheema, Maryam and Seifi, Hasti and Fazli, Pooyan},
  booktitle={Proceedings of the 2025 ACM Designing Interactive Systems Conference (DIS)},
  pages={458--474},
  year={2025},
  url={https://dl.acm.org/doi/full/10.1145/3715336.3735685}
}

@inproceedings{anderer2025lecture,
  title={{Making Lecture Videos Accessible for Students who are Blind or have Low Vision through AI-Assisted Navigation and Visual Question Answering}},
  author={Anderer, Katharina and M{\"u}ller, Karin and Strobel, Lukas and W{\"o}lfel, Matthias and Niehues, Jan and Gerling, Kathrin},
  booktitle={Proceedings of the 27th International ACM SIGACCESS Conference on Computers and Accessibility (ASSETS)},
  pages={1--15},
  year={2025},
  url={https://dl.acm.org/doi/full/10.1145/3663547.3746349}
}

@inproceedings{sechayk2025veasyguide,
  title={{VeasyGuide: Personalized Visual Guidance for Low-vision Learners on Instructor Actions in Presentation Videos}},
  author={Sechayk, Yotam and Shamir, Ariel and Pavel, Amy and Igarashi, Takeo},
  booktitle={Proceedings of the 27th International ACM SIGACCESS Conference on Computers and Accessibility (ASSETS)},
  pages={1--18},
  year={2025},
  url={https://dl.acm.org/doi/full/10.1145/3663547.3746372},
}

@article{do2026scalableaudiodescriptionquality,
  title={{Toward Scalable Audio Description Quality Control: A Workflow for Evaluating Human and VLM Raters}},
  author={Do, Lana and Jung, Gio and Barajas, Juvenal Francisco and Scott, Andrew Taylor and Ihorn, Shasta and Blum, Alexander Mario and Athitsos, Vassilis and Yoon, Ilmi},
  journal={arXiv preprint arXiv:2602.01390},
  year={2026},
  url={https://arxiv.org/abs/2602.01390}
}

@inproceedings{Jiang_2024,
  title={{“It’s Kind of Context Dependent”: Understanding Blind and Low Vision People’s Video Accessibility Preferences Across Viewing Scenarios}},
  author={Jiang, Lucy and Jung, Crescentia and Phutane, Mahika and Stangl, Abigale and Azenkot, Shiri},
  booktitle={Proceedings of the 2024 Conference on Human Factors in Computing Systems (CHI)},
  pages={1--20},
  year={2024},
  url={https://dl.acm.org/doi/full/10.1145/3613904.3642238}
}

@article{lee2023lpm,
  title={{Multimodal Lecture Presentations Dataset: Understanding Multimodality in Educational Slides}},
  author={Lee, Dong Won and Ahuja, Chaitanya and Liang, Paul Pu and Natu, Sanika and Morency, Louis-Philippe},
  journal={arXiv preprint arXiv:2208.08080},
  year={2022},
  url={https://arxiv.org/abs/2208.08080}
}

@inproceedings{chen2024m3av,
  title={{M3AV: A Multimodal, Multigenre, and Multipurpose Audio-Visual Academic Lecture Dataset}},
  author={Chen, Zhe and Liu, Heyang and Yu, Wenyi and Sun, Guangzhi and Liu, Hongcheng and Wu, Ji and Zhang, Chao and Wang, Yu and Wang, Yanfeng},
  booktitle={Proceedings of the 62nd Annual Meeting of the Association for Computational Linguistics (Volume 1: Long Papers)},
  pages={9041--9060},
  year={2024},
  url={https://aclanthology.org/2024.acl-long.489/}
}

@misc{w3c2024wcag22,
  title={{Web Content Accessibility Guidelines (WCAG) 2.2}},
  author={{World Wide Web Consortium (W3C)}},
  year={2024},
  howpublished={W3C Recommendation, 12 December 2024},
  url={https://www.w3.org/TR/WCAG22/}
}

@misc{dcmp2026descriptionkey,
  title={{Description Key: Guidelines for Description of Educational Media}},
  author={{Described and Captioned Media Program (DCMP)}},
  year={2026},
  howpublished={Described and Captioned Media Program, U.S. Department of Education},
  url={https://dcmp.org/learn/descriptionkey},
  note={Accessed July 2026}
}

@misc{ncam2026stem,
  title={{Effective Practices for Description of Science Content within Digital Talking Books}},
  author={{Carl and Ruth Shapiro Family National Center for Accessible Media (NCAM), WGBH}},
  year={2026},
  howpublished={WGBH Educational Foundation},
  url={https://www.wgbh.org/ncam-resources/2026-08-25/effective-practices-for-description-of-science-content-digital-talking-books},
  note={Accessed July 2026}
}

@misc{w3c2015maur,
  title={{Media Accessibility User Requirements}},
  author={{World Wide Web Consortium (W3C)}},
  year={2015},
  howpublished={W3C Working Group Note, 3 December 2015},
  url={https://www.w3.org/TR/media-accessibility-reqs/}
}

@misc{netflix2024adstyle,
  title={{Audio Description Style Guide v2.5}},
  author={{Netflix}},
  year={2026},
  howpublished={Netflix Partner Help Center},
  url={https://partnerhelp.netflixstudios.com/hc/en-us/articles/215510667-Audio-Description-Style-Guide-v2-5},
  note={Accessed July 2026}
}

@inproceedings{pavel2020rescribe,
  title={{Rescribe: Authoring and Automatically Editing Audio Descriptions}},
  author={Pavel, Amy and Reyes, Gabriel and Bigham, Jeffrey P},
  booktitle={Proceedings of the 33rd Annual ACM Symposium on User Interface Software and Technology (UIST)},
  pages={747--759},
  year={2020},
  url={https://dl.acm.org/doi/abs/10.1145/3379337.3415864}
}

@article{frankel2016clearspeak,
  title={{Development and Initial Evaluation of the ClearSpeak Style for Automated Speaking of Algebra}},
  author={Frankel, Lois and Brownstein, Beth and Soiffer, Neil and Hansen, Eric},
  journal={ETS Research Report Series},
  volume={2016},
  number={2},
  pages={1--43},
  year={2016},
  publisher={Wiley Online Library},
  url={https://onlinelibrary.wiley.com/doi/full/10.1002/ets2.12103}
}

@inproceedings{popovic2015chrf,
  title={{chrF: Character n-gram F-score for Automatic MT Evaluation}},
  author={Popovi{\'c}, Maja},
  booktitle={Proceedings of the Tenth Workshop on Statistical Machine Translation},
  pages={392--395},
  year={2015},
  url={https://aclanthology.org/W15-3049.pdf}
}

@inproceedings{post2018sacrebleu,
  title={{A Call for Clarity in Reporting BLEU Scores}},
  author={Post, Matt},
  booktitle={Proceedings of the Third Conference on Machine Translation (WMT): Research Papers},
  pages={186--191},
  year={2018},
  url={https://aclanthology.org/W18-6319/}
}

@inproceedings{zhang2020bertscore,
  title={{BERTScore: Evaluating Text Generation with BERT}},
  author={Zhang, Tianyi and Kishore, Varsha and Wu, Felix and Weinberger, Kilian Q. and Artzi, Yoav},
  booktitle={Proceedings of the Eighth International Conference on Learning Representations (ICLR)},
  year={2020},
  pages={1-43},
  url={https://openreview.net/pdf?id=SkeHuCVFDr}
}

@inproceedings{hessel2021clipscore,
  title={{CLIPScore: A Reference-free Evaluation Metric for Image Captioning}},
  author={Hessel, Jack and Holtzman, Ari and Forbes, Maxwell and Le Bras, Ronan and Choi, Yejin},
  booktitle={Proceedings of the 2021 Conference on Empirical Methods in Natural Language Processing (EMNLP)},
  pages={7514--7528},
  year={2021},
  url={https://aclanthology.org/2021.emnlp-main.595/}
}

@inproceedings{vedantam2015cider,
  title={{CIDEr: Consensus-based Image Description Evaluation}},
  author={Vedantam, Ramakrishna and Zitnick, C. Lawrence and Parikh, Devi},
  booktitle={Proceedings of the IEEE Conference on Computer Vision and Pattern Recognition (CVPR)},
  year={2015},
  pages={4566-4575},
  url={https://www.cv-foundation.org/openaccess/content_cvpr_2015/html/Vedantam_CIDEr_Consensus-Based_Image_2015_CVPR_paper.html}
}

@inproceedings{anderson2016spice,
  title={{SPICE: Semantic Propositional Image Caption Evaluation}},
  author={Anderson, Peter and Fernando, Basura and Johnson, Mark and Gould, Stephen},
  booktitle={Proceedings of the 14th European Conference on Computer Vision (ECCV)},
  year={2016},
  pages={382–398},
  url={https://link.springer.com/chapter/10.1007/978-3-319-46454-1_24}
}

@inproceedings{hu2023tifa,
  title={{TIFA: Accurate and Interpretable Text-to-Image Faithfulness Evaluation with Question Answering}},
  author={Hu, Yushi and Liu, Benlin and Kasai, Jungo and Wang, Yizhong and Ostendorf, Mari and Krishna, Ranjay and Smith, Noah A.},
  booktitle={2023 IEEE/CVF International Conference on Computer Vision (ICCV)},
  year={2023},
  pages={20349-20360},
  url={https://ieeexplore.ieee.org/abstract/document/10377168}
}

@inproceedings{li2023pope,
  title={{Evaluating Object Hallucination in Large Vision-Language Models}},
  author={Li, Yifan and Du, Yifan and Zhou, Kun and Wang, Jinpeng and Zhao, Xin and Wen, Ji-Rong},
  booktitle={Proceedings of the 2023 Conference on Empirical Methods in Natural Language Processing (EMNLP)},
  year={2023},
  pages={292--305},
  url={https://aclanthology.org/2023.emnlp-main.20/},
}

@inproceedings{rohrbach2018chair,
  title={{Object Hallucination in Image Captioning}},
  author={Rohrbach, Anna and Hendricks, Lisa Anne and Burns, Kaylee and Darrell, Trevor and Saenko, Kate},
  booktitle={Proceedings of the 2018 Conference on Empirical Methods in Natural Language Processing (EMNLP)},
  year={2018},
  pages={4035--4045},
  url={https://aclanthology.org/D18-1437/},
}

@inproceedings{kala2025wysiwa,
  title={{What You See is What You Ask: Evaluating Audio Descriptions}},
  author={Kala, Divy and Khandelwal, Eshika and Tapaswi, Makarand},
  booktitle={Proceedings of the 2025 Conference on Empirical Methods in Natural Language Processing (EMNLP)},
  year={2025},
  pages={23496--23518},
  url={https://aclanthology.org/2025.emnlp-main.1199/}
}

@inproceedings{rafailov2023dpo,
  title={{Direct Preference Optimization: Your Language Model is Secretly a Reward Model}},
  author={Rafailov, Rafael and Sharma, Archit and Mitchell, Eric and Manning, Christopher D. and Ermon, Stefano and Finn, Chelsea},
  booktitle={Advances in Neural Information Processing Systems (NeurIPS)},
  year={2023},
  pages={53728--53741},
  url={https://papers.nips.cc/paper_files/paper/2023/hash/a85b405ed65c6477a4fe8302b5e06ce7-Abstract-Conference.html}
}

@inproceedings{liu2025openrubrics,
  title={{OpenRubrics: Towards Scalable Synthetic Rubric Generation for Reward Modeling and LLM Alignment}},
  author={Liu, Tianci and Xu, Ran and Yu, Tony and Hong, Ilgee and Yang, Carl and Zhao, Tuo and Wang, Haoyu},
  booktitle={Proceedings of the 64th Annual Meeting of the Association for Computational Linguistics (Volume 1: Long Papers)},
  pages={17417--17437},
  year={2026},
  url={https://aclanthology.org/2026.acl-long.791/}
}

@inproceedings{gunjal2025rubrics,
  title={{Rubrics as Rewards: Reinforcement Learning Beyond Verifiable Domains}},
  author={Gunjal, Anisha and Wang, Anthony and Lau, Elaine and Nath, Vaskar and He, Yunzhong and Liu, Bing and Hendryx, Sean},
  booktitle={Proceedings of the Fourteenth International Conference on Learning Representations (ICLR)},
  pages={127924--127945},
  year={2026},
  url={https://proceedings.iclr.cc/paper_files/paper/2026/hash/cfd7bee7a651ee9af525098ef67a9e45-Abstract-Conference.html}
}

@article{zheng2023judging,
  title={{Judging LLM-as-a-Judge with MT-Bench and Chatbot Arena}},
  author={Zheng, Lianmin and Chiang, Wei-Lin and Sheng, Ying and Zhuang, Siyuan and Wu, Zhanghao and Zhuang, Yonghao and Lin, Zi and Li, Zhuohan and Li, Dacheng and Xing, Eric and others},
  journal={Advances in Neural Information Processing Systems (NeurIPS)},
  pages={46595--46623},
  year={2023},
  url={https://proceedings.neurips.cc/paper_files/paper/2023/hash/91f18a1287b398d378ef22505bf41832-Abstract-Datasets_and_Benchmarks.html}
}

@article{gu2024survey,
  title={{A Survey on LLM-as-a-Judge}},
  author={Gu, Jiawei and
             Jiang, Xuhui and
             Shi, Zhichao and
             Tan, Hexiang and
             Zhai, Xuehao and
             Xu, Chengjin and
             Li, Wei and
             Shen, Yinghan and
             Ma, Shengjie and
             Liu, Honghao and
             Wang, Saizhuo and
             Zhang, Kun and
             Lin, Zhouchi and
             Zhang, Bowen and
             Ni, Lionel and
             Gao, Wen and
             Wang, Yuanzhuo and
             Guo, Jian},
  journal={The Innovation},
  volume={7},
  number={6},
  pages={101253},
  year={2026},
  url={https://doi.org/10.1016/j.xinn.2025.101253}
}

@inproceedings{li2024judgesurvey,
  title={{From Generation to Judgment: Opportunities and Challenges of LLM-as-a-Judge}},
  author={Li, Dawei and Jiang, Bohan and Huang, Liangjie and Beigi, Alimohammad and Zhao, Chengshuai and Tan, Zhen and Bhattacharjee, Amrita and Jiang, Yuxuan and Chen, Canyu and Wu, Tianhao and Shu, Kai and Cheng, Lu and Liu, Huan},
  booktitle={Proceedings of the 2025 Conference on Empirical Methods in Natural Language Processing (EMNLP)},
  year={2025},
  pages={2757--2791},
  url={https://aclanthology.org/2025.emnlp-main.138/}
}

@inproceedings{panickssery2024selfpreference,
  title={{LLM Evaluators Recognize and Favor Their Own Generations}},
  author={Panickssery, Arjun and Bowman, Samuel R. and Feng, Shi},
  booktitle={Advances in Neural Information Processing Systems (NeurIPS)},
  year={2024},
  pages = {68772--68802},
  url={https://proceedings.neurips.cc/paper_files/paper/2024/hash/7f1f0218e45f5414c79c0679633e47bc-Abstract-Conference.html}
}

@article{mcnemar1947,
  title={{Note on the Sampling Error of the Difference between Correlated Proportions or Percentages}},
  author={McNemar, Quinn},
  journal={Psychometrika},
  volume={12},
  number={2},
  pages={153--157},
  year={1947},
  doi={10.1007/BF02295996}
}

@book{krippendorff2018content,
  title={{Content Analysis: An Introduction to Its Methodology}},
  author={Krippendorff, Klaus},
  edition={4},
  year={2018},
  publisher={SAGE Publications},
  address={Thousand Oaks, CA},
  url={https://methods.sagepub.com/book/mono/content-analysis-4e/toc#_}
}

@article{fisch2024robust,
  title={{Robust Preference Optimization through Reward Model Distillation}},
  author={Fisch, Adam and Eisenstein, Jacob and Zayats, Vicky and Agarwal, Alekh and Beirami, Ahmad and Nagpal, Chirag and Shaw, Peter and Berant, Jonathan},
  journal={Transactions on Machine Learning Research},
  year={2025},
  pages={1-27},
  url={https://openreview.net/forum?id=E2zKNuwNDc}
}

@article{qi2024onlinedpo,
  title={{Online DPO: Online Direct Preference Optimization with Fast-Slow Chasing}},
  author={Qi, Biqing and Li, Pengfei and Li, Fangyuan and Gao, Junqi and Zhang, Kaiyan and Zhou, Bowen},
  journal = {arXiv preprint arXiv:2406.05534},
  year={2024},
  url={https://arxiv.org/abs/2406.05534}
}

@inproceedings{hu2024desiderata,
  title={{New Desiderata for Direct Preference Optimization}},
  author={Hu, Xiangkun and He, Tong and Wipf, David},
  booktitle={ICML 2024 Workshop on Models of Human Feedback for AI Alignment},
  year={2024},
  pages={1-22},
  url={https://openreview.net/forum?id=Fgf0iAOb22}
}

@inproceedings{ethayarajh2024kto,
  title={{KTO: Model Alignment as Prospect Theoretic Optimization}},
  author={Ethayarajh, Kawin and Xu, Winnie and Muennighoff, Niklas and Jurafsky, Dan and Kiela, Douwe},
  booktitle={Proceedings of the 41st International Conference on Machine Learning (ICML)},
  year={2024},
  pages={1-21},
  url={https://arxiv.org/abs/2402.01306}
}

@inproceedings{hu2022lora,
  title={{LoRA: Low-Rank Adaptation of Large Language Models}},
  author={Hu, Edward J. and Shen, Yelong and Wallis, Phillip and Allen-Zhu, Zeyuan and Li, Yuanzhi and Wang, Shean and Wang, Lu and Chen, Weizhu},
  booktitle={Proceedings of the Tenth International Conference on Learning Representations (ICLR)},
  year={2022},
  pages={1-13},
  url={https://openreview.net/pdf?id=nZeVKeeFYf9}
}

@inproceedings{marafioti2025smolvlm,
  title={{SmolVLM: Redefining Small and Efficient Multimodal Models}},
  author={Marafioti, Andr{\'e}s and Zohar, Orr and Farr{\'e}, Miquel and Noyan, Merve and Bakouch, Elie and Cuenca, Pedro and Zakka, Cyril and Ben Allal, Loubna and Lozhkov, Anton and Tazi, Nouamane and Srivastav, Vaibhav and Lochner, Joshua and Larcher, Hugo and Morlon, Mathieu and Tunstall, Lewis and von Werra, Leandro and Wolf, Thomas},
  booktitle={Proceedings of the Second Conference on Language Modeling (COLM)},
  year={2025},
  pages={1-19},
  url={https://openreview.net/forum?id=qMUbhGUFUb}
}

@article{bai2025qwen3vl,
  title={{Qwen3-VL Technical Report}},
  author={Bai, Shuai and Cai, Yuxuan and Chen, Ruizhe and Chen, Keqin and Chen, Xionghui and Cheng, Zesen and Deng, Lianghao and Ding, Wei and Gao, Chang and Ge, Chunjiang and Ge, Wenbin and Guo, Zhifang and Huang, Qidong and Huang, Jie and Huang, Fei and Hui, Binyuan and Jiang, Shutong and Li, Zhaohai and Li, Mingsheng and Li, Mei and Li, Kaixin and Lin, Zicheng and Lin, Junyang and Liu, Xuejing and Liu, Jiawei and Liu, Chenglong and Liu, Yang and Liu, Dayiheng and Liu, Shixuan and Lu, Dunjie and Luo, Ruilin and Lv, Chenxu and Men, Rui and Meng, Lingchen and Ren, Xuancheng and Ren, Xingzhang and Song, Sibo and Sun, Yuchong and Tang, Jun and Tu, Jianhong and Wan, Jianqiang and Wang, Peng and Wang, Pengfei and Wang, Qiuyue and Wang, Yuxuan and Xie, Tianbao and Xu, Yiheng and Xu, Haiyang and Xu, Jin and Yang, Zhibo and Yang, Mingkun and Yang, Jianxin and Yang, An and Yu, Bowen and Zhang, Fei and Zhang, Hang and Zhang, Xi and Zheng, Bo and Zhong, Humen and Zhou, Jingren and Zhou, Fan and Zhou, Jing and Zhu, Yuanzhi and Zhu, Ke},
  journal={arXiv preprint arXiv:2511.21631},
  year={2025},
  url={https://arxiv.org/abs/2511.21631}
}

@article{wang2025internvl35,
  title={{InternVL3.5: Advancing Open-Source Multimodal Models in Versatility, Reasoning, and Efficiency}},
  author={Wang, Weiyun and Gao, Zhangwei and Gu, Lixin and Pu, Hengjun and Cui, Long and Wei, Xingguang and Liu, Zhaoyang and Jing, Linglin and Ye, Shenglong and Shao, Jie and Wang, Zhaokai and Chen, Zhe and Zhang, Hongjie and Yang, Ganlin and Wang, Haomin and Wei, Qi and Yin, Jinhui and Li, Wenhao and Cui, Erfei and Chen, Guanzhou and Ding, Zichen and Tian, Changyao and Wu, Zhenyu and Xie, Jingjing and Li, Zehao and Yang, Bowen and Duan, Yuchen and Wang, Xuehui and Hou, Zhi and Hao, Haoran and Zhang, Tianyi and Li, Songze and Zhao, Xiangyu and Duan, Haodong and Deng, Nianchen and Fu, Bin and He, Yinan and Wang, Yi and He, Conghui and Shi, Botian and He, Junjun and Xiong, Yingtong and Lv, Han and Wu, Lijun and Shao, Wenqi and Zhang, Kaipeng and Deng, Huipeng and Qi, Biqing and Ge, Jiaye and Guo, Qipeng and Zhang, Wenwei and Zhang, Songyang and Cao, Maosong and Lin, Junyao and Tang, Kexian and Gao, Jianfei and Huang, Haian and Gu, Yuzhe and Lyu, Chengqi and Tang, Huanze and Wang, Rui and Lv, Haijun and Ouyang, Wanli and Wang, Limin and Dou, Min and Zhu, Xizhou and Lu, Tong and Lin, Dahua and Dai, Jifeng and Su, Weijie and Zhou, Bowen and Chen, Kai and Qiao, Yu and Wang, Wenhai and Luo, Gen},
  journal={arXiv preprint arXiv:2508.18265},
  year={2025},
  url={https://arxiv.org/abs/2508.18265}
}

@article{gemma2025gemma3,
  title={{Gemma 3 Technical Report}},
  author={Kamath, Aishwarya and
             Ferret, Johan and
             Pathak, Shreya and
             Vieillard, Nino and
             Merhej, Ramona and
             Perrin, Sarah and
             Matejovicova, Tatiana and
             Ram{\'e}, Alexandre and
             Rivi{\`e}re, Morgane and
             Rouillard, Louis and
             Mesnard, Thomas and
             Cideron, Geoffrey and
             Grill, Jean-bastien and
             Ramos, Sabela and
             Yvinec, Edouard and
             Casbon, Michelle and
             Pot, Etienne and
             Penchev, Ivo and
             Liu, Ga{\"e}l and
             Visin, Francesco and
             Kenealy, Kathleen and
             Beyer, Lucas and
             Zhai, Xiaohai and
             Tsitsulin, Anton and
             Busa-Fekete, Robert and
             Feng, Alex and
             Sachdeva, Noveen and
             Coleman, Benjamin and
             Gao, Yi and
             Mustafa, Basil and
             Barr, Iain and
             Parisotto, Emilio and
             Tian, David and
             Eyal, Matan and
             Cherry, Colin and
             Peter, Jan-Thorsten and
             Sinopalnikov, Danila and
             Bhupatiraju, Surya and
             Agarwal, Rishabh and
             Kazemi, Mehran and
             Malkin, Dan and
             Kumar, Ravin and
             Vilar, David and
             Brusilovsky, Idan and
             Luo, Jiaming and
             Steiner, Andreas and
             Friesen, Abe and
             Sharma, Abhanshu and
             Sharma, Abheesht and
             Gilady, Adi Mayrav and
             Goedeckemeyer, Adrian and
             Saade, Alaa and
             Kolesnikov, Alexander and
             Bendebury, Alexei and
             Abdagic, Alvin and
             Vadi, Amit and
             Gy{\"o}rgy, Andr{\'a}s and
             Pinto, Andr{\'e} Susano and
             Das, Anil and
             Bapna, Ankur and
             Miech, Antoine and
             Yang, Antoine and
             Paterson, Antonia and
             Shenoy, Ashish and
             Chakrabarti, Ayan and
             Piot, Bilal and
             Wu, Bo and
             Shahriari, Bobak and
             Petrini, Bryce and
             Chen, Charlie and
             Le Lan, Charline and
             Choquette-Choo, Christopher A. and
             Carey, CJ and
             Brick, Cormac and
             Deutsch, Daniel and
             Eisenbud, Danielle and
             Cattle, Dee and
             Cheng, Derek and
             Paparas, Dimitris and
             Sreepathihalli, Divyashree Shivakumar and
             Reid, Doug and
             Tran, Dustin and
             Zelle, Dustin and
             Noland, Eric and
             Huizenga, Erwin and
             Kharitonov, Eugene and
             Liu, Frederick and
             Amirkhanyan, Gagik and
             Cameron, Glenn and
             Hashemi, Hadi and
             Klimczak-Pluci{\'n}ska, Hanna and
             Singh, Harman and
             Mehta, Harsh and
             Lehri, Harshal Tushar and
             Hazimeh, Hussein and
             Ballantyne, Ian and
             Szpektor, Idan and
             Nardini, Ivan and
             Pouget-Abadie, Jean and
             Chan, Jetha and
             Stanton, Joe and
             Wieting, John and
             Lai, Jonathan and
             Orbay, Jordi and
             Fernandez, Joseph and
             Newlan, Josh and
             Ji, Ju-yeong and
             Singh, Jyotinder and
             Black, Kat and
             Yu, Kathy and
             Hui, Kevin and
             Vodrahalli, Kiran and
             Greff, Klaus and
             Qiu, Linhai and
             Valentine, Marcella and
             Coelho, Marina and
             Ritter, Marvin and
             Hoffman, Matt and
             Watson, Matthew and
             Chaturvedi, Mayank and
             Moynihan, Michael and
             Ma, Min and
             Babar, Nabila and
             Noy, Natasha and
             Byrd, Nathan and
             Roy, Nick and
             Momchev, Nikola and
             Chauhan, Nilay and
             Bunyan, Oskar and
             Botarda, Pankil and
             Caron, Paul and
             Rubenstein, Paul Kishan and
             Culliton, Phil and
             Schmid, Philipp and
             Sessa, Pier Giuseppe and
             Xu, Pingmei and
             Stanczyk, Piotr and
             Tafti, Pouya and
             Shivanna, Rakesh and
             Wu, Renjie and
             Pan, Renke and
             Rokni, Reza and
             Willoughby, Rob and
             Vallu, Rohith and
             Mullins, Ryan and
             Jerome, Sammy and
             Smoot, Sara and
             Girgin, Sertan and
             Iqbal, Shariq and
             Reddy, Shashir and
             Sheth, Shruti and
             P{\~o}der, Siim and
             Bhatnagar, Sijal and
             Panyam, Sindhu Raghuram and
             Eiger, Sivan and
             Zhang, Susan and
             Liu, Tianqi and
             Yacovone, Trevor and
             Liechty, Tyler and
             Kalra, Uday and
             Evci, Utku and
             Misra, Vedant and
             Roseberry, Vincent and
             Feinberg, Vlad and
             Kolesnikov, Vlad and
             Han, Woohyun and
             Kwon, Woosuk and
             Chen, Xi and
             Chow, Yinlam and
             Zhu, Yuvein and
             Wei, Zichuan and
             Egyed, Zoltan and
             Cotruta, Victor and
             Giang, Minh and
             Kirk, Phoebe and
             Rao, Anand and
             Lo, Jessica and
             Moreira, Erica and
             Martins, Luiz Gustavo and
             Sanseviero, Omar and
             Gonzalez, Lucas and
             Gleicher, Zach and
             Warkentin, Tris and
             Mirrokni, Vahab and
             Senter, Evan and
             Collins, Eli and
             Barral, Joelle and
             Ghahramani, Zoubin and
             Hadsell, Raia and
             Matias, Yossi and
             Sculley, D. and
             Petrov, Slav and
             Fiedel, Noah and
             Shazeer, Noam and
             Vinyals, Oriol and
             Dean, Jeff and
             Hassabis, Demis and
             Kavukcuoglu, Koray and
             Farabet, Clement and
             Buchatskaya, Elena and
             Alayrac, Jean-Baptiste and
             Anil, Rohan and
             Lepikhin, Dmitry and
             Borgeaud, Sebastian and
             Bachem, Olivier and
             Joulin, Armand and
             Andreev, Alek and
             Hardin, Cassidy and
             Dadashi, Robert and
             Hussenot, L{\'e}onard},
  journal={arXiv preprint arXiv:2503.19786},
  year={2025},
  url={https://arxiv.org/abs/2503.19786}
}

@inproceedings{vidscribe2026,
   title={{ViDscribe: Multimodal AI for Customizing Audio Description and Question Answering in Online Videos}},
   author={Cheema, Maryam S and Elahimanesh, Sina and Fazli, Pooyan and Seifi, Hasti},
   booktitle={Proceedings of the Extended Abstracts of the 2026 Conference on Human Factors in Computing Systems (CHI)},
   year={2026},
   pages={1-6},
   url={https://dl.acm.org/doi/full/10.1145/3772363.3798744},
}

@article{holm1979simple,
  title={{A Simple Sequentially Rejective Multiple Test Procedure}},
  author={Sture Holm},
  journal={Scandinavian Journal of Statistics},
  year={1979},
  volume={6},
  pages={65-70},
  url={https://www.jstor.org/stable/4615733}
}

@techreport{who2019vision,
  title={{World Report on Vision}},
  author={{World Health Organization}},
  institution={World Health Organization},
  address={Geneva},
  year={2019},
  url={https://www.who.int/publications/i/item/9789241516570}
}

@techreport{eurostudent2021,
  title={{Social and Economic Conditions of Student Life in Europe: {EUROSTUDENT} {VII} Synopsis of Indicators 2018--2021}},
  author={Hauschildt, Kristina and Gwos{\'c}, Christoph and Schirmer, Hendrik and Wartenbergh-Cras, Froukje},
  institution={German Centre for Higher Education Research and Science Studies (DZHW)},
  year={2021},
  url={https://eurostudent.eu/publications/eurostudent-vii-synopsis-of-indicators}
}

@techreport{uncrpd2006,
  title={{Convention on the Rights of Persons with Disabilities}},
  author={{United Nations}},
  howpublished={Treaty Series, vol.\ 2515, p.\ 3; adopted 13 December 2006, entered into force 3 May 2008},
  year={2006},
  url={https://www.un.org/development/desa/disabilities/convention-on-the-rights-of-persons-with-disabilities.html}
}

@techreport{un2015sdg,
  title={{Transforming Our World: The 2030 Agenda for Sustainable Development}},
  author={{United Nations General Assembly}},
  howpublished={Resolution A/RES/70/1, adopted 25 September 2015},
  year={2015},
  url={https://sdgs.un.org/2030agenda}
}

@inproceedings{radford2023whisper,
  title={{Robust Speech Recognition via Large-Scale Weak Supervision}},
  author={Radford, Alec and Kim, Jong Wook and Xu, Tao and Brockman, Greg and Mcleavey, Christine and Sutskever, Ilya},
  booktitle={Proceedings of the 40th International Conference on Machine Learning (ICML)},
  pages={28492--28518},
  year={2023},
  url = 	 {https://proceedings.mlr.press/v202/radford23a.html},
}

@misc{silero2024vad,
  title={{Silero VAD: Pre-trained Enterprise-Grade Voice Activity Detector}},
  author={{Silero Team}},
  year={2024},
  howpublished={Software, GitHub repository},
  url= {https://github.com/snakers4/silero-vad}
}

@misc{ffmpeg,
  title={{FFmpeg: A Complete, Cross-Platform Solution to Record, Convert and Stream Audio and Video}},
  author={{FFmpeg Developers}},
  howpublished={Software},
  url={https://ffmpeg.org/},
  note={Accessed September 2026}
}

@misc{jaided2020easyocr,
  title={{EasyOCR: Ready-to-Use OCR with 80+ Supported Languages}},
  author={{Jaided AI}},
  year={2020},
  howpublished={Software, GitHub repository},
  url={https://github.com/JaidedAI/EasyOCR}
}

@misc{google2026gemini31,
  title={{Gemini 3.1 Pro} Model Card},
  author={{Google DeepMind}},
  year={2026},
  howpublished = {\url{https://storage.googleapis.com/deepmind-media/Model-Cards/Gemini-3-1-Pro-Model-Card.pdf}},
  note={Accessed September 2026}
}

@misc{openai2026gpt55,
  title={{Introducing GPT-5.5}},
  author={{OpenAI}},
  year={2026},
  howpublished={\url{https://openai.com/index/introducing-gpt-5-5/}},
  note={Accessed September 2026}
}

@misc{anthropic2026sonnet46,
  title={{Introducing Claude Sonnet 4.6}},
  author={{Anthropic}},
  year={2026},
  howpublished = {\url{https://www.anthropic.com/system-cards}},
  note={Accessed September 2026}
}

@inproceedings{fischer2025swissadt,
    title={{SwissADT: An Audio Description Translation System for Swiss Languages}},
    author={Fischer, Lukas  and
      Gao, Yingqiang  and
      Lintner, Alexa  and
      Rios, Annette  and
      Ebling, Sarah},
    booktitle={Proceedings of the 2025 Conference of the Nations of the Americas Chapter of the Association for Computational Linguistics: Human Language Technologies (Volume 3: Industry Track)},
    year={2025},
    pages={370--379},
    url={https://aclanthology.org/2025.naacl-industry.31/}

}

\appendix

\section{The 45-Rule Lecture-AD Standard}
\label{app:rules}
The complete 45-rule standard for audio description of slide-based university lectures, reproduced from the standard's source file and numbered by the published rule index, the same numbering the per-rule compliance prompts of Appendix~\ref{app:prompts} and the per-rule analyses use, so that a rule keeps its number when the standard gains one. Each entry closes with its category, whether it is documented (DOC, grounded in a cited source) or provisional (PROV, with a recorded derivation basis), and its identifier in the standard's source file, which also records each rule's citation, scoring route, and, for the 13 rules reviewed by a BLV collaborator, the review verdict.


\noindent\rulebadge{\textbf{Rule \#1.}}
REDUNDANCY GATE (governs every 'describe' rule in \S{}2): describe on-screen content only to the extent the lecturer's speech does not already convey it. The element rules in \S{}2 specify HOW to describe a figure/table/chart/equation/etc.; this rule decides WHETHER to, and how much --- always only the gap the voice leaves. When the lecturer's words already carry the content, suppress entirely. This is a system for covering what speech omits, not for reading the screen aloud. \textit{[cross-cutting; DOC; \texttt{cross\_cutting\_001}]}

\noindent\rulebadge{\textbf{Rule \#2.}}
Use present tense, active voice, third-person narration. Refer to the pointer as the cursor or pointer, or to the person as the lecturer; never a gendered pronoun (he/she) -- the listener cannot see who is speaking. \textit{[cross-cutting; DOC; \texttt{cross\_cutting\_002}]}

\noindent\rulebadge{\textbf{Rule \#3.}}
Describe observable visual content; do not infer emotions, intent, or significance. \textit{[cross-cutting; DOC; \texttt{cross\_cutting\_003}]}

\noindent\rulebadge{\textbf{Rule \#4.}}
Use consistent terminology within a lecture and across a course; adopt the lecturer's wording when terms are introduced. \textit{[cross-cutting; DOC; \texttt{cross\_cutting\_004}]}

\noindent\rulebadge{\textbf{Rule \#5.}}
When the AD system's confidence in what is on screen is low, prefer silence over a confident wrong description. A BLV student cannot verify a description; a wrong one is worse than silence. \textit{[cross-cutting; DOC; \texttt{cross\_cutting\_005}]}

\noindent\rulebadge{\textbf{Rule \#6.}}
Do not re-describe visual content already described earlier in the same lecture. After first describing an element (a diagram, a definition, a figure), subsequent appearances within the same lecture get either a name-only reference or silence, not a full re-description. \textit{[cross-cutting; DOC; \texttt{cross\_cutting\_006}]}

\noindent\rulebadge{\textbf{Rule \#7.}}
Lookahead redundancy (extends the redundancy gate cross\_cutting\_001 across time): suppress a description if the lecturer covers the same content shortly AFTER the visual event, not only simultaneously. OFFLINE dataset generation: look ahead over the transcript and suppress the description if the lecturer verbalizes the content within a window LOCKED to: before the next pedagogical event, or 45 seconds, whichever comes first. LIVE deployment: hold a non-urgent description for a brief grace window; if the lecturer begins covering it within the window, suppress, otherwise describe. Prevents the AD from pre-empting an explanation the lecturer is about to give (the covered-later case for reveals, equations, tables, figures). \textit{[cross-cutting; PROV; \texttt{cross\_cutting\_007}]}

\noindent\rulebadge{\textbf{Rule \#8.}}
Describing the lecturer's gestures and board drawings --- and resolving the deixis attached to them --- is a core (not peripheral) lecture-AD task. The WCAG 1.2.7 physics-lecture example is the canonical case: a professor sketches rapidly on a whiteboard while gesturing and pauses are insufficient --- the case proving gap-insertion alone fails. In a live system this is handled by the fallback ladder (length\_003), not by pausing (WCAG 1.2.7 extended AD is prerecorded-only). The per-element machinery lives in \S{}2.Deixis (rules deixis\_001..006); this rule frames it as a first-class concern of lecture AD. \textit{[cross-cutting; DOC; \texttt{cross\_cutting\_008}]}

\noindent\rulebadge{\textbf{Rule \#9.}}
Gap-insertion is the base delivery mechanism: insert AD into existing pauses in the lecturer's speech. \textit{[length and delivery; DOC; \texttt{length\_001}]}

\noindent\rulebadge{\textbf{Rule \#10.}}
Live mode prioritization: deixis-resolution \textgreater{} slide-text gist \textgreater{} figure summary \textgreater{} full figure. \textit{[length and delivery; PROV; \texttt{length\_002}]}

\noindent\rulebadge{\textbf{Rule \#11.}}
Live fallback ladder (the live replacement for extended AD, which is unavailable because the live stream cannot be paused for the room). The BLV listener's audio is a per-user track, so it may run behind the live feed without affecting anyone else. When a describable visual cannot be conveyed in the available speech gap, descend until one rung succeeds: (1) full description if it fits the gap; (2) compressed description if a shorter gap fits; (3) time-shift --- let the listener's track fall behind the live feed to deliver the full description, then catch up during the next pause (or by briefly faster playback), up to a maximum lag cap L (team-chosen, default \textasciitilde{}20s); (4) honest placeholder naming what is on screen without faking coverage (e.g. 'equation on screen, long derivation in progress') when time-shift cannot keep up --- lag would exceed L, or no upcoming pause exists to catch up in; (5) drop, only as a last resort. Collision-losers (two events competing for one gap) are out of scope for this standard. This ladder is the live replacement for every former per-element 'default to extended AD' instruction (figures, typeset math, ink writing, handwritten derivations). Tables and code keep their own element-specific compression and reference this ladder only for overflow. \textit{[length and delivery; PROV; \texttt{length\_003}]}

\noindent\rulebadge{\textbf{Rule \#12.}}
Concise insertion default = 25 words; detailed override = 100 words. \textit{[length and delivery; DOC; \texttt{length\_004}]}

\noindent\rulebadge{\textbf{Rule \#13.}}
Avoid continuous description; leave silence on most lecturer-narrated stretches. \textit{[length and delivery; PROV; \texttt{length\_005}]}

\noindent\rulebadge{\textbf{Rule \#14.}}
Read on-screen text central to comprehension that is not already conveyed by the lecturer's speech. This covers the slide title and any individual body item: surface only the specific uncovered item, gated by cross\_cutting\_001 --- never read the slide top-to-bottom. \textit{[slide text; DOC; \texttt{slide\_text\_001}]}

\noindent\rulebadge{\textbf{Rule \#15.}}
Do not read aloud slide text the lecturer is reading aloud (redundancy). \textit{[slide text; DOC; \texttt{slide\_text\_002}]}

\noindent\rulebadge{\textbf{Rule \#16.}}
Do not describe transition animations (dissolves, wipes, slide-in). \textit{[slide transition; PROV; \texttt{slide\_transition\_001}]}

\noindent\rulebadge{\textbf{Rule \#17.}}
Do not describe animation effects (fly-in, fade-up). \textit{[slide animation; PROV; \texttt{slide\_animation\_001}]}

\noindent\rulebadge{\textbf{Rule \#18.}}
Apply the redundancy gate to each revealed increment independently: when a build adds a bullet, equation step, or chart series, describe only the delta the new step adds, and only if speech leaves it uncovered --- never re-read the accumulated slide state. Stepwise typeset-equation builds follow the same delta convention; chart build animations are described as the completed data state, not frame by frame. \textit{[slide animation; PROV; \texttt{slide\_animation\_002}]}

\noindent\rulebadge{\textbf{Rule \#19.}}
Drill down: brief summary first; then main components; then details only if a gap remains. Describe only the part the lecturer's speech does not already convey (redundancy gate, cross\_cutting\_001). \textit{[figure; DOC; \texttt{figure\_001}]}

\noindent\rulebadge{\textbf{Rule \#20.}}
Foreground data-carrying elements; suppress decorative styling. \textit{[figure; DOC; \texttt{figure\_002}]}

\noindent\rulebadge{\textbf{Rule \#21.}}
Do not describe decorative elements (ornamental backgrounds, branded logos, semantically inert color). \textit{[figure; DOC; \texttt{figure\_003}]}

\noindent\rulebadge{\textbf{Rule \#22.}}
Open with single-sentence type-and-purpose ('A block diagram of \ldots{}'), then a brief list of components in reading order. \textit{[figure; DOC; \texttt{figure\_004}]}

\noindent\rulebadge{\textbf{Rule \#23.}}
Provide brief summary before reading the table (what it shows, what the dimensions are). Describe only the part the lecturer's speech does not already convey (redundancy gate, cross\_cutting\_001). \textit{[table; DOC; \texttt{table\_001}]}

\noindent\rulebadge{\textbf{Rule \#24.}}
Split multi-layered tables into smaller sub-tables organized by category. \textit{[table; DOC; \texttt{table\_002}]}

\noindent\rulebadge{\textbf{Rule \#25.}}
Do not describe cell-level styling unless it encodes semantic meaning. \textit{[table; DOC; \texttt{table\_003}]}

\noindent\rulebadge{\textbf{Rule \#26.}}
Open with chart type and what it measures, then a one-sentence trend, then data structure. Describe only the part the lecturer's speech does not already convey (redundancy gate, cross\_cutting\_001). \textit{[chart; DOC; \texttt{chart\_001}]}

\noindent\rulebadge{\textbf{Rule \#27.}}
Where data is sparse (4 or fewer values), present as a short text list rather than a full table. \textit{[chart; DOC; \texttt{chart\_002}]}

\noindent\rulebadge{\textbf{Rule \#28.}}
Do not describe bar/line color unless colours encode categories the listener must distinguish. \textit{[chart; DOC; \texttt{chart\_003}]}

\noindent\rulebadge{\textbf{Rule \#29.}}
Surface the equation aurally; mark it semantically so a renderer can verbalize per user preference. Describe only the part the lecturer's speech does not already convey (redundancy gate, cross\_cutting\_001). \textit{[typeset mathematics; DOC; \texttt{math\_typeset\_001}]}

\noindent\rulebadge{\textbf{Rule \#30.}}
Pair every non-trivial equation with a one-line natural-language gloss in the same insertion. \textit{[typeset mathematics; PROV; \texttt{math\_typeset\_002}]}

\noindent\rulebadge{\textbf{Rule \#31.}}
Use ClearSpeak as the default verbalization grammar, applied consistently within a course; defer to learner preference where the learner has one. (ClearSpeak chosen over MathSpeak because it reads closer to natural spoken mathematics; no universal standard exists.) \textit{[typeset mathematics; PROV; \texttt{math\_typeset\_003}]}

\noindent\rulebadge{\textbf{Rule \#32.}}
Announce language and purpose ('Python snippet, five lines, defines a binary-search function'). Describe only the part the lecturer's speech does not already convey (redundancy gate, cross\_cutting\_001). \textit{[code; PROV; \texttt{code\_001}]}

\noindent\rulebadge{\textbf{Rule \#33.}}
Surface error or output highlights when the lecturer references them deictically. \textit{[code; PROV; \texttt{code\_002}]}

\noindent\rulebadge{\textbf{Rule \#34.}}
Do not re-read code the lecturer has read aloud line-by-line (redundancy). \textit{[code; DOC; \texttt{code\_003}]}

\noindent\rulebadge{\textbf{Rule \#35.}}
Surface the referent ('the kinetic-energy term, $\tfrac{1}{2}$mv\textsuperscript{2}'), not the act of pointing. \textit{[deixis; PROV; \texttt{deixis\_001}]}

\noindent\rulebadge{\textbf{Rule \#36.}}
When referent cannot be resolved with confidence, prefer silence to a wrong guess. \textit{[deixis; DOC; \texttt{deixis\_002}]}

\noindent\rulebadge{\textbf{Rule \#37.}}
Do not describe idle cursor motion uncoupled from speech. \textit{[deixis; PROV; \texttt{deixis\_003}]}

\noindent\rulebadge{\textbf{Rule \#38.}}
Do not describe the pointing act as the description ('the cursor is pointing at it'). \textit{[deixis; PROV; \texttt{deixis\_004}]}

\noindent\rulebadge{\textbf{Rule \#39.}}
Resolve bare deictics ('this', 'here', 'this term') from the cursor's dwell anchor at utterance time: sustained dwell or deliberate circling over a slide region counts as a point; transit motion between regions is non-deictic (deixis\_003). In this scope the cursor --- with highlights and ink --- is the only pointing modality, so an unanchored bare deictic resolves to silence (deixis\_002). \textit{[deixis; PROV; \texttt{deixis\_005}]}

\noindent\rulebadge{\textbf{Rule \#40.}}
Treat highlight boxes, callout shapes, and colour emphasis drawn over a slide as deictic acts: surface the highlighted referent ('the boundary condition, u(0)=0'), not the overlay's appearance --- describe colour, shape, or stroke only when it encodes meaning the lecturer relies on. \textit{[deixis; PROV; \texttt{deixis\_006}]}

\noindent\rulebadge{\textbf{Rule \#41.}}
Do not describe the lecturer's face thumbnail --- its presence, position, facial expressions, clothing, background, or appearance/disappearance; in this scope it is decoration and the speaker is already carried by the voice. If the lecturer explicitly directs attention to the webcam view (e.g. holds an object to the camera), treat it as an off-slide surface and describe it as such. \textit{[edge cases; PROV; \texttt{edge\_cases\_001}]}

\noindent\rulebadge{\textbf{Rule \#42.}}
When the lecturer zooms into, scrolls within, or otherwise reframes the shared slide, describe the content the new framing presents ('the lower-right quadrant of the diagram fills the screen: ...'), not the navigation mechanics; treat a sustained close-up on a region as deictic emphasis and resolve it per \S{}2.Deixis. Name the region now shown rather than the camera-like move. \textit{[edge cases; PROV; \texttt{edge\_cases\_002}]}

\noindent\rulebadge{\textbf{Rule \#43.}}
VISUAL-CONTENT CONDITION (the second gate beside the redundancy gate): describe a moment only when a visual referent's FORM is needed to follow the lecturer's point and is not conveyed by speech. Form means one of: which element is meant, the shape or trend of a curve, the spatial arrangement of parts, relative magnitude, a count, a value read off the figure, or a colour code. The concept being spoken does not make the visual redundant; the slide existing does not make it describable; text has no form to describe. A moment with no visual referent, or whose form is already spoken, is not described. \textit{[visual content; PROV; \texttt{visual\_content\_001}]}

\noindent\rulebadge{\textbf{Rule \#44.}}
On a slide whose only content is text (title, bullets, prose), the AD is at most one orientation line naming the slide's topic, given once when the slide first appears; never a paraphrase or summary of the bullets. If the lecturer reads or paraphrases the text, say nothing (slide\_text\_002). \textit{[slide text; PROV; \texttt{slide\_text\_003}]}

\noindent\rulebadge{\textbf{Rule \#45.}}
A constructed defect in a preference pair must be perceptible to a listener: checkable against the speech, the named visual referent, or a stated form property. A perturbation of a slide-text token on a text-only moment (e.g. 'ten-fiber' inserted into a bullet paraphrase) is imperceptible by design and the pair is uninformative. Length defects must be audible overruns (the longer side at least 8 words and 1.4x the shorter, and over the rung budget), not marginal ones. \textit{[pair perceptibility; PROV; \texttt{pair\_perceptibility\_001}]}

\section{Prompts}
\label{app:prompts}

We list every prompt in this study, one subsection per prompt. Placeholders in braces are filled at run time, and typographic characters are rendered in ASCII (em and en dashes as two hyphens, the plus--minus sign as +/-, and the unit-separator token as [US]); nothing else is altered. Each lead-in states where the prompt was employed. A transcript-window span named inside a prompt, such as the $\pm 8$\,s of Prompt~\ref{app:prompt-p1}, is that prompt's own window; the two windows used for reference generation are given in Appendix~\ref{app:training}. Not every prompt listed here feeds a reported figure: those of Prompts~\ref{app:prompt-rejudge-text}--\ref{app:prompt-valdim}, \ref{app:prompt-rank}, \ref{app:prompt-compliance} and \ref{app:prompt-cert}--\ref{app:prompt-cqa} were used in development, pilot or auxiliary analyses and produce no number reported in the paper, and every natural preference pair in this work was ordered by the judge of Prompt~\ref{app:prompt-j1}. Two de-duplications are applied: the 42 rendered per-rule compliance prompts instantiate one template, shown once, with Appendix~\ref{app:training} saying why three of the 45 rules carry none; and the standard spliced into the shared prefix is Appendix~\ref{app:rules} and is not repeated.

\subsection{Shared standard prefix (prompt set, \S 0)}
\label{app:prompt-prefix}
The following prefix was prepended, byte-identically, to the reference-generation (P2), candidate-generation (P3), controlled-pair (P4), and judge (J1, J2) prompts of Section~\ref{sec:standard}. The line \verb|<<<rule YAML>>>| marks where the full standard of Appendix~\ref{app:rules} is spliced in verbatim; it is not repeated here.
\begin{promptbox}{Prompt B.1: Shared standard prefix (prompt set, \S 0)}
\begin{lstlisting}[style=promptbox]
You are an audio description (AD) author for blind and low-vision (BLV) students watching
screencast-style lecture recordings.

[rules_for_slides.yaml -- 45 rules, verbatim]
<<< rules YAML >>>

OPERATING PRINCIPLES

REDUNDANCY GATE (cross_cutting_001) is the spine: describe on-screen content only to the
extent the lecturer's speech does not already convey it. Every coverage claim must be
evidenced from the transcript window itself -- quote the covering span verbatim when you
claim coverage, and name the specific uncovered visual content when you claim none. Judge
each moment on its own evidence.

KEYFRAMES are the authoritative source of visual content. OCR and transcript are aids; where
they conflict with the keyframes, trust the keyframes.

DEIXIS (deixis_001, deixis_004): surface the referent -- what is pointed at, with its on-slide
identity -- never the act of pointing. "The kinetic-energy term, one half m v squared." Never
"the lecturer points at an equation."

UNCERTAINTY (cross_cutting_005): where you cannot determine what is on screen, prefer silence
to a confident wrong description. A BLV student cannot verify.

STYLE (cross_cutting_002/003): present tense, active voice, third person, observable content
only. FORBIDDEN: "next slide", "on the slide there is", "the slide shows", "the image shows",
"the figure shows", "the screen shows", "an image of", "a picture of", "this slide", "we can
see", "is pointing", "the presenter", "now the lecturer", slide numbers, template chrome,
footers, the webcam thumbnail, transition animations. Describe the CONTENT directly.

MATH (math_typeset_003): ClearSpeak verbalization, consistent within a lecture.

NO RE-DESCRIPTION (cross_cutting_006): if described_so_far lists an element, use a name-only
reference or stay silent.

DELIVERY RUNGS (length_003): 1 = in-gap; 2 = compressed, max 15 words; 3 = time-shifted,
lag cap 20 s; 4 = placeholder, max 5 words. Rung 5 (drop) is never a valid output -- prefer
emit=false. Report the rung in the `rung` field. Do NOT mark it in the description text: the
AD is what a student hears, and the delivery schedule is not part of it.
\end{lstlisting}
\end{promptbox}

\subsection{Moment classification (P1)}
\label{app:prompt-p1}
The following prompt was employed by the VLM classifier that assigns each detector candidate a moment type or rejects it (Section~\ref{sec:standard}). It is rendered here with the full type list; the classifier receives ten candidates per call.
\begin{promptbox}{Prompt B.2: Moment classification (P1)}
\begin{lstlisting}[style=promptbox]
You are classifying candidate events detected in a slide-based lecture recording. A computer-
vision detector proposed each candidate; your task is to say what each one is, or to reject it.

You are NOT deciding whether the event deserves description, and NOT considering whether the
lecturer's speech covers it. Those are separate later judgements.

TYPES - use exactly one:
  slide       a new slide replaces the previous one
  transition  an animated change between slide states
  reveal      an increment appears on the existing slide (build step, bullet, overlay)
  pointing    the cursor, laser or highlight comes to rest on, dwells on, or traces a
              specific on-slide element
  ink         handwritten annotation is drawn on the slide
  figure      a diagram, image or illustration first appears
  chart       a plot or graph first appears
  table       a table first appears
  math        typeset mathematics first appears
  code        a code block first appears
  animation   embedded animated content plays (video, 3D model, simulation)
  other       a change on the slide surface fitting none of the above
  reject      not a change on the slide surface, or an artifact

Classify `pointing` only where the cursor comes to rest, dwells, or traces. Cursor transit
across the screen is not pointing. Reject: webcam-thumbnail movement, slide numbers, footers,
template chrome, compression artifacts, and returns to a slide already shown with no new
content (mark those `slide` with revisit=true).

Apply a strict standard: assign a type only where it is clearly identifiable from the frames.
Where the frames do not show an event, say so in what_on_screen and set type to reject. Do not
infer an event from the detector channel: the channel says a keyframe changed, not what changed,
and the change may predate these frames.

Compare each CONSECUTIVE PAIR of frames, not the set as a whole. A change between the last two
frames counts exactly as much as one between the first two: the event often sits near the end of
the window, so several identical frames followed by one different frame IS a change, not a static
slide.

"Nothing changes" and "the same content is moving" are different. An embedded video, a rotating
3D model or a playing simulation shows the SAME elements in a DIFFERENT position or state across
the frames -- that is `animation`, not reject, even when consecutive frames differ only slightly.
Reject requires the slide surface to be genuinely static.

When a revealed increment is itself a typed content element, use the content type rather than
`reveal`: an equation appearing as a build step is `math`, a table appearing is `table`. The type
records what the element IS, which is what determines the rules that apply to it; how it arrived
is secondary.

INPUT -- for each candidate:
  id, frames (3-5, each carrying a burned-in t = MM:SS.SS overlay), the detector channel
  that proposed it, and the transcript within +/-8 s.

Read the timestamp off the overlay. Do not estimate it.

OUTPUT -- one JSON object per candidate, in input order, no preamble, no fences:
{"id": "<id>", "t": "<MM:SS from the overlay>", "type": "<type>", "revisit": <bool>,
 "what_on_screen": "<what is visible or what changed, one sentence, observable only>"}

what_on_screen must describe only what these frames show. If nothing changes across them, write
that, and set type to reject.
\end{lstlisting}
\end{promptbox}

\subsection{Reference generation (P2)}
\label{app:prompt-p2}

The following prompt was employed to generate reference descriptions using the three teacher models introduced in Section~\ref{sec:standard}. The same prompt template was applied across all models to ensure consistency in the generation procedure, with placeholders enclosed in braces replaced by the corresponding moment-specific information. 
\begin{promptbox}{Prompt B.3: Reference generation (P2)}
\begin{lstlisting}[style=promptbox]
MOMENT
- type: {type}
- transcript_window (-5 s / +8 s, word timestamps): {transcript_window}
- slide_ocr: {slide_ocr}
- pause_after: {pause_after} s
- reachable_rung: {reachable_rung}          # computed in code; do not recompute
- described_so_far (this lecture): {described_so_far}
- keyframes: [3-5 attached]

TASK

Step 1 -- coverage. From the transcript window and keyframes alone:
  coverage = FULLY_COVERED if the lecturer's words in the window already name or convey the
             visual content of this moment. Put the covering span in covering_quote, verbatim.
  coverage = PARTLY_COVERED if some but not all is spoken. State what is left in
             uncovered_content.
  coverage = UNCOVERED if the visual carries content the window does not convey. State it in
             uncovered_content.
  An unresolved deictic -- the lecturer says "this" or "here" and the window never names the
  referent -- is UNCOVERED.

Step 2 -- emit. emit=false if coverage == FULLY_COVERED, or the moment carries no pedagogical
content, or your confidence in what is on screen is genuinely low. Otherwise emit=true.

Step 3 -- if emit=true, write the AD for exactly the uncovered content, at the word budget for
reachable_rung: rung 1 -> whatever fits the pause, default 25 words; rung 2 -> max 15;
rung 3 -> up to 25, or up to 60 only for a complex multi-element visual that is UNCOVERED;
rung 4 -> max 5. Put the rung in the `rung` field only -- no marker in the AD text.

Step 4 -- rationale: one sentence citing the driving rule IDs.

Output ONLY this JSON, no preamble, no fences:
{{"coverage": "FULLY_COVERED"|"PARTLY_COVERED"|"UNCOVERED",
 "covering_quote": <string or null>,
 "uncovered_content": <string or null>,
 "emit": <bool>,
 "ad_text": <string or null>,
 "rung": <1-4 or null>,
 "rationale": <string>}}
\end{lstlisting}
\end{promptbox}

\subsection{Reference generation, 45\,s look-ahead variant (P2-W45)}
\label{app:prompt-p2w45}
The following variant of P2 widens the transcript window used by the redundancy rule to 45 seconds after the moment; it produced the repair-run references described in the standard.
\begin{promptbox}{Prompt B.4: Reference generation, 45\,s look-ahead variant (P2-W45)}
\begin{lstlisting}[style=promptbox]
MOMENT
- type: {type}
- transcript_window (-5 s, forward to the next pedagogical event or +45 s, whichever comes first; word timestamps): {transcript_window}
- slide_ocr: {slide_ocr}
- pause_after: {pause_after} s
- reachable_rung: {reachable_rung}          # computed in code; do not recompute
- described_so_far (this lecture): {described_so_far}
- keyframes: [3-5 attached]

TASK

Step 1 -- coverage. From the transcript window and keyframes alone:
  coverage = FULLY_COVERED if the lecturer's words in the window already name or convey the
             visual content of this moment. Put the covering span in covering_quote, verbatim.
  coverage = PARTLY_COVERED if some but not all is spoken. State what is left in
             uncovered_content.
  coverage = UNCOVERED if the visual carries content the window does not convey. State it in
             uncovered_content.
  An unresolved deictic -- the lecturer says "this" or "here" and the window never names the
  referent -- is UNCOVERED.

Step 2 -- emit. emit=false if coverage == FULLY_COVERED, or the moment carries no pedagogical
content, or your confidence in what is on screen is genuinely low. Otherwise emit=true.

Step 3 -- if emit=true, write the AD for exactly the uncovered content, at the word budget for
reachable_rung: rung 1 -> whatever fits the pause, default 25 words; rung 2 -> max 15;
rung 3 -> up to 25, or up to 60 only for a complex multi-element visual that is UNCOVERED;
rung 4 -> max 5. Put the rung in the `rung` field only -- no marker in the AD text.

Step 4 -- rationale: one sentence citing the driving rule IDs.

Output ONLY this JSON, no preamble, no fences:
{{"coverage": "FULLY_COVERED"|"PARTLY_COVERED"|"UNCOVERED",
 "covering_quote": <string or null>,
 "uncovered_content": <string or null>,
 "emit": <bool>,
 "ad_text": <string or null>,
 "rung": <1-4 or null>,
 "rationale": <string>}}
\end{lstlisting}
\end{promptbox}

\subsection{Reference generation, decision-only variant (P2-DECISION)}
\label{app:prompt-p2dec}
The following variant asks only for the describe-or-stay-silent decision and was used for the decision re-draws that set the flip-down rule.
\begin{promptbox}{Prompt B.5: Reference generation, decision-only variant (P2-DECISION)}
\begin{lstlisting}[style=promptbox]
MOMENT
- type: {type}
- transcript_window (-5 s, forward to the next pedagogical event or +45 s, whichever comes first; word timestamps): {transcript_window}
- slide_ocr: {slide_ocr}
- pause_after: {pause_after} s
- reachable_rung: {reachable_rung}          # computed in code; do not recompute
- described_so_far (this lecture): {described_so_far}
- keyframes: [3-5 attached]

TASK

Step 1 -- coverage. From the transcript window and keyframes alone:
  coverage = FULLY_COVERED if the lecturer's words in the window already name or convey the
             visual content of this moment. Put the covering span in covering_quote, verbatim.
  coverage = PARTLY_COVERED if some but not all is spoken. State what is left in
             uncovered_content.
  coverage = UNCOVERED if the visual carries content the window does not convey. State it in
             uncovered_content.
  An unresolved deictic -- the lecturer says "this" or "here" and the window never names the
  referent -- is UNCOVERED.

Step 2 -- emit. emit=false if coverage == FULLY_COVERED, or the moment carries no pedagogical
content, or your confidence in what is on screen is genuinely low. Otherwise emit=true.

Step 3 -- there is no step 3. Do NOT write an AD.

Output ONLY this JSON, no preamble, no fences:
{{"coverage": "FULLY_COVERED"|"PARTLY_COVERED"|"UNCOVERED",
 "covering_quote": <string or null>,
 "uncovered_content": <string or null>,
 "emit": <bool>}}
\end{lstlisting}
\end{promptbox}

\subsection{Reference generation, decision-only control (P2-DECISION-F8)}
\label{app:prompt-p2decf8}
The following variant is the instrument control for P2-DECISION, with the fixed transcript window.
\begin{promptbox}{Prompt B.6: Reference generation, decision-only control (P2-DECISION-F8)}
\begin{lstlisting}[style=promptbox]
MOMENT
- type: {type}
- transcript_window (-5 s / +8 s, word timestamps): {transcript_window}
- slide_ocr: {slide_ocr}
- pause_after: {pause_after} s
- reachable_rung: {reachable_rung}          # computed in code; do not recompute
- described_so_far (this lecture): {described_so_far}
- keyframes: [3-5 attached]

TASK

Step 1 -- coverage. From the transcript window and keyframes alone:
  coverage = FULLY_COVERED if the lecturer's words in the window already name or convey the
             visual content of this moment. Put the covering span in covering_quote, verbatim.
  coverage = PARTLY_COVERED if some but not all is spoken. State what is left in
             uncovered_content.
  coverage = UNCOVERED if the visual carries content the window does not convey. State it in
             uncovered_content.
  An unresolved deictic -- the lecturer says "this" or "here" and the window never names the
  referent -- is UNCOVERED.

Step 2 -- emit. emit=false if coverage == FULLY_COVERED, or the moment carries no pedagogical
content, or your confidence in what is on screen is genuinely low. Otherwise emit=true.

Step 3 -- there is no step 3. Do NOT write an AD.

Output ONLY this JSON, no preamble, no fences:
{{"coverage": "FULLY_COVERED"|"PARTLY_COVERED"|"UNCOVERED",
 "covering_quote": <string or null>,
 "uncovered_content": <string or null>,
 "emit": <bool>}}
\end{lstlisting}
\end{promptbox}

\subsection{Reference generation, verifiable variant (P2-V)}
\label{app:prompt-p2v}
The following variant of P2 additionally requires the quoted evidence that a description relies on, used for the quote-verification pass.
\begin{promptbox}{Prompt B.7: Reference generation, verifiable variant (P2-V)}
\begin{lstlisting}[style=promptbox]
MOMENT
- type: {type}
- transcript_window (-5 s, forward to the next pedagogical event or +45 s, whichever comes first; word timestamps): {transcript_window}
- slide_ocr: {slide_ocr}
- pause_after: {pause_after} s
- reachable_rung: {reachable_rung}          # computed in code; do not recompute
- described_so_far (this lecture): {described_so_far}
- keyframes: [3-5 attached]

TASK

Step 1 -- coverage. From the transcript window and keyframes alone:
  coverage = FULLY_COVERED if the lecturer's words in the window already name or convey the
             visual content of this moment. Put the covering span in covering_quote, verbatim.
  coverage = PARTLY_COVERED if some but not all is spoken. State what is left in
             uncovered_content.
  coverage = UNCOVERED if the visual carries content the window does not convey. State it in
             uncovered_content.
  An unresolved deictic -- the lecturer says "this" or "here" and the window never names the
  referent -- is UNCOVERED.

Step 1b -- visual content (visual_content_001, slide_text_003). From the keyframes:
  content = VISUAL if a non-text element is on screen (figure, diagram, chart, table, equation,
            code, image, animation) or the lecturer's cursor, laser or hand indicates a specific
            element. Name that element in visual_referent. Then set unspoken_form to the ONE
            property of it that the window does not convey and a listener needs to follow the
            point -- which_element, shape_trend, spatial_arrangement, relative_magnitude, count,
            value, colour_code -- or null if the speech already conveys its form. Hearing the
            concept is not hearing the form; the slide existing does not make it describable.
  content = TEXT_ONLY if the slide surface carries only text (title, bullets, prose). Text has no
            form to describe; a paraphrase of it is not audio description.
  content = NONE if nothing on screen carries content (blank, transition frame, webcam only).
  visual_referent and unspoken_form are null unless content == VISUAL.

Step 2 -- emit. emit=true ONLY if coverage != FULLY_COVERED and either
  (a) content == VISUAL and unspoken_form is not null, or
  (b) content == TEXT_ONLY and coverage == UNCOVERED -- then Step 3 writes at most one
      orientation line naming the slide's topic (slide_text_003), never a paraphrase of the
      bullets, and only if the topic has not been given already.
Otherwise emit=false. If your confidence in what is on screen is genuinely low, emit=false.

Step 3 -- if emit=true, write the AD for exactly the uncovered content, at the word budget for
reachable_rung: rung 1 -> whatever fits the pause, default 25 words; rung 2 -> max 15;
rung 3 -> up to 25, or up to 60 only for a complex multi-element visual that is UNCOVERED;
rung 4 -> max 5. Put the rung in the `rung` field only -- no marker in the AD text.

Step 4 -- rationale: one sentence citing the driving rule IDs.

Output ONLY this JSON, no preamble, no fences:
{{"coverage": "FULLY_COVERED"|"PARTLY_COVERED"|"UNCOVERED",
 "covering_quote": <string or null>,
 "uncovered_content": <string or null>,
 "content": "NONE"|"TEXT_ONLY"|"VISUAL",
 "visual_referent": <string or null>,
 "unspoken_form": "which_element"|"shape_trend"|"spatial_arrangement"|"relative_magnitude"|"count"|"value"|"colour_code"|null,
 "emit": <bool>,
 "ad_text": <string or null>,
 "rung": <1-4 or null>,
 "rationale": <string>}}
\end{lstlisting}
\end{promptbox}

\subsection{Reference generation, verifiable decision-only variant (P2-DECISION-V)}
\label{app:prompt-p2decv}
The following variant combines the decision-only and quote-verification forms.
\begin{promptbox}{Prompt B.8: Reference generation, verifiable decision-only variant (P2-DECISION-V)}
\begin{lstlisting}[style=promptbox]
MOMENT
- type: {type}
- transcript_window (-5 s, forward to the next pedagogical event or +45 s, whichever comes first; word timestamps): {transcript_window}
- slide_ocr: {slide_ocr}
- pause_after: {pause_after} s
- reachable_rung: {reachable_rung}          # computed in code; do not recompute
- described_so_far (this lecture): {described_so_far}
- keyframes: [3-5 attached]

TASK

Step 1 -- coverage. From the transcript window and keyframes alone:
  coverage = FULLY_COVERED if the lecturer's words in the window already name or convey the
             visual content of this moment. Put the covering span in covering_quote, verbatim.
  coverage = PARTLY_COVERED if some but not all is spoken. State what is left in
             uncovered_content.
  coverage = UNCOVERED if the visual carries content the window does not convey. State it in
             uncovered_content.
  An unresolved deictic -- the lecturer says "this" or "here" and the window never names the
  referent -- is UNCOVERED.

Step 1b -- visual content (visual_content_001, slide_text_003). From the keyframes:
  content = VISUAL if a non-text element is on screen (figure, diagram, chart, table, equation,
            code, image, animation) or the lecturer's cursor, laser or hand indicates a specific
            element. Name that element in visual_referent. Then set unspoken_form to the ONE
            property of it that the window does not convey and a listener needs to follow the
            point -- which_element, shape_trend, spatial_arrangement, relative_magnitude, count,
            value, colour_code -- or null if the speech already conveys its form. Hearing the
            concept is not hearing the form; the slide existing does not make it describable.
  content = TEXT_ONLY if the slide surface carries only text (title, bullets, prose). Text has no
            form to describe; a paraphrase of it is not audio description.
  content = NONE if nothing on screen carries content (blank, transition frame, webcam only).
  visual_referent and unspoken_form are null unless content == VISUAL.

Step 2 -- emit. emit=true ONLY if coverage != FULLY_COVERED and either
  (a) content == VISUAL and unspoken_form is not null, or
  (b) content == TEXT_ONLY and coverage == UNCOVERED -- then Step 3 -- there is no step 3. Do NOT write an AD.

Output ONLY this JSON, no preamble, no fences:
{{"coverage": "FULLY_COVERED"|"PARTLY_COVERED"|"UNCOVERED",
 "covering_quote": <string or null>,
 "uncovered_content": <string or null>,
 "content": "NONE"|"TEXT_ONLY"|"VISUAL",
 "visual_referent": <string or null>,
 "unspoken_form": "which_element"|"shape_trend"|"spatial_arrangement"|"relative_magnitude"|"count"|"value"|"colour_code"|null,
 "emit": <bool>}}
\end{lstlisting}
\end{promptbox}

\subsection{Candidate generation with focus directives (P3)}
\label{app:prompt-p3}
The following prompt was employed to generate the candidate descriptions from which natural preference pairs are drawn (Section~\ref{sec:standard}); the five focus directives are listed first, one of which is substituted into the body.
\begin{promptbox}{Prompt B.9: Candidate generation with focus directives (P3)}
\begin{lstlisting}[style=promptbox]
FOCUS DIRECTIVES
[REFERENT]
name the specific element the speech leaves unresolved (the pointed, highlighted or newly salient item) and its on-slide identity or role. Say 'the cursor' or 'the lecturer', never he/she.

[STRUCTURE]
type-and-purpose in one sentence, then main components in reading order (figure_001/002/004), levels 1-2 only.

[CHANGE]
only the delta this moment adds relative to the previous slide state (rule 18). Never re-read accumulated content.

[DATA]
the quantitative content: trend, key values, or the equation with a one-line ClearSpeak gloss (rules 23-31).

[SPATIAL]
where the uncovered elements sit and how they relate positionally. Content-bearing layout only, no styling (rules 20-21).

BODY
This moment warrants description. Produce ONE candidate AD under the focus directive below.
The redundancy gate still applies WITHIN your focus: never restate content the transcript
window already conveys -- check it against the window before writing.

MOMENT
- type: {type}
- transcript_window (-5 s / +8 s, word timestamps): {transcript_window}
- slide_ocr: {slide_ocr}
- pause_after: {pause_after} s
- reachable_rung: {reachable_rung}
- described_so_far (this lecture): {described_so_far}
- keyframes: [3-5 attached]

FOCUS DIRECTIVE: {directive}
  REFERENT  -- name the specific element the speech leaves unresolved (the pointed, highlighted
              or newly salient item) and its on-slide identity or role. Say 'the cursor' or
              'the lecturer', never he/she.
  STRUCTURE -- type-and-purpose in one sentence, then main components in reading order
              (figure_001/002/004), levels 1-2 only.
  CHANGE    -- only the delta this moment adds relative to the previous slide state (rule 18).
              Never re-read accumulated content.
  DATA      -- the quantitative content: trend, key values, or the equation with a one-line
              ClearSpeak gloss (rules 23-31).
  SPATIAL   -- where the uncovered elements sit and how they relate positionally. Content-bearing
              layout only, no styling (rules 20-21).

LENGTH BAND: {band} -- write between {band_min} and {band_max} words.
Count the words. The upper limit is computed for THIS moment from the gap actually available
after it, so it is often tighter than the band's nominal width. Stay inside it -- a hard
constraint, not a target.

Choose the rung honestly for the text you produced.

Output ONLY the JSON object, no preamble, no fences:
{{"ad_text": <string>, "rung": <1-4>, "rationale": <string>}}
\end{lstlisting}
\end{promptbox}

\subsection{Candidate generation, verbalized-sampling variant (P3-VS)}
\label{app:prompt-p3vs}
The following variant, which asks for five candidates per call, is the default arm that produced the natural pairs.
\begin{promptbox}{Prompt B.10: Candidate generation, verbalized-sampling variant (P3-VS)}
\begin{lstlisting}[style=promptbox]
This moment warrants description. Produce ONE candidate AD under the focus directive below.
The redundancy gate still applies WITHIN your focus: never restate content the transcript
window already conveys -- check it against the window before writing.

MOMENT
- type: {type}
- transcript_window (-5 s / +8 s, word timestamps): {transcript_window}
- slide_ocr: {slide_ocr}
- pause_after: {pause_after} s
- reachable_rung: {reachable_rung}
- described_so_far (this lecture): {described_so_far}
- keyframes: [3-5 attached]

FOCUS DIRECTIVE: {directive}
  REFERENT  -- name the specific element the speech leaves unresolved (the pointed, highlighted
              or newly salient item) and its on-slide identity or role. Say 'the cursor' or
              'the lecturer', never he/she.
  STRUCTURE -- type-and-purpose in one sentence, then main components in reading order
              (figure_001/002/004), levels 1-2 only.
  CHANGE    -- only the delta this moment adds relative to the previous slide state (rule 18).
              Never re-read accumulated content.
  DATA      -- the quantitative content: trend, key values, or the equation with a one-line
              ClearSpeak gloss (rules 23-31).
  SPATIAL   -- where the uncovered elements sit and how they relate positionally. Content-bearing
              layout only, no styling (rules 20-21).

LENGTH BAND: {band} -- write between {band_min} and {band_max} words.
Count the words. The upper limit is computed for THIS moment from the gap actually available
after it, so it is often tighter than the band's nominal width. Stay inside it -- a hard
constraint, not a target.

Choose the rung honestly for the text you produced.

Produce FIVE candidate ADs for this moment under the focus directive and length band above,
each with the probability you would assign it as the description a professional describer
would actually write for this moment. Sample across the full distribution, not only the most
obvious phrasing. The candidates must differ in content selection or framing, not only in
wording.

Every candidate must independently satisfy the standard: inside the band, rule-compliant,
faithful to the keyframes, non-redundant with the transcript window.

Output ONLY a JSON array of five objects, no preamble, no fences:
[{{"ad_text": <string>, "rung": <1-4>, "probability": <float>, "rationale": <string>}}, ...]
\end{lstlisting}
\end{promptbox}

\subsection{Candidate generation, verbalized-sampling with rationale (P3-VS-COT)}
\label{app:prompt-p3vscot}
The following variant adds a rationale field and was run only in the bake-off that selected P3-VS.
\begin{promptbox}{Prompt B.11: Candidate generation, verbalized-sampling with rationale (P3-VS-COT)}
\begin{lstlisting}[style=promptbox]
This moment warrants description. Produce ONE candidate AD under the focus directive below.
The redundancy gate still applies WITHIN your focus: never restate content the transcript
window already conveys -- check it against the window before writing.

MOMENT
- type: {type}
- transcript_window (-5 s / +8 s, word timestamps): {transcript_window}
- slide_ocr: {slide_ocr}
- pause_after: {pause_after} s
- reachable_rung: {reachable_rung}
- described_so_far (this lecture): {described_so_far}
- keyframes: [3-5 attached]

FOCUS DIRECTIVE: {directive}
  REFERENT  -- name the specific element the speech leaves unresolved (the pointed, highlighted
              or newly salient item) and its on-slide identity or role. Say 'the cursor' or
              'the lecturer', never he/she.
  STRUCTURE -- type-and-purpose in one sentence, then main components in reading order
              (figure_001/002/004), levels 1-2 only.
  CHANGE    -- only the delta this moment adds relative to the previous slide state (rule 18).
              Never re-read accumulated content.
  DATA      -- the quantitative content: trend, key values, or the equation with a one-line
              ClearSpeak gloss (rules 23-31).
  SPATIAL   -- where the uncovered elements sit and how they relate positionally. Content-bearing
              layout only, no styling (rules 20-21).

LENGTH BAND: {band} -- write between {band_min} and {band_max} words.
Count the words. The upper limit is computed for THIS moment from the gap actually available
after it, so it is often tighter than the band's nominal width. Stay inside it -- a hard
constraint, not a target.

Choose the rung honestly for the text you produced.

First, in at most 40 words, name the distinct content choices available at this moment under the
focus directive -- the different things a describer could reasonably decide are worth the pause.
Do not write any description yet.

Then produce FIVE candidate ADs realising those choices, each with the probability you would
assign it as the description a professional describer would actually write. The candidates must
differ in content selection or framing, not only in wording.

Every candidate must independently satisfy the standard: inside the band, rule-compliant,
faithful to the keyframes, non-redundant with the transcript window.

Output ONLY:
{{"plan": <string>,
  "candidates": [{{"ad_text": <string>, "rung": <1-4>, "probability": <float>,
                  "rationale": <string>}}, ...]}}
\end{lstlisting}
\end{promptbox}

\subsection{Controlled-pair generation (P4)}
\label{app:prompt-p4}
The following prompt was employed to generate the controlled pairs that differ on exactly one rule category (Section~\ref{sec:standard}).
\begin{promptbox}{Prompt B.12: Controlled-pair generation (P4)}
\begin{lstlisting}[style=promptbox]
Produce TWO audio descriptions of this moment that are EQUALLY FAITHFUL to the keyframes and
state IDENTICAL FACTS, and that differ ONLY in the property named below. Everything else --
content selected, facts stated, word count -- must match as closely as possible.

MOMENT
- type: {type}
- transcript_window (-5 s / +8 s, word timestamps): {transcript_window}
- slide_ocr: {slide_ocr}
- pause_after: {pause_after} s
- reachable_rung: {reachable_rung}
- described_so_far (this lecture): {described_so_far}
- keyframes: [3-5 attached]

AXIS: {axis}
  style          A follows the style specification; B uses forbidden meta-language
                 ("the slide shows", "we can see") while stating the same facts.
  terminology    A uses the on-slide term exactly as written; B substitutes a different term
                 that is also on the slide, making the reference wrong.
  length         A fits the reachable rung's budget; B overruns it substantially while adding
                 no information.
  deixis         A names the referent; B narrates the pointing act ("the cursor moves to...")
                 without naming what is pointed at.
  faithfulness   A states only what the keyframes support; B adds one specific detail the
                 keyframes do not support.
  non_redundancy A states only what the transcript window leaves uncovered; B restates content
                 the window already conveys.

A is the compliant side, B the defective side. Keep |words(A) - words(B)| <= 3 except on the
length axis, where the difference is the manipulation.

Output ONLY: {{"axis": "{axis}", "a": <string>, "b": <string>,
              "what_differs": "<one sentence>"}}
\end{lstlisting}
\end{promptbox}

\subsection{Pairwise video judge (J1)}
\label{app:prompt-j1}
The following prompt was employed for the video-watching judge that ordered the natural pairs, judging each pair in both presentation orders (Section~\ref{sec:standard}).
\begin{promptbox}{Prompt B.13: Pairwise video judge (J1)}
\begin{lstlisting}[style=promptbox]
Watch this clip of a lecture moment. Two audio descriptions of THIS moment, written for a
blind student who cannot see the screen and hears only the lecturer plus the description.

Judge which description better gives that student the information the speech leaves out: does it
convey what is on screen and what changed, in terms the student can use, without repeating what
the lecturer already says.

The clip is the evidence, not the target. A description that asserts something the clip does not
support is disqualified however well it reads. But do not reward a description for narrating
visible activity when the other conveys what that activity is about.

Description 1: {d1}
Description 2: {d2}

Answer with ONLY '1', '2', or 'TIE'.
\end{lstlisting}
\end{promptbox}

\subsection{Best-of-four jury (J2)}
\label{app:prompt-j2}
The following prompt was employed on the training side only, to select one of four candidate descriptions per moment.
\begin{promptbox}{Prompt B.14: Best-of-four jury (J2)}
\begin{lstlisting}[style=promptbox]
Four audio descriptions of the same lecture moment, for a blind student. Pick the one that best
serves the learner, judging on: does it surface what the speech leaves uncovered; is it faithful
to the keyframes; does it name the referent rather than the pointing act; is it free of
meta-language; does it fit the stated delivery rung.

Do not prefer a description because it reads more smoothly or more conventionally. Judge only
against the five criteria above.

MOMENT
- Visible: {what_on_screen}
- Lecturer said (+/-8 s): {transcript_window}
- Reachable rung: {reachable_rung}
- keyframes: [attached]

A: {a}   B: {b}   C: {c}   D: {d}

Answer with ONLY 'A', 'B', 'C' or 'D'.
\end{lstlisting}
\end{promptbox}

\subsection{Judge-family screening}
\label{app:prompt-screen}
The following prompt was employed to screen candidate judge families for video reception before J1 was run.
\begin{promptbox}{Prompt B.15: Judge-family screening}
\begin{lstlisting}[style=promptbox]
Watch this short clip of a lecture slide.

Answer in at most 25 words: what MOVES or CHANGES on screen during the clip? If nothing moves,
say "nothing moves". If you cannot see any video, say exactly "NO VIDEO RECEIVED".
\end{lstlisting}
\end{promptbox}

\subsection{Prior-ablation sentences (P2 arms B and C)}
\label{app:prompt-prior}
The following two sentences are spliced into P2 as the prior-ablation arms.
\begin{promptbox}{Prompt B.16: Prior-ablation sentences (P2 arms B and C)}
\begin{lstlisting}[style=promptbox]
[B]
('gemini_prompt.txt:16', 'A typical lecture has multiple REDUNDANT moments -- usually at least 5-10 per 30 minutes of footage. If your output has zero or near-zero REDUNDANT rows for a normal lecture, you are missing them.')

[C]
('gemini_prompt.txt:76', 'A real lecture has BOTH types. If you find yourself producing all-YES or all-NO pointing rows, re-examine -- the actual distribution should be mixed, typically somewhere between 30% and 70% NO depending on lecturer style.')
\end{lstlisting}
\end{promptbox}

\subsection{Ground-truth moment marking}
\label{app:prompt-f4}
The following prompt was employed for the check of detector coverage against independently marked moments.
\begin{promptbox}{Prompt B.17: Ground-truth moment marking}
\begin{lstlisting}[style=promptbox]
You are marking candidate pedagogical moments in a slide-based lecture recording,
for the construction of ground truth in an audio-description benchmark. Your output
will be checked by a human annotator, who will confirm, reject, and add to it.

Report every candidate you observe. Do not filter for importance, do not decide
whether a moment deserves description, and do not consider whether the lecturer's
speech already covers it -- those are separate downstream judgements. Your only task
is: did something change on the slide surface, and when.

## Definition

A pedagogical moment is a time point at which a change occurs on the slide surface.
Mark one candidate per distinct change, at the time the change begins.

Types (use exactly these strings):
  slide       a new slide replaces the previous one
  transition  an animated change between slide states
  reveal      an increment appears on the existing slide (build step, bullet, overlay)
  pointing    the cursor, laser, or a highlight indicates a specific on-slide element
  ink         handwritten annotation is drawn on the slide
  figure      a diagram, image, or illustration first appears
  chart       a plot or graph first appears
  table       a table first appears
  math        typeset mathematics first appears
  code        a code block first appears
  animation   embedded animated content plays (video, 3D model, simulation)
  other       a change on the slide surface fitting none of the above

Mark as `pointing` only when the cursor or highlight comes to rest on, dwells on,
or traces a specific element. Cursor movement across the screen that does not settle
on anything is not a candidate. Do not mark the lecturer's webcam thumbnail, slide
numbers, footers, or template chrome.

If a single instant carries two changes (a new slide that is also the first
appearance of a chart), emit one candidate per type at the same timestamp.

## Input

The attached video is a {clip_len:.0f} second clip.

Its first {lead_in:.0f} seconds are LEAD-IN, provided only so you can see the slide state
that already existed. **Do not report any candidate in the lead-in.** Report only
candidates at or after {lead_in:.0f} seconds into the clip.

TRANSCRIPT for this clip, timestamps relative to the clip start:
{transcript_window}

The transcript is context to help you locate and identify what is on screen.
It does not determine whether something is a candidate: a change that the lecturer
describes in full is still a candidate.

## Output

One JSON object per line, no preamble, no markdown fences, no commentary.

**Timestamps are MM:SS measured from the START OF THIS CLIP** -- not from the start of
the lecture. The clip begins at 00:00. Do not attempt to convert to lecture time.

{{"t": "<MM:SS>", "type": "<type>", "what_on_screen": "<what is visible or what changed, one sentence, observable content only>", "confidence": <1|2|3>, "evidence": "<what in the video indicated this -- one short phrase>"}}

confidence: 3 = unambiguous; 2 = probable; 1 = uncertain, flagged for the human
to adjudicate. Use 1 freely -- an uncertain candidate a human rejects costs less
than a moment nobody sees.

Order by timestamp. If a window contains no candidates, output nothing.
\end{lstlisting}
\end{promptbox}

\subsection{Pair judge, text rubric}
\label{app:prompt-rejudge-text}
The following prompt scores a pair simply from the two descriptions alone, without the moment's frames.
\begin{promptbox}{Prompt B.18: Pair judge, text rubric}
\begin{lstlisting}[style=promptbox]
Two audio descriptions (AD) of the SAME lecture moment, for a blind student. Which better follows the AD standard: faithful to the visual, non-redundant with the speech, names the referent (not the act of pointing), concise for the pause, no meta-language ('the slide shows')?

MOMENT: {moment}

Description 1: {d1}
Description 2: {d2}

Answer with ONLY '1', '2', or 'TIE'.
\end{lstlisting}
\end{promptbox}

\subsection{Pair judge, visual}
\label{app:prompt-rejudge-visual}
The following prompt orders a pair from the moment's frames and differs from the J1 prompt above.
\begin{promptbox}{Prompt B.19: Pair judge, visual}
\begin{lstlisting}[style=promptbox]
Below are two audio descriptions (AD) of the SAME lecture moment, for a blind student. Judge which one better matches WHAT ACTUALLY HAPPENS AT THIS MOMENT -- faithful to what is shown, names the right on-screen things, resolves any pointing to the real referent.

MOMENT: {moment}

Description 1: {d1}
Description 2: {d2}

Answer with ONLY '1', '2', or 'TIE'.
\end{lstlisting}
\end{promptbox}

\subsection{Clip judge}
\label{app:prompt-videojudge}
The following prompt orders a pair from the moment's video clip.
\begin{promptbox}{Prompt B.20: Clip judge}
\begin{lstlisting}[style=promptbox]
Watch this short lecture video clip. Below are two audio descriptions (AD) of THIS moment, written for a blind student who cannot see the screen and hears only the lecturer plus the description.

Judge which description better gives that student the information the speech leaves out: does it convey what is on screen and what changed, in terms the student can use, without repeating what the lecturer already says, and briefly enough to fit the pause.

The clip is the evidence, not the target. A description that asserts something the clip does not support is disqualified however well it reads. But do not reward a description for narrating visible activity when the other conveys what that activity is about.

Description 1: {d1}
Description 2: {d2}

Answer with ONLY '1', '2', or 'TIE'.
\end{lstlisting}
\end{promptbox}

\subsection{Per-category pair judge}
\label{app:prompt-judgecat}
The following prompt scores a pair on each of the six rule categories separately.
\begin{promptbox}{Prompt B.21: Per-category pair judge}
\begin{lstlisting}[style=promptbox]
You are judging two audio descriptions (AD) of the SAME lecture moment, written for a blind student. The images are stills sampled across the moment.

LECTURER IS SAYING: {transcript}
SILENCE AVAILABLE IN THE NEXT 5 SECONDS: {pause}s

Description 1: {d1}

Description 2: {d2}

For EACH property below, say which description is better, or 'tie' if they are equivalent:
  faithfulness   - matches what the stills actually show; invents nothing
  non_redundancy - adds visual information the lecturer is not already saying
  deixis         - names what things ARE rather than leaving 'this'/'here' unresolved
  length         - deliverable in the time available; not padded, not uselessly sparse
  style          - direct present-tense narration; no 'the slide shows' meta-language
  terminology    - uses the lecture's own terms for on-screen things

Return ONLY compact JSON, no markdown fences:
{{"faithfulness":"1|2|tie","non_redundancy":"1|2|tie","deixis":"1|2|tie","length":"1|2|tie","style":"1|2|tie","terminology":"1|2|tie"}}
\end{lstlisting}
\end{promptbox}

\subsection{Dimension attribution}
\label{app:prompt-valdim}
The following prompt names the dimension on which the weaker side of a pair fails.
\begin{promptbox}{Prompt B.22: Dimension attribution}
\begin{lstlisting}[style=promptbox]
"Watch this short lecture video clip. Below are two audio descriptions (AD) of this moment for a "
    "blind student. The SECOND one is the weaker of the two.\n\n"
    "In ONE word, which single property does the weaker one most violate?\n"
    "  faithfulness - says something the clip does not show, or gets a visual detail wrong\n"
    "  redundancy   - repeats what the lecturer is already saying\n"
    "  length       - far too long, or too sparse, for the moment\n"
    "  deixis       - leaves pointing words unresolved ('this', 'here') instead of naming the referent\n"
    "  style        - meta-language ('the slide shows'), wrong register, or awkward phrasing\n\n"
    "LECTURER SAID: {transcript}\n\nBETTER: {best}\nWORSE: {worst}\n\n"
    "Answer with ONLY one of: " + ", ".join(DIMS) + "."
\end{lstlisting}
\end{promptbox}

\subsection{Style held-out generator}
\label{app:prompt-styleheld}
The following prompt generated the held-out style negatives used to test the style component (Section~\ref{sec:res-metric}).
\begin{promptbox}{Prompt B.23: Style held-out generator}
\begin{lstlisting}[style=promptbox]
Produce TWO audio descriptions of this moment that state IDENTICAL FACTS and are
EQUALLY FAITHFUL to the keyframes, differing ONLY in the property named below.

MOMENT
- type: {type}
- transcript_window (-5 s / +8 s): {transcript_window}
- slide_ocr: {slide_ocr}
- keyframes: [3-5 attached]

DIFFERENCE: {variant_desc}

A is the compliant side, B the defective side. Keep |words(A) - words(B)| <= 3.

**HARD CONSTRAINT on B**: it must NOT contain any of these exact phrases, in any casing:
{banned}
B must break the style rule in the way named above, using DIFFERENT wording. If you cannot do
that, return {{"skip": true}}.

Output ONLY the JSON object, no preamble, no fences:
{{"a": <string>, "b": <string>, "variant": "{variant}", "what_differs": "<one sentence>"}}
\end{lstlisting}
\end{promptbox}

\subsection{Pool rating and dimension tagging}
\label{app:prompt-rank}
The following two prompts are the pair-pool helpers, which rate a single description and tag the dimension a pair differs on.
\begin{promptbox}{Prompt B.24: Pool rating and dimension tagging}
\begin{lstlisting}[style=promptbox]
RATING
Rate this audio description (AD) of a lecture moment for a blind student, 0-10, by the AD standard (faithful to the visual, non-redundant with the speech, names the referent not the pointing act, concise for the pause, no meta-language like 'the slide shows').

MOMENT: {moment}

AD: {ad}

Return ONLY an integer 0-10.

DIMENSION
Two audio descriptions of the same lecture moment; the SECOND is worse. In ONE word, which dimension does the second most violate: {dims}?

MOMENT: {moment}

BETTER: {best}
WORSE: {worst}

Return only one of: {dims}.
\end{lstlisting}
\end{promptbox}

\subsection{Near-tie screen}
\label{app:prompt-tie}
The following prompt screens whether two descriptions of one moment are substantively equivalent before a pair is admitted to the benchmark pool.
\begin{promptbox}{Prompt B.25: Near-tie screen}
\begin{lstlisting}[style=promptbox]
You are screening two audio descriptions (AD) written for the SAME lecture moment, for a blind student. Judge only whether they are SUBSTANTIVELY EQUIVALENT in the information they convey -- not which is better, not their wording.

Description 1: {a}
Description 2: {b}

Answer with exactly one word: EQUIVALENT (they convey essentially the same content) or DIFFERENT (a blind learner would get materially different information).
\end{lstlisting}
\end{promptbox}

\subsection{Synthetic defect instructions}
\label{app:prompt-perturb}
The following per-category instructions, and the wrapper below them, produced synthetic defective sides for the perturbation checks; the wrapper is shown as source because it is assembled at run time.
\begin{promptbox}{Prompt B.26: Synthetic defect instructions}
\begin{lstlisting}[style=promptbox]
[style]
Rewrite the audio description so it slips into forbidden meta-language a describer should avoid (phrases like 'the slide shows', 'we can see', 'an image of'), while conveying the SAME content.

[deixis]
Rewrite so it describes the ACT of pointing (the cursor/presenter pointing or hovering) instead of naming the actual thing being indicated. Keep it a single natural sentence.

[redundancy]
Rewrite so it redundantly restates what the lecturer ALREADY said in this transcript, adding no new visual information: '{transcript}'. Keep it natural.

[length]
Rewrite to be far too long and verbose for a short speech gap (at least 120 words), padding with extra visual detail while staying on-topic.

WRAPPER (source)
def make_llm_perturber(
    model: str = _LLM_MODEL,
    cache_path: str | Path | None = None,
    client: Any = None,
) -> Callable[[str, str, dict[str, Any]], str | None]:
    """Build a ``(dimension, ad, context) -> degraded`` natural-rewrite perturber (PAID OpenRouter).

    Degradations are more realistic than the deterministic templates (a harder metric test). Cached
    to JSONL for reproducibility. ``client`` may be injected (tests/offline); else an OpenRouter
    client is built from ``OPENROUTER_API_KEY``. Returns None on failure -> caller falls back to the
    deterministic perturber.
    """
    cache_path = Path(cache_path) if cache_path else None
    if cache_path:
        cache_path.parent.mkdir(parents=True, exist_ok=True)
    cache: dict[str, str] = {}
    if cache_path and cache_path.exists():
        for line in cache_path.read_text(encoding="utf-8").splitlines():
            if line.strip():
                d = json.loads(line)
                cache[d["key"]] = d["rejected"]

    def _client() -> Any:
        if client is not None:
            return client
        from openai import OpenAI

        key = os.environ.get("OPENROUTER_API_KEY")
        if not key:
            raise RuntimeError("OPENROUTER_API_KEY not set (LLM perturbation is a PAID call).")
        return OpenAI(base_url="https://openrouter.ai/api/v1", api_key=key)

    holder: dict[str, Any] = {}

    def perturb(dimension: str, ad: str, ctx: dict[str, Any]) -> str | None:
        key = f"{dimension}[US]{ad}"
        if key in cache:
            return cache[key]
        instr = _LLM_INSTRUCTION[dimension].format(transcript=(ctx.get("transcript_window") or "")[:400])
        prompt = (
            f"You are editing an audio description (AD) of a lecture moment to create a WORSE version for "
            f"a benchmark. {instr}\n\nOriginal AD: {ad}\n\nReturn ONLY the rewritten AD, no preamble."
        )
        try:
            cli = holder.get("c") or holder.setdefault("c", _client())
            resp = cli.chat.completions.create(
                model=model, messages=[{"role": "user", "content": prompt}], temperature=0.7, max_tokens=300
            )
            out = (resp.choices[0].message.content or "").strip()
        except Exception:
            return None
        if not out:
            return None
        if cache_path:
            with open(cache_path, "a", encoding="utf-8") as f:
                f.write(json.dumps({"key": key, "model": model, "rejected": out}) + "\n")
        cache[key] = out
        return out

    return perturb
\end{lstlisting}
\end{promptbox}

\subsection{\textsc{AD4Edu-Eval} rubric judge}
\label{app:prompt-rubric}
The following prompt was employed for the rubric mode of \textsc{AD4Edu-Eval} (Section~\ref{sec:metric}) both on the 400 evaluation pairs and at system level; the \texttt{\{standard\}} slot receives the compressed standard shown as the training system prompt below, and the \texttt{\{refs\}} slot the moment's references with the leakage rule applied.
\begin{promptbox}{Prompt B.27: \textsc{AD4Edu-Eval} rubric judge}
\begin{lstlisting}[style=promptbox]
The operative standard for audio description of slide-based lectures:

{standard}

---
MOMENT CONTEXT (what a sighted viewer has):
What is on screen: {on_screen}
Slide text (OCR): {ocr}
Spoken (transcript window): {transcript}

REFERENCE ADs of this moment (independent describers):
{refs}

CANDIDATE AD:
{cand}

Grade the CANDIDATE against the standard and the references. Reply with EXACTLY seven lines:
style: <1-5>
terminology: <1-5>
length: <1-5>
deixis: <1-5>
faithfulness: <1-5>
non_redundancy: <1-5>
overall: <1-10>
No other text.
\end{lstlisting}
\end{promptbox}

\subsection{LLM-AD-Eval (single reference)}
\label{app:prompt-llmad}
The following prompt was employed for the LLM-AD-Eval column (Section~\ref{sec:auto-eval}).
\begin{promptbox}{Prompt B.28: LLM-AD-Eval (single reference)}
\begin{lstlisting}[style=promptbox]
You are evaluating an audio description (AD) of a video moment for a blind viewer.

REFERENCE AD:
{ref}

CANDIDATE AD:
{cand}

Rate how well the CANDIDATE conveys the same visual information as the REFERENCE, on a 1-5 scale:
1 = unrelated or contradictory, 2 = poor, 3 = partial, 4 = good, 5 = equivalent.
Answer with a single integer from 1 to 5 and nothing else.
\end{lstlisting}
\end{promptbox}

\subsection{LLM-AD-Eval (multi-reference, 400-pair validation)}
\label{app:prompt-llmad400}
The following variant of LLM-AD-Eval, with all references of a moment, was employed for the human-agreement analysis of Section~\ref{sec:res-human}.
\begin{promptbox}{Prompt B.29: LLM-AD-Eval (multi-reference, 400-pair validation)}
\begin{lstlisting}[style=promptbox]
You are evaluating an audio description (AD) of a lecture-video moment for a blind viewer.

REFERENCE ADs of the same moment (independent describers):
{refs}

CANDIDATE AD:
{cand}

Rate how well the CANDIDATE conveys the same visual information as the references, on a 1-5 scale: 1 = unrelated or contradictory, 2 = poor, 3 = partial, 4 = good, 5 = equivalent.
Answer with a single integer from 1 to 5 and nothing else.
\end{lstlisting}
\end{promptbox}

\subsection{Per-rule compliance judge (template)}
\label{app:prompt-compliance}
The following template is rendered once per rule into the 42 per-rule judge prompts of the compliance judge; the rendered files are byte-identical up to the rule text and grounding slot, and are not reproduced individually.
\begin{promptbox}{Prompt B.30: Per-rule compliance judge (template)}
\begin{lstlisting}[style=promptbox]
You are evaluating whether an audio description (AD) example for a lecture moment satisfies a specific rule from the operative AD standard.

## The rule

rule_id: {rule_id}
section: {section}   |   element: {element}   |   type: {rule_type}

{rule_body}

Grounding / basis (verbatim from the standard):
{grounding}

## Context

Lecture moment:
- Type: {type}
- Visible on slide: {what_on_screen}
- Lecturer said (+/-8s window): {transcript_window}
- Redundant: {redundant}
- Pointing clear: {pointing_clear}
- Emit decision: {emit}

## The AD to evaluate

{ad_text}

## Your task

This rule has been pre-determined to apply to this moment. Evaluate whether the AD satisfies it.

Output ONLY this JSON (no preamble, no markdown fences):
{"verdict": "PASS" | "FAIL", "reasoning": "<one sentence>"}
\end{lstlisting}
\end{promptbox}

\subsection{Head-to-head win-rate judge}
\label{app:prompt-winrate}
The following prompt was employed by the two machine annotators of Section~\ref{sec:human-eval} to judge a pair of generated descriptions, in both presentation orders.
\begin{promptbox}{Prompt B.31: Head-to-head win-rate judge}
\begin{lstlisting}[style=promptbox]
You are judging two candidate audio descriptions (AD) of the same lecture moment for a blind
learner, against the operative AD standard (describe only what speech omits; faithful to the
slide; concise enough for the pause; no meta-language).

MOMENT
- Type: {type}
- Lecturer said (+/-8s): {transcript_window}
- On-screen text (OCR): {slide_ocr}
- Pause available: {pause_after}s

DESCRIPTION 1:
{ad1}

DESCRIPTION 2:
{ad2}

Which description better serves the blind learner at this moment?
Answer with exactly one token: "1", "2", or "TIE".
\end{lstlisting}
\end{promptbox}

\subsection{Machine annotator on the rating sheet}
\label{app:prompt-sheet}
The following prompt let a model stand in for a rater on the 400-pair sheet; its outputs appear as the two machine-annotator rows of Table~\ref{tab:human}.
\begin{promptbox}{Prompt B.32: Machine annotator on the rating sheet}
\begin{lstlisting}[style=promptbox]
You are standing in for a blind student who is judging audio descriptions of a university lecture.

The student hears the lecturer and, at this moment, one short spoken description of what is on the
screen. They cannot see the slide. Two descriptions were written for THIS moment. You are shown both,
together with the slide, the text on it, and what the lecturer is saying around the moment.

MOMENT
- Subject: {subject}
- Moment type: {moment_type}
- What is on screen: {on_screen}
- Text on the slide: {slide_ocr}
- The lecturer is saying: {transcript_window}

DESCRIPTION 1:
{ad1}

DESCRIPTION 2:
{ad2}

Which description better gives that student the information the speech leaves out: does it convey what
is on the screen and what changed, in terms the student can use, without repeating what the lecturer
already says? The slide is the evidence, not the target: a description that asserts something the slide
does not support is disqualified however well it reads, but do not reward a description for narrating
visible activity when the other conveys what that activity is about. If neither is better for the
student, say TIE.

Answer with ONLY "1", "2", or "TIE" on the first line.
On a second line write "REASON:" and at most twenty words.
\end{lstlisting}
\end{promptbox}

\subsection{Fact and claim decomposition (certificate sidecar)}
\label{app:prompt-cert}
The following two prompts decompose a moment and a description into atomic facts and claims for the certificate sidecar of the benchmark definition.
\begin{promptbox}{Prompt B.33: Fact and claim decomposition (certificate sidecar)}
\begin{lstlisting}[style=promptbox]
FACTS
A lecture-video moment. Context:
What is on screen: {on_screen}
Slide text (OCR): {ocr}
Spoken (transcript window): {transcript}

Independent audio descriptions of this moment:
{refs}

List the REQUIRED atomic visual facts a competent audio description of this moment must convey for a blind listener -- the visual content the speech does not already carry. One short declarative sentence per fact, at most 6 facts, no numbering beyond a leading dash.
Reply with ONLY the dashed list.

CLAIMS
Split this audio description into its atomic VISUAL claims (one short declarative sentence per claim; only what it asserts about the visuals; at most 8):

{ad}

Reply with ONLY a dashed list.
\end{lstlisting}
\end{promptbox}

\subsection{Single-description quality rating}
\label{app:prompt-rwjudge}
The following prompt was employed in a pilot single-description rating study.
\begin{promptbox}{Prompt B.34: Single-description quality rating}
\begin{lstlisting}[style=promptbox]
You are rating one audio description (AD) of a slide-lecture moment for a blind or low-vision student.

MOMENT CONTEXT
Spoken (transcript window):
{transcript}

Slide text (OCR):
{ocr}

What is on screen: {on_screen}

AD TO RATE:
{ad}

Rate the AD's overall quality for a BLV student on a 1-5 scale:
1 = unusable or misleading, 2 = poor, 3 = acceptable, 4 = good, 5 = excellent (accurate, concise, adds visual information the speech does not already carry, resolves any pointing references like 'this' or 'here').
Answer with a single integer from 1 to 5 and nothing else.
\end{lstlisting}
\end{promptbox}

\subsection{Comprehension questions (optional metric)}
\label{app:prompt-cqa}
The following two prompts implement an optional comprehension-question metric.
\begin{promptbox}{Prompt B.35: Comprehension questions (optional metric)}
\begin{lstlisting}[style=promptbox]
GENERATE
You are creating comprehension questions to test whether an AUDIO DESCRIPTION would help a blind or low-vision learner understand a lecture moment.

Given what is VISUALLY on the slide and what the lecturer SAID, write {n} multiple-choice questions whose correct answers depend on VISUAL content that the speech does NOT convey. Each question must be answerable from the visuals but NOT from the transcript alone. Prefer concrete visual facts: spatial layout, quantities, labels, structure, relations.

VISUAL (slide OCR + on-screen description):
{visual}

TRANSCRIPT (what the lecturer said around the moment):
{transcript}

Output ONLY a JSON array of {n} objects, no prose:
[{{"dimension": "spatial|quantity|label|structure|relation", "question": "...", "options": {{"A": "...", "B": "...", "C": "...", "D": "..."}}, "answer": "A"}}]


ANSWER
Answer the multiple-choice question using ONLY the information below. If it is insufficient, choose the single best guess.

INFORMATION:
{info}

QUESTION: {question}
A) {A}
B) {B}
C) {C}
D) {D}

Output ONLY the letter of the best answer (A, B, C, or D).
\end{lstlisting}
\end{promptbox}

\subsection{Training and inference system prompt}
\label{app:prompt-system}
The following system prompt is baked into every SFT and DPO example and replayed verbatim at inference (Section~\ref{sec:models}); it is the compressed standard that the rubric judge also receives.
\begin{promptbox}{Prompt B.36: Training and inference system prompt}
\begin{lstlisting}[style=promptbox]
You are an audio describer for slide-based lecture videos, writing for blind and low-vision learners. For each lecture moment you receive the on-screen visuals (keyframe images and slide OCR text), the lecturer's speech around the moment, the moment type, and the length of the pause available.

Follow the operative standard:
- Redundancy gate: describe on-screen content ONLY to the extent the lecturer's speech does not already convey it. If the speech already carries the content, suppress (do not describe).
- Fit the pause: keep the description within what the available pause allows.
- Be faithful: describe only what is actually visible; never invent content.
- Style: present tense, active voice, third person; concise; no meta-language ("the slide shows", "an image of").

Output ONLY a JSON object: {"emit": <true|false>, "ad_text": <string or null>, "rung": <1-4 or null>, "rationale": <string>}. Use emit=false with ad_text=null when the moment should not be described; give the reason for suppressing in rationale.
\end{lstlisting}
\end{promptbox}

\subsection{Training and inference user message (schema)}
\label{app:prompt-user}
The following function builds the user message of every training example from the moment context; it is shown as source because the message is assembled per moment.
\begin{promptbox}{Prompt B.37: Training and inference user message (schema)}
\begin{lstlisting}[style=promptbox]
def _user_text(row: dict[str, Any], text_only: bool, screen_hint: bool = False) -> str:
    lines = [f"Moment type: {row.get('type')}"]
    lines.append(f"Lecturer said (+/-8s): {(row.get('transcript_window') or '').strip()}")
    lines.append(f"On-screen text (OCR): {(row.get('slide_ocr') or '').strip()}")
    # text_only: the annotated visual description stands in as the visual proxy.
    # screen_hint: keep it ALONGSIDE the keyframes (vision as additive signal, not a replacement --
    # disentangles "raw pixels vs no vision" from "raw pixels vs oracle summary").
    if text_only or screen_hint:
        lines.append(f"What is on screen: {(row.get('what_on_screen') or '').strip()}")
    pause = row.get("pause_after")
    lines.append(f"Pause available after the moment: {pause if pause is not None else 0}s")
    return "\n".join(lines)
\end{lstlisting}
\end{promptbox}

\section{Example Moments}
\label{app:moments}
\begin{figure*}[p]
\centering
\pdftooltip{\includegraphics[width=\textwidth]{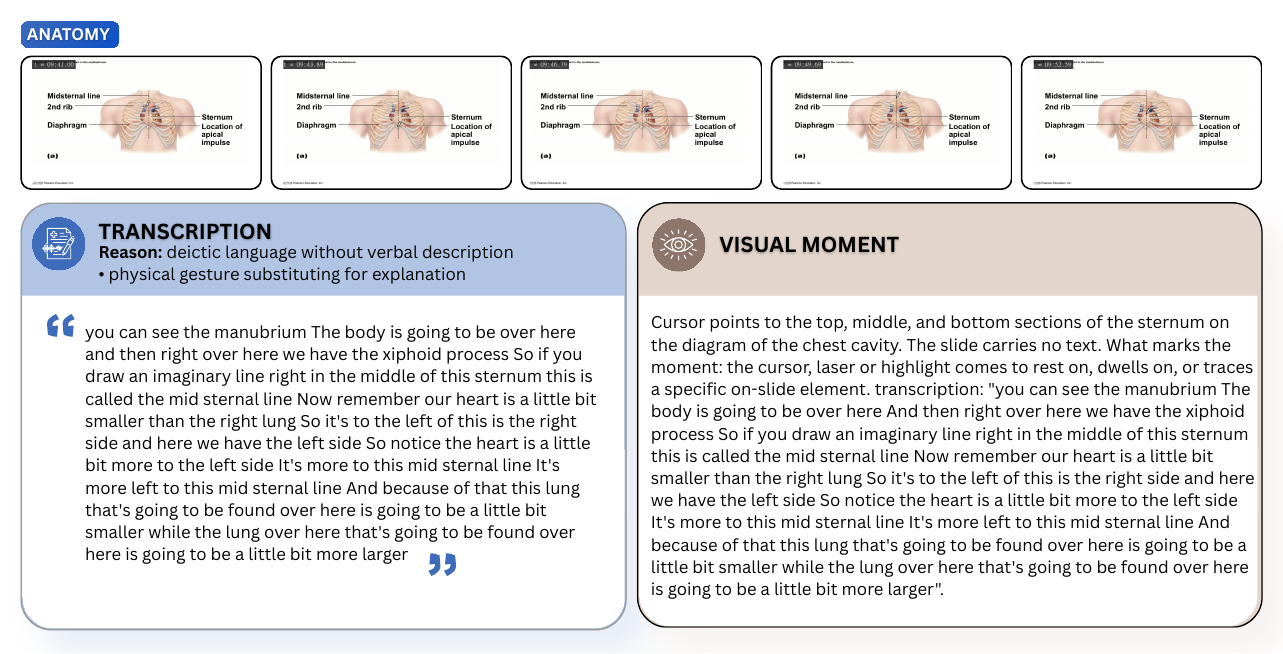}}{Five stills of an anatomy slide showing a labeled diagram of the chest cavity, with a cursor pointing to the top, middle, and bottom of the sternum. Below, the transcript window with the classifier's reason, and the on-screen description.}
\caption{\textbf{A pointing moment from an anatomy lecture (LPM).} On a labeled diagram of the chest cavity, the cursor points in turn to the top, middle, and bottom of the sternum while the lecturer names the manubrium, the body, and the xiphoid process and then contrasts the two lungs. The parts are named, but their positions are given only as ``over here'', and the two lungs only as ``this lung'' and ``the lung over here'', so a listener needs the visual channel to know which side is meant. The slide is reproduced from the lecture recording; its copyright remains with its publisher.}
\label{fig:moment-anatomy-1}
\end{figure*}

\begin{figure*}[p]
\centering
\pdftooltip{\includegraphics[width=\textwidth]{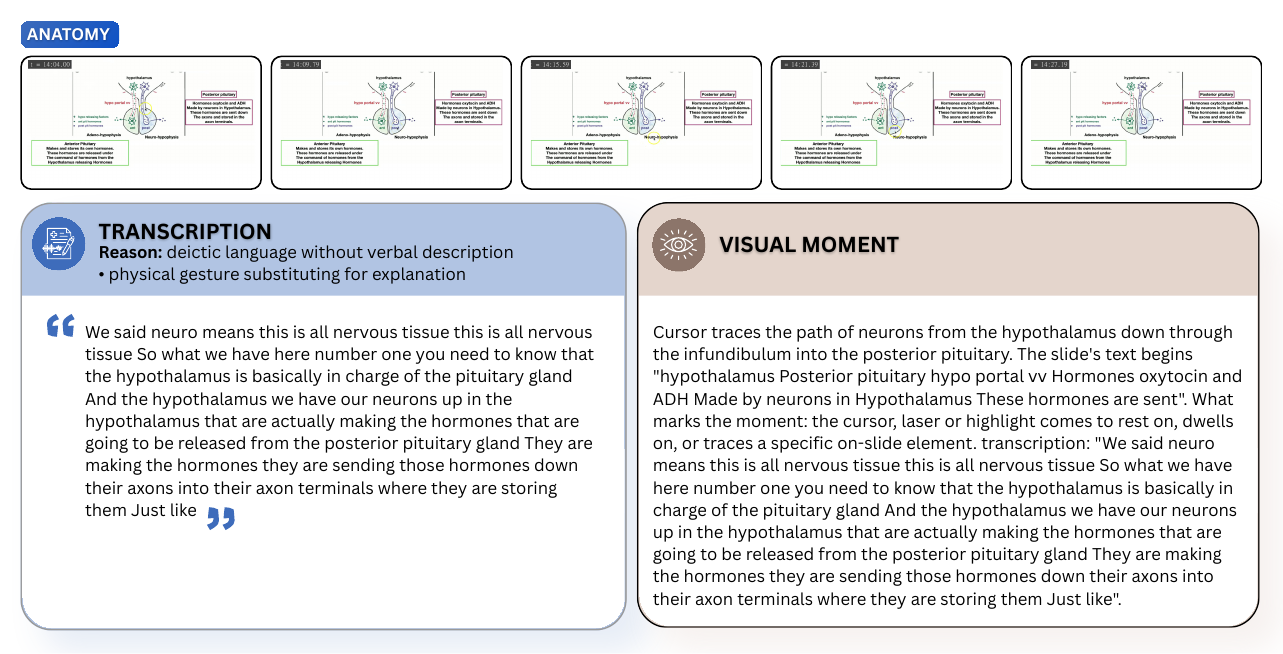}}{Five stills of an anatomy slide showing a diagram of the hypothalamus and the posterior pituitary, with a cursor moving down the connecting stalk. Below, the transcript window with the classifier's reason, and the on-screen description.}
\caption{\textbf{A pointing moment from an anatomy lecture (LPM).} On a diagram of the hypothalamus and the pituitary gland, the cursor traces the path of the neurons from the hypothalamus down the infundibulum into the posterior pituitary. The lecturer speaks of hormones being sent ``down their axons'' without naming the structures the cursor passes, so the path is available only visually. The slide is reproduced from the lecture recording; its copyright remains with its publisher.}
\label{fig:moment-anatomy-2}
\end{figure*}

\begin{figure*}[p]
\centering
\pdftooltip{\includegraphics[width=\textwidth]{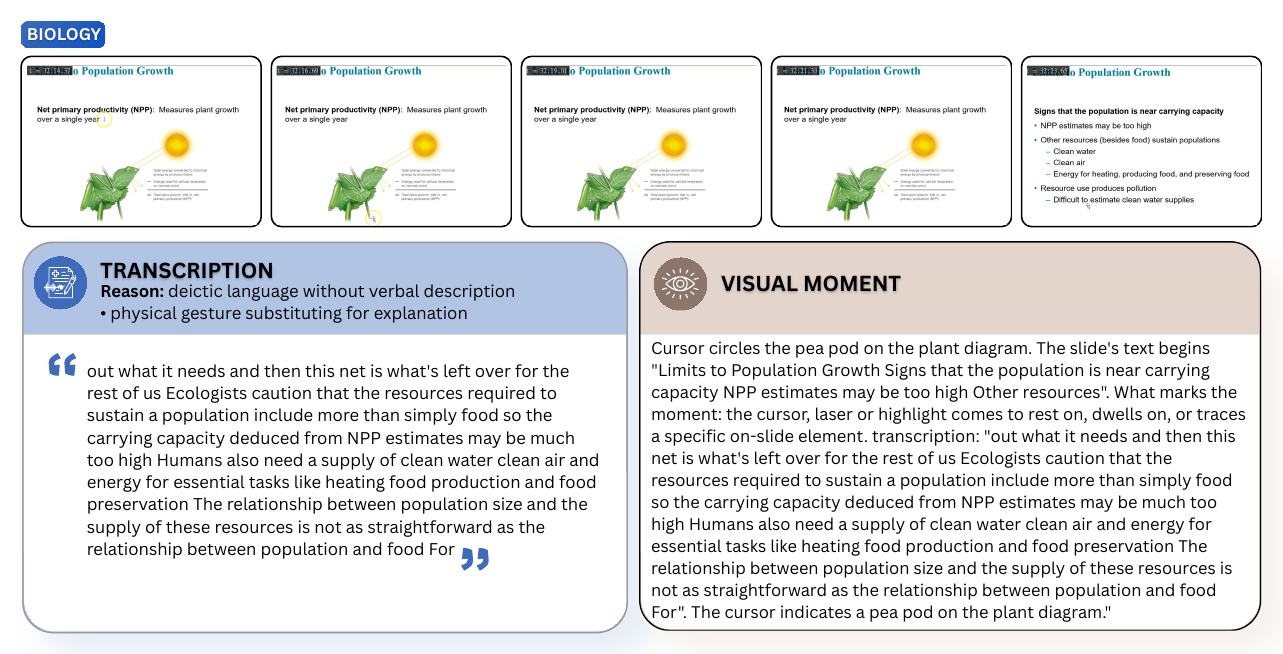}}{Five stills of a biology slide titled Population Growth, with a plant diagram and a sun icon and a cursor circling a pea pod on the plant; the last still shows the next slide, a bulleted list. Below, the transcript window with the classifier's reason, and the on-screen description.}
\caption{\textbf{A pointing moment from a biology lecture (LPM).} On a slide about limits to population growth, the cursor circles the pea pod on the plant diagram while the lecturer explains that net primary productivity overstates carrying capacity. The speech never names what is circled, so a listener needs the visual channel to know which part of the diagram is meant. The slide is reproduced from the lecture recording; its copyright remains with its publisher.}
\label{fig:moment-biology}
\end{figure*}

\begin{figure*}[p]
\centering
\pdftooltip{\includegraphics[width=\textwidth]{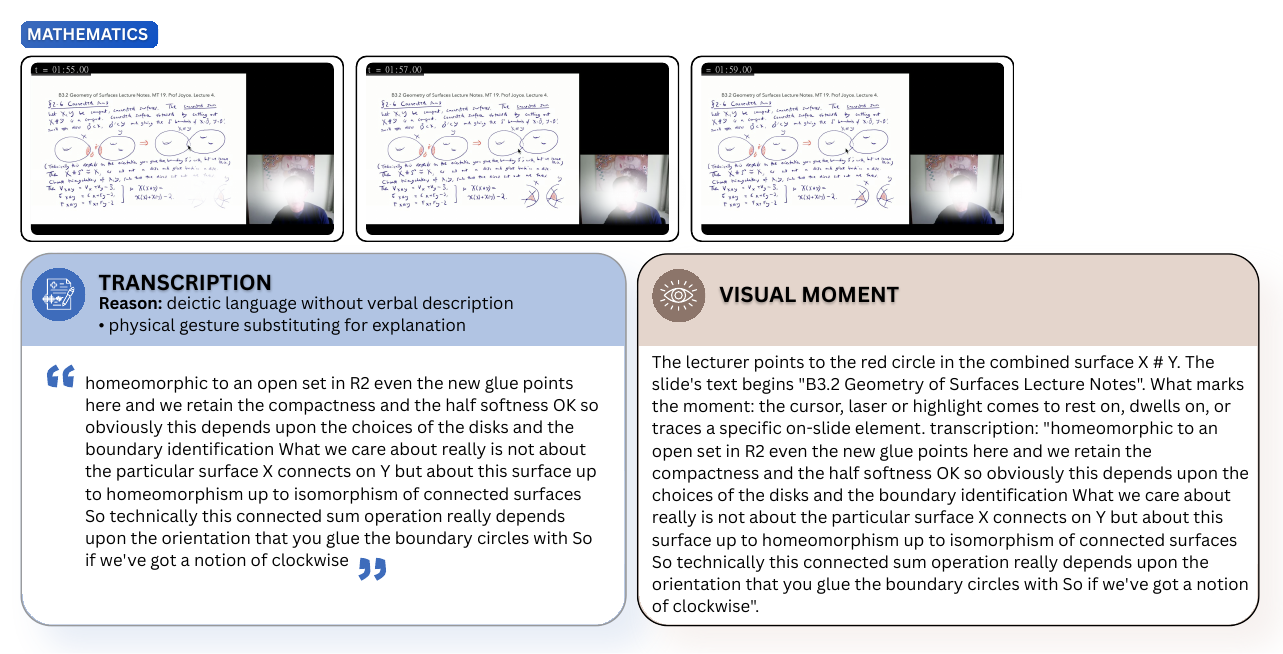}}{Three stills of handwritten lecture notes on the geometry of surfaces, showing two surfaces glued into a connected sum with a red circle at the glued boundary, and a lecturer webcam inset. Below, the transcript window with the classifier's reason, and the on-screen description.}
\caption{\textbf{A pointing moment from a mathematics lecture (M3AV).} On handwritten notes on the geometry of surfaces, the lecturer points to the red circle marking the glued boundary of the connected sum $X\#Y$ while explaining that the operation depends on the orientation with which the boundary circles are glued. The phrase ``the new glue points here'' is resolvable only from the pointing. Three stills are shown because the gap between the triggering keyframes was at most 3\,s (Appendix~\ref{app:training}). The notes are reproduced from the lecture recording; their copyright remains with their publisher.}
\label{fig:moment-math-1}
\end{figure*}

\begin{figure*}[p]
\centering
\pdftooltip{\includegraphics[width=\textwidth]{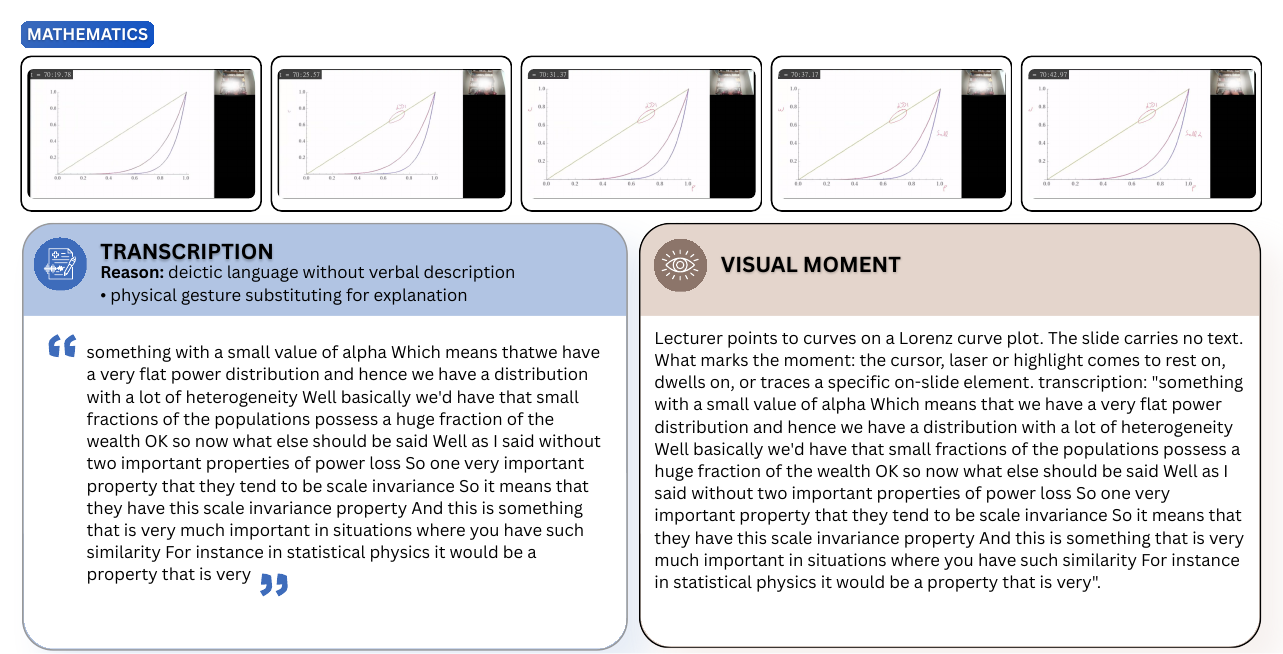}}{Five stills of a plot of Lorenz curves with no text on the slide, a lecturer webcam inset, and handwritten marks added near the curves. Below, the transcript window with the classifier's reason, and the on-screen description.}
\caption{\textbf{A pointing moment from a mathematics lecture (M3AV).} On a plot of Lorenz curves that carries no text, the lecturer points to the curves while explaining that a small exponent $\alpha$ yields a very flat power-law distribution, in which a small fraction of the population holds a large fraction of the wealth. Because the slide has no text, neither the OCR nor the speech says which curve corresponds to which case; only the pointing does. The plot is reproduced from the lecture recording; its copyright remains with its publisher.}
\label{fig:moment-math-2}
\end{figure*}

\begin{figure*}[p]
\centering
\pdftooltip{\includegraphics[width=\textwidth]{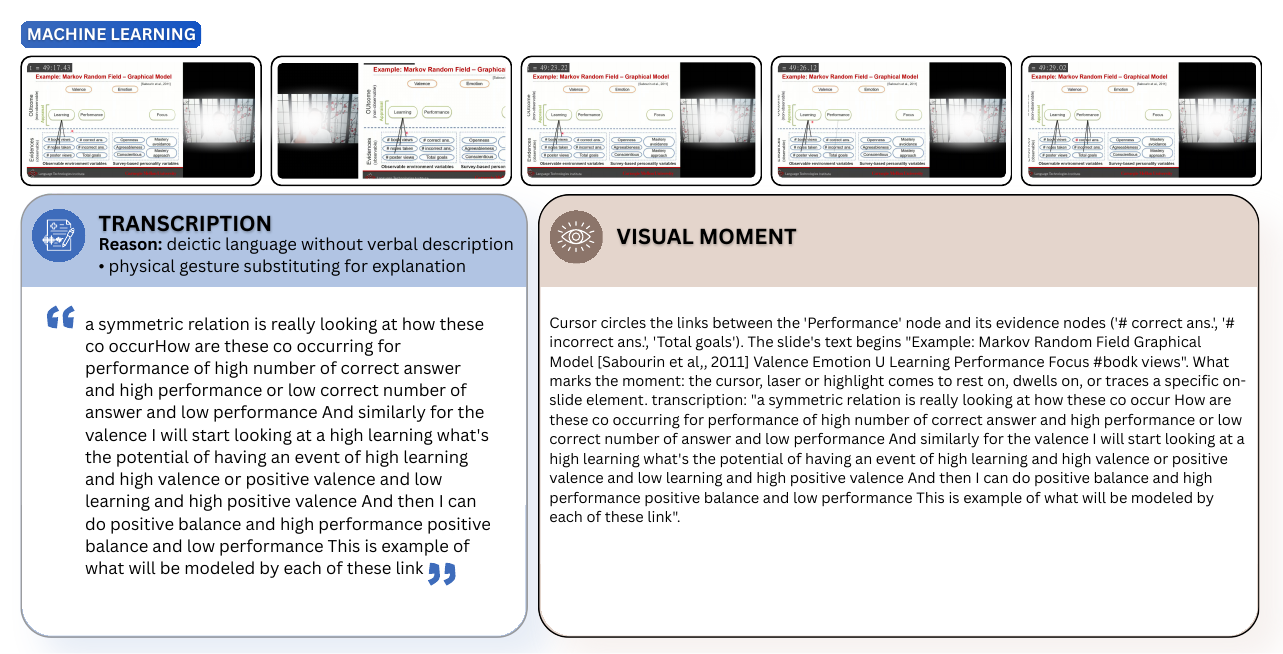}}{Five stills of a machine-learning slide showing a Markov random field graphical model with a lecturer webcam inset, and a cursor circling the links between a Performance node and three evidence nodes. Below, the transcript window with the classifier's reason, and the on-screen description.}
\caption{\textbf{A pointing moment from a machine-learning lecture (LPM).} On a slide showing a Markov random field over learning, performance, and emotion variables and their observable evidence, the cursor circles the links between the \emph{Performance} node and its evidence nodes (correct answers, incorrect answers, and total goals) while the lecturer explains what each of these links models. Which links are meant is shown only by the cursor. The slide is reproduced from the lecture recording; its copyright remains with its publisher.}
\label{fig:moment-ml}
\end{figure*}

Figures~\ref{fig:moment-anatomy-1}--\ref{fig:moment-ml} show six retained moments from four subjects, four from the LPM collection (anatomy, biology, and machine learning) and two from M3AV (mathematics). Each figure gives the stills the generators received, the transcript window with the classifier's reason for retaining the candidate, and the on-screen description used in the text-only setting (Section~\ref{sec:models}). All six are pointing moments, the most frequent type in the corpus, in which the lecturer's deictic speech leaves the referent to the visual channel.

\section{Corpus and Training Details}
\label{app:training}

This appendix reports how the corpus was built, how the models were trained,
and which models act as generators and judges, underlying the post-training
study of Section~\ref{sec:res-training}. 

\subsection{Corpus}

\textbf{Moment Detection and Classification.} The moment inventory was built in
two passes. A deterministic detector first proposed candidate visual events
from artifacts already computed: content words added between consecutive OCR
frames, OCR turnover below a Jaccard threshold, mean pixel difference where OCR
is stable, and contiguous voice-activity silence. Between them these channels
registered that something on the slide changed or that the lecturer paused, and
they were tuned for recall, since a candidate the detector never proposed could not
be recovered downstream. A VLM pass (Gemini 3.1 Pro) then examined each
candidate's frames and decided whether it was a genuine pedagogical event,
assigning one of twelve moment types, slide, transition, reveal, pointing,
ink, figure, chart, table, math, code, animation, other, or rejecting it. It was this second pass that fixed which candidates became moments; the rejected
candidates were overwhelmingly ones whose frames showed no change at all, such as
a static slide behind a moving webcam inset. The classifier
worked from the moment's stills and a transcript slice and did not see the
standard, so it judged what kind of teaching event is on screen rather than whether that event needs describing; the latter was decided at generation time,
when each model either wrote a description or stayed silent under the 45-rule
standard. Appendix~\ref{app:moments} shows six retained moments.

\textbf{Generation Context.} A moment's stills were sampled across a window
derived from the gap between the two keyframes that triggered the candidate:
three stills when that gap was at most 3\,s and five when it was wider,
with at least one sampling interval falling after the event. The transcript
window opened 5\,s before the moment and closed 8\,s after it
for 2,534 reference rows and at the next pedagogical event, or 45\,s,
whichever came first, for the remaining 4,334; each reference row records
which window produced it.

\textbf{Subject Coverage.} The corpus covers six science, medicine, and psychology subjects from the LPM and M3AV collections. The system evaluation rests on 17 lectures from three subjects (machine learning, dentistry, and mathematics), and the rating study on 27 lectures from four subjects (machine learning, anatomy, psychology, and biology), of which machine learning is the only subject shared with the system evaluation; both sets are more pointing-heavy than the training side (77\% and 72\% of moments, against 45\%).

\textbf{Preference-Pair Generators.} Candidates for the natural pairs were
generated by four VLMs (Gemini 3.1 Pro, GPT-5.5, Claude Sonnet 4.6 and
Qwen3-VL-235B) and the controlled pairs by Gemini 3.1 Pro alone.

\subsection{Training}

\textbf{Backbones and SFT.} All seven backbones are open-weight,
instruction-tuned VLMs: SmolVLM2-2.2B \citep{marafioti2025smolvlm},
Qwen3-VL-2B, Qwen3-VL-4B, Qwen3-VL-8B \citep{bai2025qwen3vl},
InternVL3.5-4B, InternVL3.5-8B \citep{wang2025internvl35}, and Gemma-3-4B
\citep{gemma2025gemma3}. Each model was fine-tuned in two arms, a multimodal
arm receiving the keyframes and a text-only arm in which an annotated
description of the screen replaces them (Prompt~B.37), using LoRA
\citep{hu2022lora} with rank 16, scaling factor 32, and dropout 0.05 on the
attention projections. Training ran for three epochs at a learning rate of
$10^{-4}$ with an effective batch size of 8, under three seeds (0, 1, 2); a
moment-disjoint 10\% development slice of the 4,963-instance training set
served for checkpoint selection. The chrF, BERTScore, CLIPScore and the two
\textsc{AD4Edu-Eval} columns of Table~\ref{tab:systems} are averages over the
three seeds.

\textbf{Inference.} All systems were evaluated under forced emission, so that every table compares descriptions rather than decisions to stay silent.
Generation ran through Hugging Face \texttt{model.generate} in half precision on one GPU, with the assistant turn prefilled so the model
continued a description rather than choosing whether to write one. The first attempt was greedy at a maximum of 160 new tokens; if its output did not parse, up to two sampled retries followed at temperature 0.7, top-$p$ 0.9 and a 256-token budget, each passed through the same salvage chain.
Forcing was therefore best-effort: a moment could still return no parsable
description, as five did on the DPO row of Table~\ref{tab:rubric}.

\textbf{DPO Configuration.} Post-training ran for two epochs at effective
batch size 8 over the 3,052 preference pairs; a moment-disjoint 10\%
development slice left 2,742 training pairs and 686 optimizer steps at a
learning rate of $1.5\times10^{-5}$ with $\beta=0.1$. DPO was trained for all
seven text-only rows but only two multimodal rows: a DPO step forwards the
chosen and rejected sequences together, and the remaining multimodal cells
exceeded the 24\,GB GPU, a limit we could not lift without changing the model
input and so making a cell incomparable with its own text-only twin.

\textbf{Hardware.} All runs were trained on one shared node with eight NVIDIA
RTX 3090 GPUs (24\,GB each), one GPU per run and no scheduler, so the
wall-clock figures below are lower bounds taken while other jobs shared the
node. Table~\ref{tab:wallclock} lists the measured cells: SFT took between one
and fourteen hours depending on backbone and arm, DPO between 43 minutes and
four hours. The 8B rows exceeded the card while sharing it and were trained
afterwards on an emptied card, so their SFT wall-clock was not recorded; their
DPO cells were trained on the emptied card and are listed.

\begin{table}[t]
\centering
\caption{\textbf{Measured training cost per cell} on one RTX 3090 (24\,GB) of a shared node, measured on seed 0 only. SFT cells are 1,674 steps (three epochs over 4,462 training rows at effective batch size 8, after the 501-row development slice); DPO cells are 686 steps (two epochs over 2,742 training pairs at the same effective batch size). Peak VRAM is the training process's memory sampled every 30\,s, a lower bound on the true peak; it was sampled on the SFT runs only, hence the dashes on the DPO rows. Wall-clock was measured while other jobs used the node.}
\footnotesize
\setlength{\tabcolsep}{4pt}
\resizebox{\columnwidth}{!}{
\begin{tabular}{llrrr}
\toprule
Backbone & Arm & Steps & Peak VRAM & Wall-clock \\
\specialrule{\lightrulewidth}{\aboverulesep}{0pt}
\rowcolor{gray!20}[1pt][1pt]
\multicolumn{5}{c}{\emph{SFT, text-only}} \\
SmolVLM2-2.2B & text & 1,674 & 5.5\,GiB & 66 min \\
Qwen3-VL-2B & text & 1,674 & 7.0\,GiB & 102 min \\
Qwen3-VL-4B & text & 1,674 & 12.0\,GiB & 199 min \\
InternVL3.5-4B & text & 1,674 & 12.6\,GiB & 195 min \\
Gemma-3-4B & text & 1,674 & 13.8\,GiB & 202 min \\
\rowcolor{gray!20}[1pt][1pt]
\multicolumn{5}{c}{\emph{SFT, multimodal}} \\
SmolVLM2-2.2B & mm & 1,674 & 6.0\,GiB & 172 min \\
Qwen3-VL-2B & mm & 1,674 & 7.7\,GiB & 416 min \\
Qwen3-VL-4B & mm & 1,674 & 13.1\,GiB & 860 min \\
InternVL3.5-4B & mm & 1,674 & 12.8\,GiB & 417 min \\
Gemma-3-4B & mm & 1,674 & 15.9\,GiB & 750 min \\
\rowcolor{gray!20}[1pt][1pt]
\multicolumn{5}{c}{\emph{DPO, text-only}} \\
SmolVLM2-2.2B & text & 686 & --- & 43 min \\
Qwen3-VL-2B & text & 686 & --- & 88 min \\
Qwen3-VL-4B & text & 686 & --- & 99 min \\
InternVL3.5-4B & text & 686 & --- & 96 min \\
Gemma-3-4B & text & 686 & --- & 98 min \\
InternVL3.5-8B & text & 686 & --- & 154 min \\
Qwen3-VL-8B & text & 686 & --- & 150 min \\
\rowcolor{gray!20}[1pt][1pt]
\multicolumn{5}{c}{\emph{DPO, multimodal}} \\
SmolVLM2-2.2B & mm & 686 & --- & 97 min \\
Qwen3-VL-2B & mm & 686 & --- & 236 min \\
\bottomrule
\end{tabular}
}

\label{tab:wallclock}
\end{table}

\textbf{DPO Convergence Dynamics.} Two development sets were in play: the
488-pair lecture-disjoint development set, removed from training and used to select
$\beta$ and the learning rate, and a further in-loop 10\% moment-disjoint slice
of the remainder, which the trainer evaluated during the run. Preference
accuracy on the in-loop slice peaked at step 400 of 686 and receded thereafter,
while the implicit
reward margin continued to rise across the whole run, a pattern consistent
with accuracy carried by pairs whose preferred side already appears among the
SFT targets.

\subsection{Judges and Evaluation}


\textbf{Judges.} LLM-AD-Eval in Table~\ref{tab:systems} uses the original judges of \citet{han2024autoad3}, GPT-3.5 Turbo and Llama-2-7B-Chat. Table~\ref{tab:human} and Appendix~\ref{app:robustness} use GPT-5.5, which also generated part of the reference set and serves as the judge for \textsc{AD4Edu-Eval}.

\textbf{Reference-Writer Compliance.} Table~\ref{tab:refwriters} scores the three reference
writers' own descriptions under both offline modes of \textsc{AD4Edu-Eval}. All three lie between
0.81 and 0.82 in the mechanical mode and between 0.80 and 0.82 in the local mode, against 0.532 and
0.501 for the slide-title readout. The best fine-tuned row of Table~\ref{tab:systems} reaches 0.858,
but on all 516 moments under forced emission rather than on a writer's self-selected subset, so the two
figures are not directly comparable.

\begin{table}[t]
\centering
\caption{\textbf{Compliance of the reference writers} under \textsc{AD4Edu-Eval}, both offline
modes. Unlike the rows of Table~\ref{tab:systems}, which are generated under forced emission on all 516 held-out
moments, a reference writer was free to stay silent, so $n$ is the number of moments that writer
chose to describe and each figure is compliance on that writer's own subset. The last two rows are
repeated from Table~\ref{tab:systems} for scale and are scored on all 516.}
\footnotesize
\setlength{\tabcolsep}{5pt}
\begin{tabular}{lrcc}
\toprule
Reference writer & $n$ & mech & local \\
\midrule
GPT-5.5 & 424 & 0.824 & 0.818 \\
Gemini 3.1 Pro & 401 & 0.813 & 0.806 \\
Claude Sonnet 4.6 & 339 & 0.811 & 0.796 \\
\midrule
Best trained row & 516 & 0.858 & 0.838 \\
Slide-title readout & 516 & 0.532 & 0.501 \\
\bottomrule
\end{tabular}

\label{tab:refwriters}
\end{table}

\textbf{Compliance Prompts.} Forty-two of the 45 rules carry a judge template.
The three without one are \texttt{visual\_content\_001}, which governs the emit
decision rather than the wording of a description;
\texttt{pair\_perceptibility\_001}, which constrains how a preference pair is built; and \texttt{slide\_text\_003}, which is enforced by the novelty term of Eq.~\ref{eq:ad4edu}.

\textbf{Statistical Power.} The raters' use of the equal option on controlled pairs ranges from 3\% to 62\%, leaving 93 to 343 decided pairs per rater and metric. At that size, agreement is distinguishable from chance at 80\% power only above 57--63\%, and paired McNemar tests could detect differences of 10.8--15.3~pp between LLM-AD-Eval or the rubric and chrF and of 6.5--9.2~pp between the fitted head and LLM-AD-Eval. Eleven of the twelve per-rater gaps we tested lie below these thresholds; only the twelfth, chrF over the rubric for R2, reaches its threshold and is significant after Holm correction.
\textbf{Held-Out Comparison.} The one-shot comparison on the held-out rater
reports a 4.9~pp discordant-pair effect, four of the 82 pairs both scorers decide, against a 10.2~pp minimum detectable effect at that denominator.

\textbf{Fitted Preference Head.} The head is a logistic regression over the differences of per-side metric features between the two descriptions of a pair, scored antisymmetrically: the decision is the sign of the weighted feature difference with no intercept, so swapping the sides flips the verdict by construction. It was fitted on the labels of R1, R2 and R4 with lecture-disjoint cross-validation and consensus weighting (a label's weight is the number of raters who chose that side). Features whose weight sign was unstable across raters were dropped (the rubric's deixis and faithfulness subscores), and the weights were frozen on 2026-09-09, before R3's labels were collected. Six features remain: chrF-to-references ($+0.78$), LLM-AD-Eval ($+0.81$), the rubric's overall score ($+0.73$), its terminology subscore ($+1.54$), its non-redundancy subscore ($-1.06$), and word count ($+3.36$). The fitted-head row of Table~\ref{tab:human} is out-of-fold for R1, R2 and R4 and a one-shot test with frozen weights for R3.

\section{Worked Examples}
\label{app:examples}
This appendix shows, concretely, what the controlled-pair construction described in
Section~\ref{sec:standard} produces: one worked example per rule category, drawn from the labeled
validation pool, followed by one rule from the 45-rule standard (Appendix~\ref{app:rules})
reproduced in full, with its citation, scoring route, and BLV review verdict. The candidate texts
quoted below are reproduced verbatim from the source pool; nothing has been paraphrased, corrected, or shortened. Pair and moment identifiers are withheld from this anonymized version because they encode the source video id; each example is instead labeled by
rule category and a running number, and the real pair and moment identifiers are recorded in the
audit file accompanying this appendix.

\subsection*{One controlled pair per rule category}

Each pair below belongs to the ``controlled'' stratum of the validation pool: a compliant
description and a single deliberately introduced defect on one rule, with everything else held
fixed, so that a preference between the two isolates that one rule. Six examples follow, one per
category: style, terminology, length, deixis, faithfulness, non-redundancy.

\paragraph{Example 1 --- Style.} \textit{Moment: an anatomy-lecture figure, a bone shown from a
single viewing angle.}
\begin{quote}\small
\textbf{Compliant.} A superior view of the right clavicle.\\[2pt]
\textbf{Violates style.} The slide shows a superior view of the right clavicle.
\end{quote}
The non-compliant side opens with the announcing phrase ``The slide shows,'' which this pair's own
construction note records as forbidden meta-language; \texttt{cross\_cutting\_002}'s mechanical
check flags exactly this framing, though the rule's own text states its requirement as present
tense, active voice, and third-person narration rather than naming meta-language directly. The
compliant side narrates the visual directly, satisfying that text.

\paragraph{Example 2 --- Terminology.} \textit{Moment: a machine-learning lecture, the lecturer
points to one labeled block of a neural-network architecture diagram.}
\begin{quote}\small
\textbf{Compliant.} The lecturer points to the Neural network boxes grouped under Layer 2.\\[2pt]
\textbf{Violates terminology.} The lecturer points to the Neural network boxes grouped under
Layer 1.
\end{quote}
The non-compliant side substitutes the label of a different layer, also present on the slide, for
the one the lecturer is actually pointing to; the compliant side names the layer actually
indicated, consistent with \texttt{cross\_cutting\_004}'s requirement to keep terminology
consistent with what is being referred to, rather than borrowing a label from elsewhere on the same
slide.

\paragraph{Example 3 --- Length.} \textit{Moment: a machine-learning lecture, the lecturer points
to one symbol inside a kernel-function equation.}
\begin{quote}\small
\textbf{Compliant.} The lecturer points to phi of x sub i.\\[2pt]
\textbf{Violates length.} The lecturer points to phi of x sub i, which is part of the equation K
of x sub i comma x sub j equals phi of x sub i transpose phi of x sub j, which equals the inner
product of phi of x sub i and phi of x sub j, where phi maps D to Z.
\end{quote}
The non-compliant side unpacks the entire equation the symbol belongs to, adding no information
beyond what the symbol itself conveys, and overruns the budget of the reachable delivery rung; the
compliant side stays within it. The mechanical check bound to this defect is
\texttt{length\_004}'s \texttt{length\_budget}: it computes the reachable rung's word budget
directly from the pause length and scores the AD's own word count against that budget, independent
of any declared rung --- the mechanism this pair's own difference note names (``B overruns the
reachable rung's budget''). \texttt{length\_003}'s registered check instead scores a
\emph{declared} rung field against feasibility; no row in this pool declares one, so it cannot
distinguish this pair at all. (As with the style example above, \texttt{length\_004}'s own recorded
rule text states a flat 25-word/100-word default, not the rung-derived budget its registered check
actually computes; the check, not the prose, is what scores this difference.)

\paragraph{Example 4 --- Deixis.} \textit{Moment: a biology lecture, the lecturer points to a
labeled organelle in a cell diagram.}
\begin{quote}\small
\textbf{Compliant.} The lecturer points to the centrioles.\\[2pt]
\textbf{Violates deixis.} The cursor moves to them.
\end{quote}
The non-compliant side narrates the cursor's motion and leaves the referent to be inferred; the
compliant side names the referent instead of the act of pointing, which is \texttt{deixis\_001}'s
defining requirement.

\paragraph{Example 5 --- Faithfulness.} \textit{Moment: an anatomy lecture, a microscopy or
cell-diagram figure with one stained structure at its center.}
\begin{quote}\small
\textbf{Compliant.} The large purple structure in the center of the diagram is the nucleus.\\[2pt]
\textbf{Violates faithfulness.} The large purple structure, which measures exactly five
micrometers across, is the nucleus.
\end{quote}
The non-compliant side adds a precise measurement the keyframe does not support; the compliant side
makes no claim beyond what is visible, following \texttt{cross\_cutting\_005}'s preference for
silence over a confident wrong description.

\paragraph{Example 6 --- Non-redundancy.} \textit{Moment: a machine-learning lecture, the lecturer
points to a node in a decision-tree diagram tied to a value-function equation.}
\begin{quote}\small
\textbf{Compliant.} The lecturer points to the V star of s prime node at the bottom left of the
tree.\\[2pt]
\textbf{Violates non-redundancy.} The lecturer points to the V star of s prime node representing
the optimal state value function for the next state.
\end{quote}
The non-compliant side restates the node's definition, which the lecturer's speech already
supplies; the compliant side instead adds the node's on-screen location, information the speech
does not carry, as the redundancy gate \texttt{cross\_cutting\_001} requires.

\subsection*{A fully recorded rule}

Table~\ref{tab:rule-math-typeset-003} reproduces every field recorded against
one rule, \texttt{math\_typeset\_003}, in the standard: its
text, category and DOC/PROV tag, citation and derivation basis, the scoring
route the training code reads, and the BLV review verdict from the
single-rater review of build \texttt{6424ab12}. This is one of only 13 rules
in the 45-rule standard carrying a recorded BLV review block, and the only
one of those 13 whose citation field is not null.

\begin{table*}[h]
\small
\centering
\caption{Every field recorded against rule \texttt{math\_typeset\_003} in the standard, reproduced verbatim.}
\begin{tabular}{@{}p{0.24\linewidth}p{0.66\linewidth}@{}}
\toprule
\textbf{Field} & \textbf{Value} \\
\midrule
Rule id & \texttt{math\_typeset\_003} \\
Section & 2.MathTypeset \\
Element & math\_typeset \\
Type (category) & how \\
Tag & PROV (provisional) \\
Basis & transfer-from (ClearSpeak is a documented rule set; the choice of it over MathSpeak is a
design decision) \\
Rule text & Use ClearSpeak as the default verbalization grammar, applied consistently within a
course; defer to learner preference where the learner has one. (ClearSpeak chosen over MathSpeak
because it reads closer to natural spoken mathematics; no universal standard exists.) \\
Max seconds & --- (null) \\
Citation & No universal spoken-math standard exists (NFB, Guidelines for Collegiate Faculty to
Teach Mathematics to Blind or Visually Impaired Students). Frankel, Brownstein, Soiffer \& Hansen
(2016), ETS Research Report RR-16-23 --- supporting (grade C): documents ClearSpeak as a defined
rule set, not a universal standard. \\
Bucket & needs\_users \\
Validation & blv\_user\_study \\
Scoring route & judge \\
Scoring checks & [] (empty; the judge route scores against the rule text directly, not a named
mechanical check) \\
Scoring basis & blv\_unresolved \\
Unscored reason & --- (null; route is not \texttt{unscored}) \\
\midrule
\multicolumn{2}{@{}p{0.9\linewidth}@{}}{\textit{BLV review --- build \texttt{6424ab12}, rules SHA
\texttt{42804e05d3ea}, $n=1$ reviewer}} \\
\midrule
Review question & ``Is this rule right?'' \\
BLV rule reviewer verdict & cannot\_tell \\
BLV rule reviewer note & ``would need an example.'' \\
\bottomrule
\end{tabular}

\label{tab:rule-math-typeset-003}
\end{table*}

\section{Metric Robustness, Stability, and Per-Category Analyses}
\label{app:robustness}

We audit whether the recomputable evaluation modes are sensitive to
surface changes that should not materially affect description quality.
This analysis serves as an implementation audit and regression test,
rather than as an independent estimate of construct validity.

\paragraph{Protocol.}
We applied four deterministic, guarded transformations to each eligible
description: whitespace normalization, ASCII/Unicode punctuation
substitution, en-GB/en-US spelling substitution, and digit/word numeral
substitution. Transformations that leave the description unchanged are
excluded from the corresponding denominator. For each recomputable
scorer, we measured (i) score sensitivity, defined as an absolute score
change greater than 0.01 on the scorer's normalized scale, and
(ii) decision inconsistency, defined as a change in the preferred
candidate after transforming one or both descriptions in a pair.

For scorers that produce pairwise decisions, we additionally reversed
candidate presentation order and mapped the resulting verdict back to
candidate identity. A scorer is inconsistent when the selected
candidate changes after this mapping. Cached LLM-AD-Eval and rubric
scores cannot be recomputed on transformed ADs and are
therefore excluded from the text-transformation audit. Reordering two
previously cached pointwise scores would test only the comparison code,
not presentation-order sensitivity of the underlying judge.

\paragraph{Results.}
Across 6,647 defined score comparisons, 170 (2.6\%) changed by more
than 0.01. Across 4,390 defined pairwise comparisons under text
transformations, 12 (0.3\%) changed the preferred candidate. Most score
sensitivity occurred under orthographic and numeral transformations,
primarily because chrF depends directly on character overlap.


\paragraph{Scope.}
The audit measures implementation stability under a small set of
controlled surface transformations. It does not establish robustness
under arbitrary paraphrases or semantic changes. Text-level robustness
of the cached LLM judges also remains unmeasured because those judges
were not rerun on transformed inputs.

\paragraph{Judge Choice and Self-Preference.}
GPT-5.5 wrote roughly a third of the reference ADs it grades against, so we
re-scored the study pairs with the two judges of Table~\ref{tab:systems}, neither
of which wrote a reference. On the 250 controlled pairs GPT-3.5 Turbo prefers the
compliant side on 71.3\% [62, 79] of the 108 pairs it decides, against 84.3\%
[75, 91] of the 83 GPT-5.5 decides; on the 46 pairs both decide they agree on
93.5\%, and the discordance is 3--0 in GPT-5.5's favor (exact McNemar
$p=0.25$). The two rank the six categories alike, localizing deixis (100\% and
89\%) and terminology (95\% and 88\%) and falling below chance on length (27\%
and 24\%). Scored the same way under both judges, per-rater agreement on the 400
pairs moves by 1.7 to 4.2~pp, up on two raters and down on two, with every
interval overlapping, so no claim of Section~\ref{sec:res-human} turns on the
choice. We kept GPT-5.5 as the judge of record for the study pairs
and report this as a robustness check rather than a replacement. Llama-2-7B-Chat
cannot serve at pair level at all: it rates 94\% of sides a 4, so it decides only
8 of the 250 controlled pairs and 6 to 8 non-tie labels per rater.

\paragraph{Ranking Stability.}
Because the compliance columns of Table~\ref{tab:systems} are used to rank systems, Table~\ref{tab:rank-stability} reports how that ranking depends on the novelty threshold $\kappa$ of Eq.~\ref{eq:ad4edu}. For $\kappa\in\{0.2,0.3,0.4,0.5,0.6\}$ it gives the Spearman correlation of the resulting ordering of the 23 trained checkpoints (the 18 trained rows of Table~\ref{tab:systems} and the five multimodal SFT checkpoints without a DPO counterpart) with the ordering at $\kappa=0.4$, the rank of the slide-title readout among these 24 systems, and the number of DPO rows that fall below their SFT precursor. The ordering is stable ($\rho\ge0.98$ in both modes), the readout is last at every $\kappa$, and the mechanical-mode count of DPO rows below SFT is 9 of 9 throughout, while the local-mode count moves between 7 and 8; no conclusion of Section~\ref{sec:results} depends on the choice of $\kappa$.

\begin{table}[t]
\centering
\caption{\textbf{Sensitivity of the compliance ranking to $\kappa$ (mechanical / local)}: Spearman correlation of the ordering of the 23 trained checkpoints with the ordering at $\kappa=0.4$, the slide-title readout's rank among these 24 systems, and the number of DPO rows below their SFT precursor; $\kappa=0.4$ is the value used throughout.}
\small
\setlength{\tabcolsep}{4pt}
\resizebox{0.9\columnwidth}{!}{%
\begin{tabular}{lccc}
\toprule
$\kappa$ & \begin{tabular}[b]{@{}c@{}}$\rho$ vs.\\$\kappa=0.4$\end{tabular} & \begin{tabular}[b]{@{}c@{}}Readout\\rank (of 24)\end{tabular} & \begin{tabular}[b]{@{}c@{}}DPO $<$ SFT\\(of 9)\end{tabular} \\
\midrule
0.2 & 0.992 / 0.987 & 24 / 24 & 9 / 7 \\
0.3 & 0.996 / 0.995 & 24 / 24 & 9 / 7 \\
0.4 & 1.00 / 1.00 & 24 / 24 & 9 / 7 \\
0.5 & 0.995 / 0.992 & 24 / 24 & 9 / 8 \\
0.6 & 0.990 / 0.982 & 24 / 24 & 9 / 8 \\
\bottomrule
\end{tabular}}

\label{tab:rank-stability}
\end{table}

\paragraph{Per-Category Localization.}
Table~\ref{tab:localization} breaks the localization figures of
Section~\ref{sec:res-metric} down by rule category. Given the differing category, the rubric finds
the defect on 97.6--100\% of pairs in every category. Pooled, without that oracle, five categories
hold above 85\% and non-redundancy falls to 62.1\%: deciding that a description repeats the
lecturer requires reading the transcript against the description, which the other five do not.
Counts are pairs on which the scorer decided; ties are excluded, which is why $n$ differs between
the two columns and across categories.

\paragraph{Per-Category Rater Agreement.}
Table~\ref{tab:percat-raters} breaks the rubric-subscore agreement of Section~\ref{sec:res-human}
down by rule category, pooled over the four raters on the controlled pairs. Agreement is highest
on deixis (89.1\%) and terminology (82.5\%), near chance on faithfulness, style, and
non-redundancy, and far below chance on length (16.5\%), where the raters prefer the longer side
that the standard penalizes. The spread from deixis to length is the evidence behind the layered reading of the standard in Section~\ref{sec:res-human}.

\begin{table*}[t]
\centering
\caption{\textbf{Per-category analysis on the controlled pairs}, which differ on exactly one rule. Left: rubric localization with the differing category given (``Given the category'') or to be identified along with the defect (``Pooled''); $n$ counts decided pairs. Right: agreement of the rubric subscores with the BLV raters, pooled over the four raters; each pair carries up to four rater labels, so the unit is the label, with pair counts alongside. Ties and tied subscores are excluded.}
\footnotesize
\setlength{\tabcolsep}{6pt}
\begin{tabular}{lccccc}
\toprule
 & \multicolumn{2}{c}{Rubric localization (\%, $n$)} & \multicolumn{3}{c}{Rubric-subscore agreement with BLV raters} \\
\cmidrule(lr){2-3}\cmidrule(l){4-6}
Rule category & Given the category & Pooled & Agreement (\%) & 95\% CI & labels / pairs \\
\midrule
style          & 100.0 (34) & 100.0 (29) & 56.9 & [45, 68] & 65 / 33 \\
terminology    & 100.0 (32) & 100.0 (31) & 82.5 & [74, 89] & 103 / 32 \\
length         & 100.0 (34) & 89.3 (28)  & 16.5 & [10, 25] & 97 / 34 \\
deixis         & 97.6 (41)  & 91.4 (35)  & 89.1 & [83, 93] & 129 / 41 \\
faithfulness   & 100.0 (27) & 85.7 (28)  & 60.0 & [49, 70] & 75 / 27 \\
non-redundancy & 100.0 (24) & 62.1 (29)  & 48.6 & [37, 60] & 72 / 24 \\
\bottomrule
\end{tabular}

\label{tab:localization}
\label{tab:percat-raters}
\end{table*}

\paragraph{Metric Correlations.}
Table~\ref{tab:metric-corr} reports Spearman rank correlations between the
metrics at two units: the 23 trained checkpoints of the ranking-stability analysis and
the 400 pairs of the BLV study. At the system level (Table~\ref{tab:metric-corr}a),
chrF and BERTScore order the systems alike
($\rho=0.89$), CLIPScore runs against both ($-0.49$ and $-0.70$), the two
compliance modes agree with each other ($0.86$) and are uncorrelated
with the reference-based metrics ($0.06$ and $-0.16$ against chrF), and
CLIPScore is negatively correlated with mechanical compliance ($-0.62$).
Adding the slide-title readout changes no sign. 

The GPT-5.5 LLM-AD-Eval and the rubric cannot join
this block, being scored on the 150-moment draw for a handful of systems;
the two-judge LLM-AD-Eval column of Table~\ref{tab:systems} can (Table~\ref{tab:metric-corr}b), and its
two halves order the trained checkpoints alike ($+0.90$) while correlating with
neither compliance mode beyond an interval covering zero ($-0.40$ and
$-0.36$ against the mechanical mode, $-0.13$ and $-0.04$ against the local
mode) nor significantly with chrF ($+0.41$ and $+0.32$, intervals covering zero) or BERTScore ($+0.10$ and $+0.07$). At the pair
level (Table~\ref{tab:metric-corr}c), on the side differences over the 400 pairs of the BLV study, the
reference- and judge-based metrics correlate moderately (chrF with
LLM-AD-Eval $0.48$, LLM-AD-Eval with the rubric $0.44$, chrF with the rubric
$0.42$) and prefer the same side on 67--86\% of the pairs both decide,
whereas the four mechanical components are close to independent of all
three ($\rho\le0.12$; same side on 54--65\%). The compliance modes thus
measure something other metrics do not, which makes them a
complementary axis for ranking systems rather than a proxy for the
reference-based metrics. 

\begin{table}[!htb]
\centering
\caption{\textbf{Spearman rank correlations between the metrics}; mech, local, and rubric are the three \textsc{AD4Edu-Eval} modes. (a)~The 23 trained checkpoints, slide-title readout excluded. (b)~The two LLM-AD-Eval judges of Table~\ref{tab:systems} on the same checkpoints: $+0.90^{*}$ with each other, $+0.10$ and $+0.07$ with BERTScore. (c)~Per-pair side differences (A minus B) on the 400 BLV-study pairs, $n$ from 356 to 391 per cell; mech is the mean of the four mechanical components (style, terminology, length, deixis); BERTScore and CLIPScore were not scored on the pairs. $^{*}$ marks a bootstrap 95\% interval excluding zero.}
\footnotesize
\setlength{\tabcolsep}{3pt}
\begin{tabular}{@{}lrrrr@{}}
\toprule
\multicolumn{5}{@{}l}{\emph{(a) System level, 23 trained checkpoints}} \\
 & BERTScore & CLIPScore & mech & local \\
\midrule
chrF       & $+0.89^{*}$ & $-0.49^{*}$ & $+0.06$     & $-0.16$ \\
BERTScore  &             & $-0.70^{*}$ & $+0.29$     & $+0.02$ \\
CLIPScore  &             &             & $-0.62^{*}$ & $-0.35$ \\
mech       &             &             &             & $+0.86^{*}$ \\
\midrule
\multicolumn{5}{@{}l}{\emph{(b) System level, LLM-AD-Eval judges}} \\
 & chrF & CLIPScore & mech & local \\
\midrule
GPT-3.5 judge & $+0.41$ & $+0.32$ & $-0.40$ & $-0.13$ \\
Llama-2 judge & $+0.32$ & $+0.28$ & $-0.36$ & $-0.04$ \\
\midrule
\multicolumn{5}{@{}l}{\emph{(c) Pair level, 400 study pairs}} \\
 & LLM-AD & rubric & mech & \\
\midrule
chrF    & $+0.48^{*}$ & $+0.42^{*}$ & $+0.11$     & \\
LLM-AD-Eval  &             & $+0.44^{*}$ & $+0.05$     & \\
rubric  &             &             & $+0.12^{*}$ & \\
\bottomrule
\end{tabular}

\label{tab:metric-corr}
\end{table}

\section{The BLV Annotation Tool}
\label{app:tool}
\begin{figure*}[p]
\centering
\pdftooltip{\includegraphics[width=\textwidth]{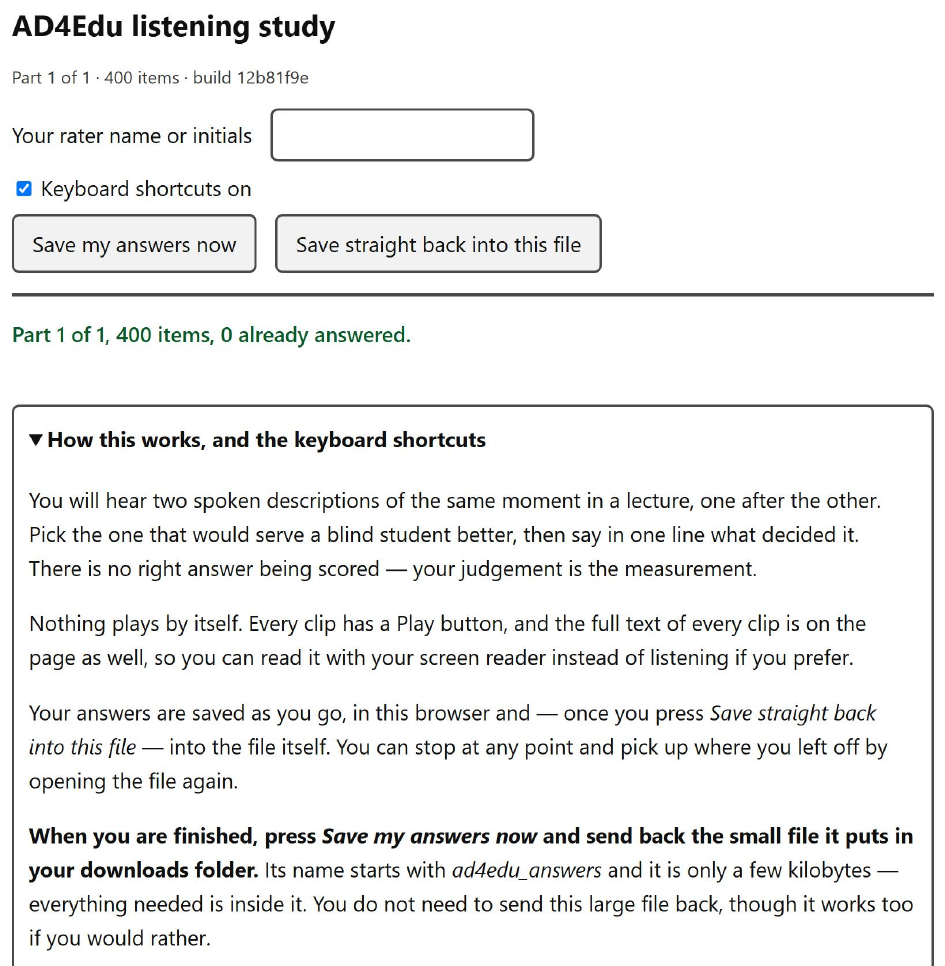}}{Screenshot of the top of the rating page. A heading reads AD4Edu listening study, with the line Part 1 of 1, 400 items, build 12b81f9e. Below are a text field labeled Your rater name or initials, a ticked checkbox labeled Keyboard shortcuts on, and two buttons, Save my answers now and Save straight back into this file. A green line reads Part 1 of 1, 400 items, 0 already answered. An expanded panel titled How this works, and the keyboard shortcuts explains in four paragraphs that the rater hears two spoken descriptions of the same lecture moment and picks the one that would serve a blind student better, then says in one line what decided it; that nothing plays by itself and the full text of every clip is on the page for screen-reader users; that answers are saved as the rater goes, in the browser and, on request, into the file itself; and that, when finished, the rater presses Save my answers now and sends back the small answers file it downloads.}
\caption{\textbf{Page header and instructions of the rating tool.} The header gives the part, the number of items, and the build identifier, and holds the rater-name field, the switch for single-key shortcuts, and the two save buttons; the progress line counts answered items. The expanded help panel states the task in the raters' own terms, that nothing plays by itself, that every clip's text is on the page, and how answers are saved and returned. The panel continues in Figure~\ref{fig:tool-shortcuts}.}
\label{fig:tool-help}
\end{figure*}

\begin{figure*}[p]
\centering
\pdftooltip{\includegraphics[width=\textwidth]{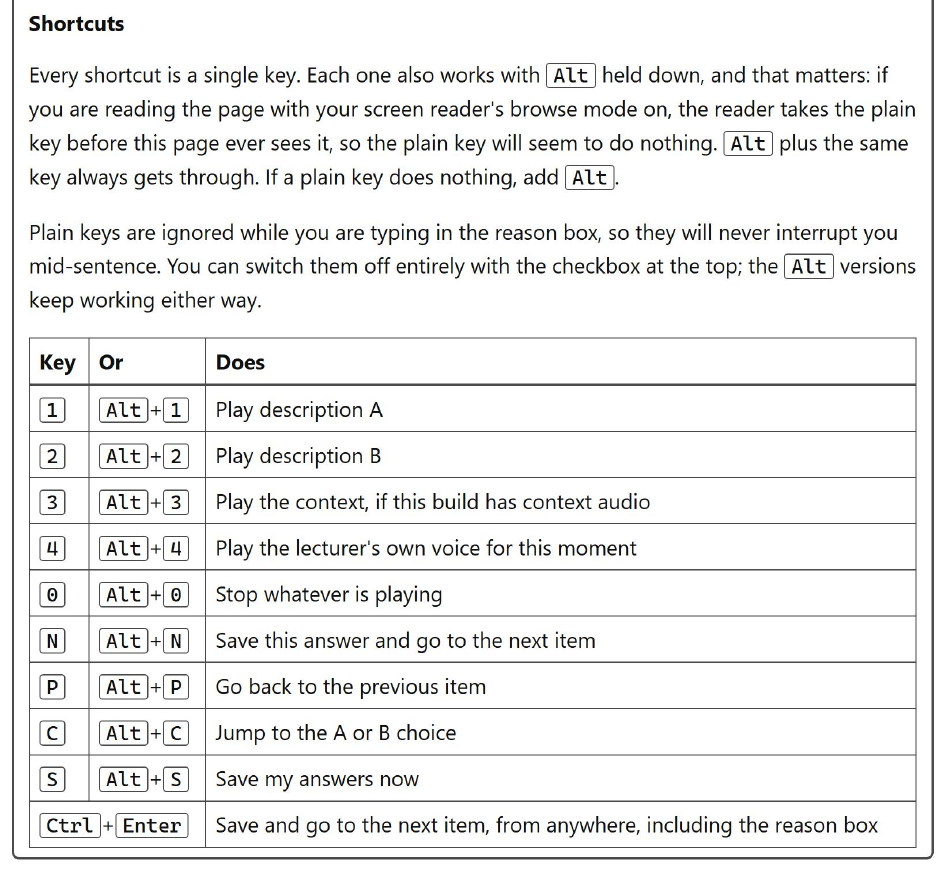}}{Screenshot of the second half of the help panel, under the heading Shortcuts. Two paragraphs explain that every shortcut is a single key that also works with Alt held down, because a screen reader in browse mode takes the plain key before the page sees it, and that plain keys are ignored while typing in the reason box and can be switched off with the checkbox at the top, while the Alt versions keep working. A three-column table lists key, alternative, and action: 1 or Alt+1 plays description A; 2 or Alt+2 plays description B; 3 or Alt+3 plays the context, if the build has context audio; 4 or Alt+4 plays the lecturer's own voice for this moment; 0 or Alt+0 stops whatever is playing; N or Alt+N saves the answer and goes to the next item; P or Alt+P goes back to the previous item; C or Alt+C jumps to the A or B choice; S or Alt+S saves the answers now; and Ctrl+Enter saves and goes to the next item from anywhere, including the reason box.}
\caption{\textbf{Keyboard shortcuts, the second half of the help panel.} Every shortcut is a single key with an Alt variant, because a screen reader in browse mode consumes plain keys before the page receives them. Plain keys are ignored while the rater types a reason and can be switched off, and Ctrl+Enter saves and advances from anywhere on the page.}
\label{fig:tool-shortcuts}
\end{figure*}

\begin{figure*}[p]
\centering
\pdftooltip{\includegraphics[width=\textwidth]{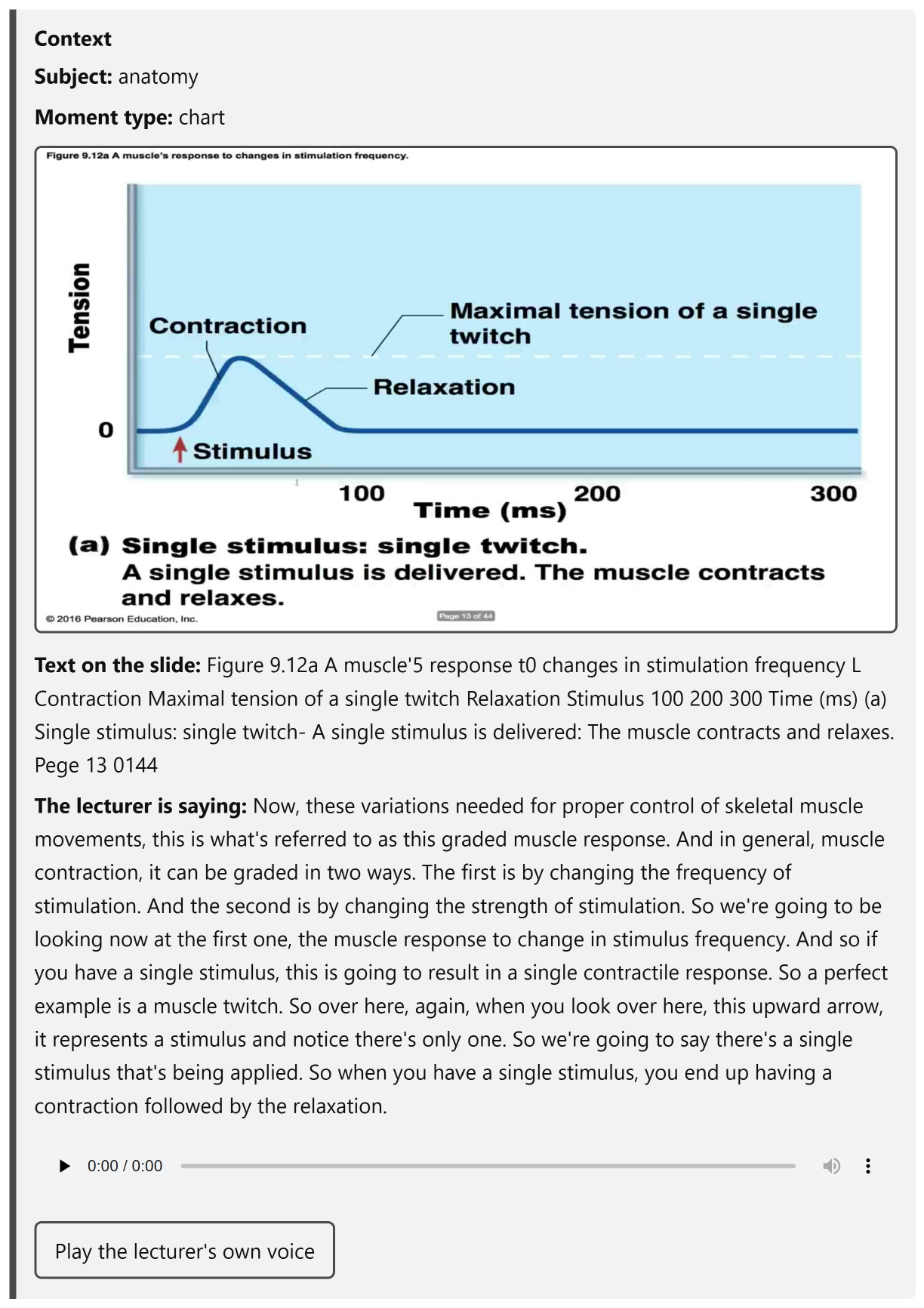}}{Screenshot of the context block of item 1 of 400. Labels read Context, Subject: anatomy, and Moment type: chart. Below is the slide still, a textbook line chart titled A muscle's response to changes in stimulation frequency: tension on the vertical axis rises from zero to a single peak labeled Contraction and falls back, labeled Relaxation, over time in milliseconds from 0 to 300, with an arrow marking the Stimulus and a dashed line marking the maximal tension of a single twitch; the caption on the slide reads Single stimulus: single twitch. A paragraph headed Text on the slide repeats the slide's text as extracted automatically, with recognition errors. A paragraph headed The lecturer is saying gives the transcript around the moment, in which the lecturer explains graded muscle responses and the single twitch. Below are an audio player and a button labeled Play the lecturer's own voice.}
\caption{\textbf{Lecture context of an item.} Item 1 of 400 (anatomy, moment type chart) shows the slide still, the slide text as our OCR pipeline extracted it, recognition errors included, the lecturer's transcript around the moment, an audio player, and a button that plays the lecturer's own voice for the moment. The slide is reproduced from the lecture recording, and its copyright remains with its publisher.}
\label{fig:tool-item}
\end{figure*}

\begin{figure*}[p]
\centering
\pdftooltip{\includegraphics[width=\textwidth]{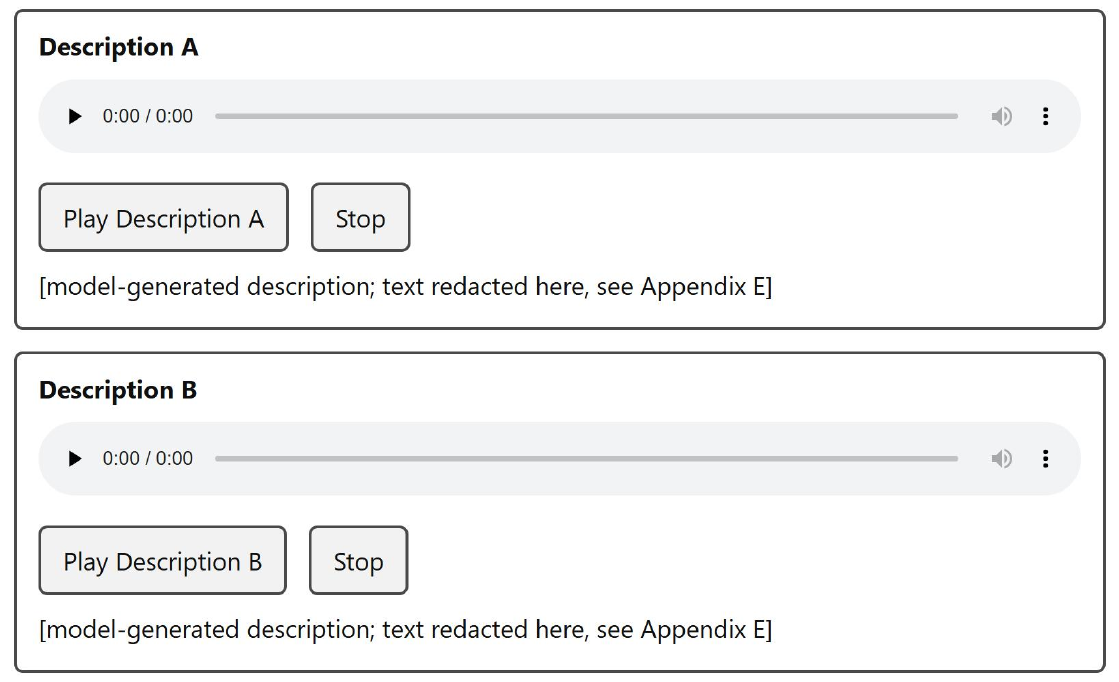}}{Screenshot of two boxes, Description A and Description B. Each holds an audio player, a Play button and a Stop button, and one line of text reading: model-generated description; text redacted here, see Appendix E.}\\[4pt]
\pdftooltip{\includegraphics[width=\textwidth]{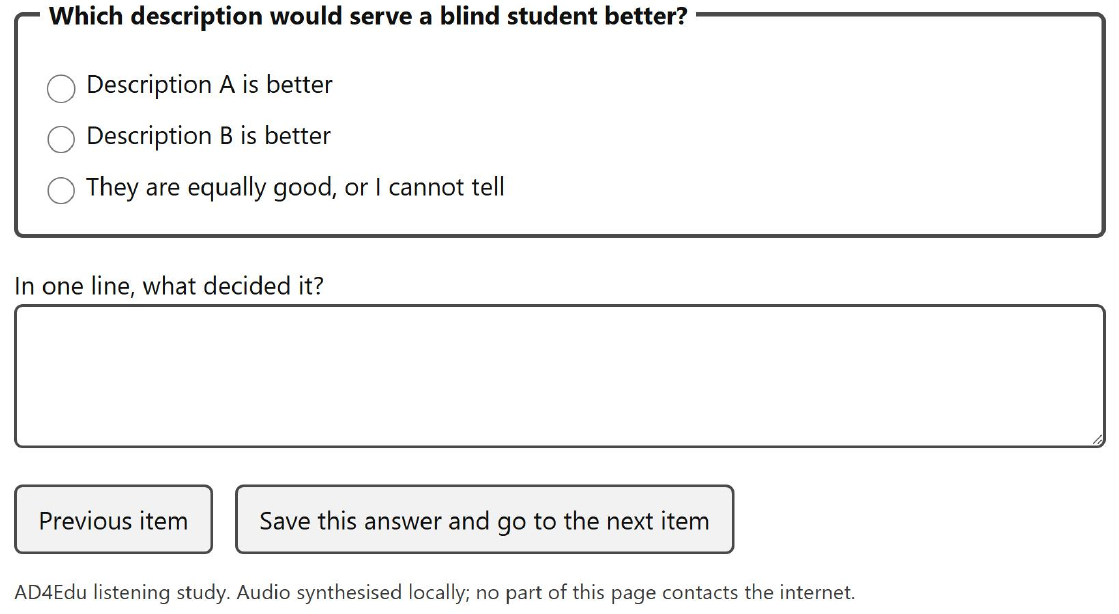}}{Screenshot of the answer area. A group titled Which description would serve a blind student better? holds three radio buttons: Description A is better; Description B is better; They are equally good, or I cannot tell. Below is a text box labeled In one line, what decided it?, then two buttons, Previous item and Save this answer and go to the next item. A footer reads: AD4Edu listening study. Audio synthesized locally; no part of this page contacts the internet.}
\caption{\textbf{Descriptions and the answer.} Top: each description has an audio player, Play and Stop buttons, and its full text below, which this rendering replaces by a placeholder (Appendix~\ref{app:examples} shows paired descriptions). Bottom: the rater chooses between A, B, and ``equally good, or I cannot tell'', writes one line on what decided it, and saves; the page refuses to advance without a choice and a reason. The footer states that the audio was synthesized locally and that the page contacts no server.}
\label{fig:tool-answer}
\end{figure*}

Raters worked in a single HTML file that runs offline in the browser, with all audio embedded and synthesized in advance; the page contacts no server. A consent screen precedes the first item and is not shown here, because it names the investigators. Each of the 400 items presents the lecture context (Figure~\ref{fig:tool-item}) and two descriptions under the neutral labels A and B (Figure~\ref{fig:tool-answer}). The page does not advance until the rater has entered a name or initials, chosen A, B, or ``equally good'', and written a one-line reason; answers are saved in the browser as the rater works and can also be written back into the file itself. Nothing plays by itself, the full text of every clip is on the page for screen-reader users, and every action has a single-key shortcut with an Alt variant, because a screen reader in browse mode intercepts plain keys (Figures~\ref{fig:tool-help} and~\ref{fig:tool-shortcuts}). The page follows WCAG~2.1 AA with axe-core rules. The screenshots are rendered from a demonstration build of the same template the raters received, before any answer was entered: the rater-name field is empty, and in Figure~\ref{fig:tool-answer} the two description texts are replaced by a placeholder.

\begin{figure*}[p]
\centering
\pdftooltip{\includegraphics[width=\textwidth]{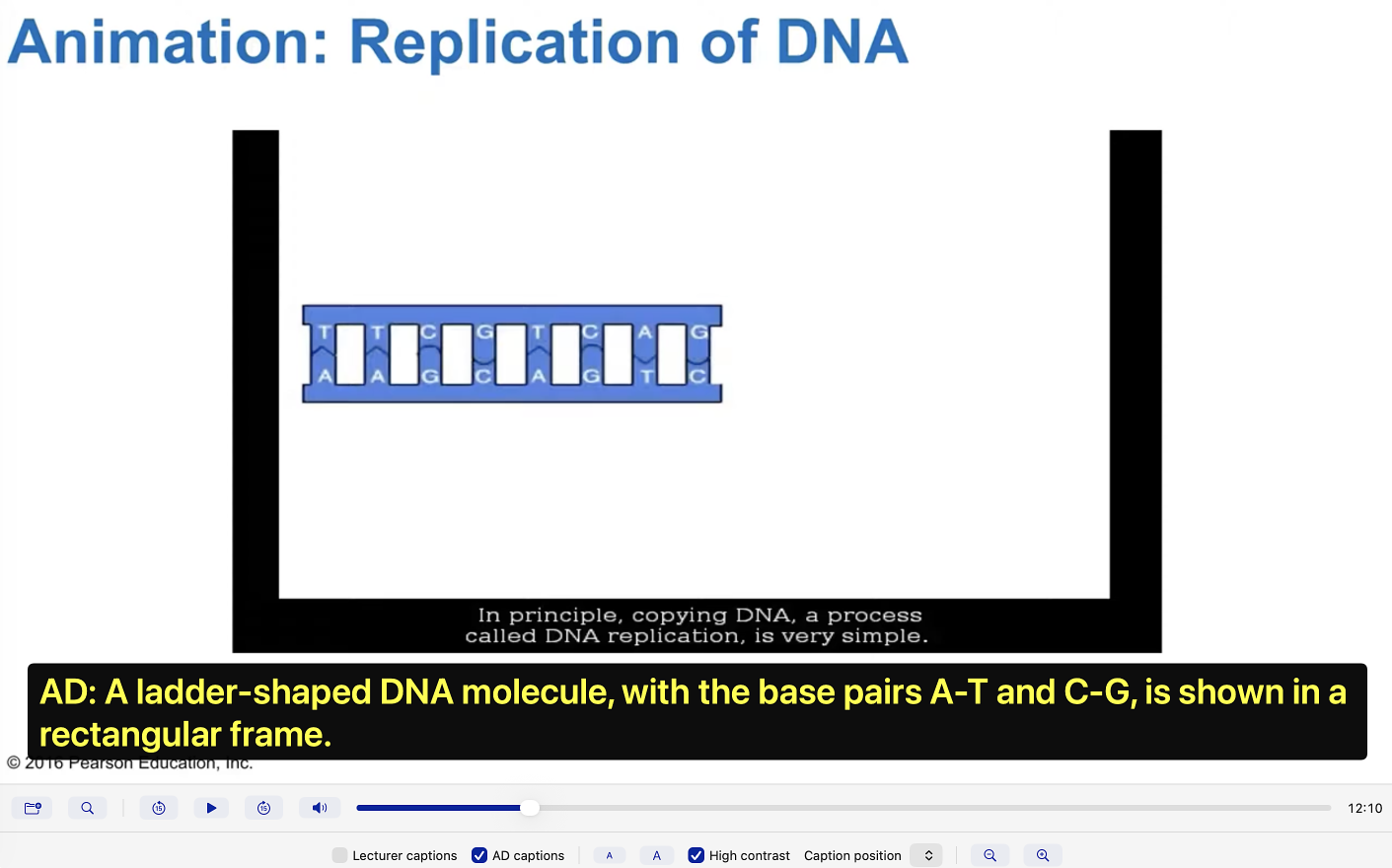}}{Screenshot of the Lecture AD Player, twelve minutes into a lecture. The slide is titled Animation: Replication of DNA and holds a video frame in which DNA is drawn as a blue ladder whose rungs carry the paired bases T-A, C-G, T-A and C-G; the subtitle burnt into the video reads: In principle, copying DNA, a process called DNA replication, is very simple. A black bar across the foot of the slide carries the audio description in yellow: A ladder-shaped DNA molecule, with the base pairs A-T and C-G, is shown in a rectangular frame. Below are the transport controls, the elapsed time 12:10, and a settings row in which AD captions and High contrast are ticked and Lecturer captions is not.}
\caption{\textbf{A description of an animation.} Twelve minutes into the lecture an animation draws DNA as a ladder and spells out its base pairs. The player pauses on the frame it is describing and shows the description as a high-contrast caption across the foot of the slide, so the same text reaches a listener as synthesized speech and a sighted reader as text. The slide is reproduced from the lecture recording, and its copyright remains with its publisher.}
\label{fig:player-dna-replication}
\end{figure*}

\begin{figure*}[p]
\centering
\pdftooltip{\includegraphics[width=\textwidth]{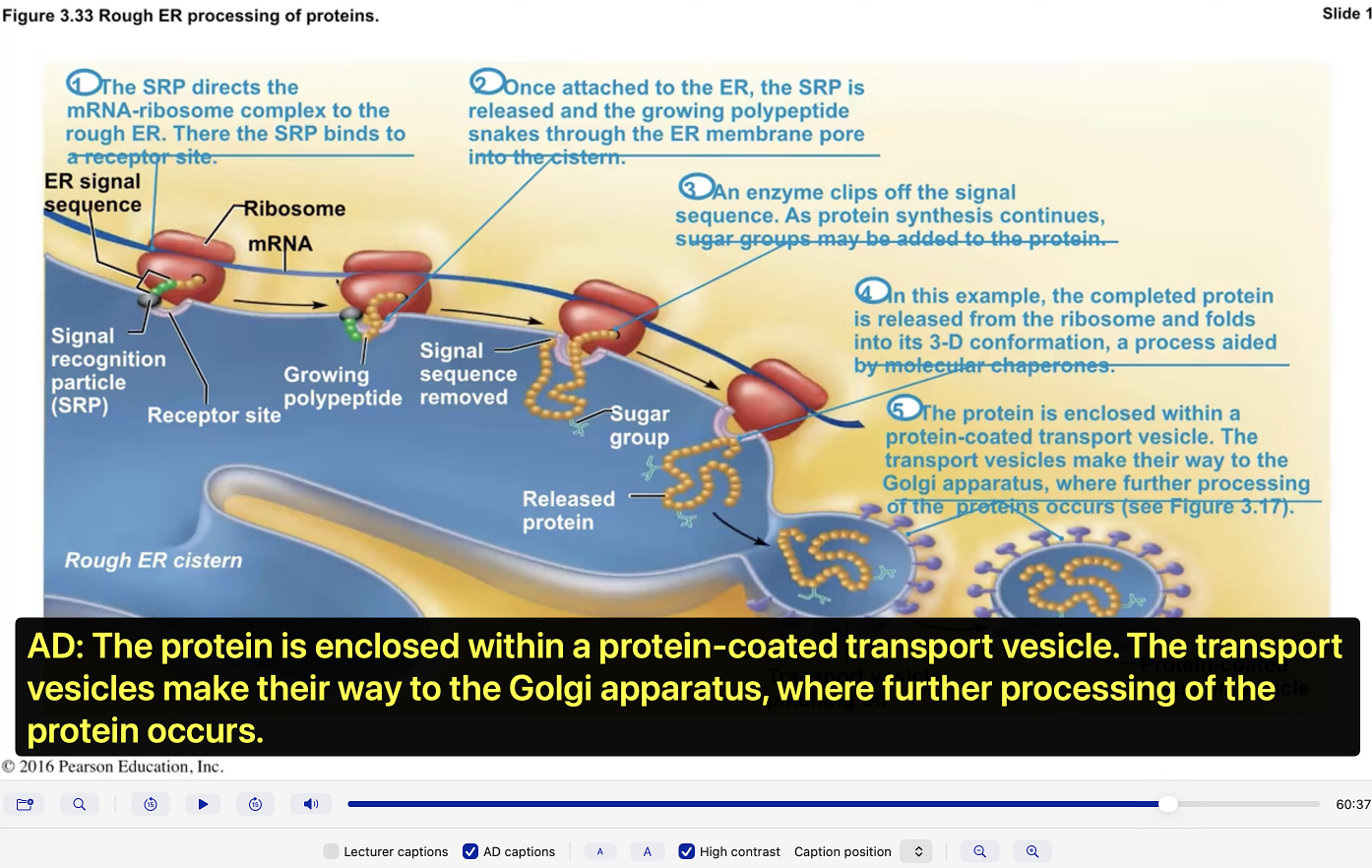}}{Screenshot of the player, an hour into a lecture. The slide, Figure 3.33 Rough ER processing of proteins, follows a ribosome along the rough endoplasmic reticulum in five numbered steps, from the signal recognition particle docking at a receptor site to the finished protein leaving inside a transport vesicle bound for the Golgi apparatus. A black bar across the foot of the slide carries the audio description in yellow: The protein is enclosed within a protein-coated transport vesicle. The transport vesicles make their way to the Golgi apparatus, where further processing of the protein occurs. Below are the transport controls, the elapsed time 60:37, and a settings row in which AD captions and High contrast are ticked and Lecturer captions is not.}
\caption{\textbf{A description an hour into the lecture.} The slide follows a protein through the rough endoplasmic reticulum in five numbered steps. The description picks up the last of them, the finished protein leaving inside a transport vesicle bound for the Golgi apparatus. The slide is reproduced from the lecture recording, and its copyright remains with its publisher.}
\label{fig:player-rough-er}
\end{figure*}

\begin{figure*}[p]
\centering
\pdftooltip{\includegraphics[width=\textwidth]{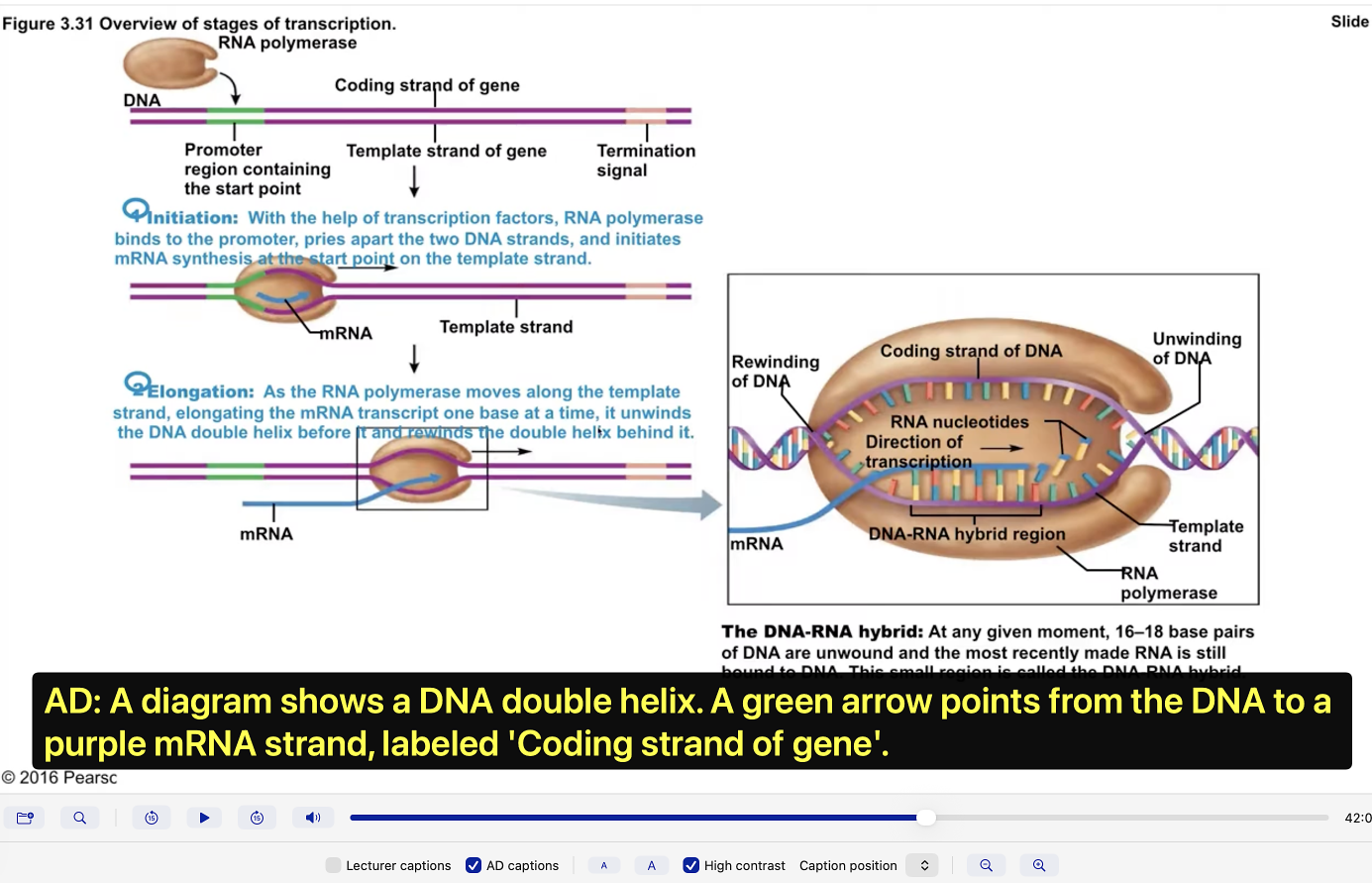}}{Screenshot of the player, forty-two minutes into a lecture. The slide, Figure 3.31 Overview of stages of transcription, diagrams RNA polymerase transcribing a gene: the promoter region, the coding and template strands and the termination signal are labeled, initiation and elongation are drawn in turn, and an inset at the right enlarges the DNA-RNA hybrid. A black bar across the foot of the slide carries the audio description in yellow: A diagram shows a DNA double helix. A green arrow points from the DNA to a purple mRNA strand, labeled `Coding strand of gene'. Below are the transport controls and a settings row in which AD captions and High contrast are ticked and Lecturer captions is not.}

\caption{\textbf{A description of a RNA diagram.} Forty-two minutes in, the slide sets out every stage of transcription at once, from the promoter region to the termination signal, with an inset enlarging the DNA-RNA hybrid. The description reduces it to the double helix and the strand the arrow points to, which is as much as the pause allows. The slide is reproduced from the lecture recording, and its copyright remains with its publisher.}
\label{fig:player-transcription}
\end{figure*}

\begin{figure*}[p]
\centering
\pdftooltip{\includegraphics[width=\textwidth]{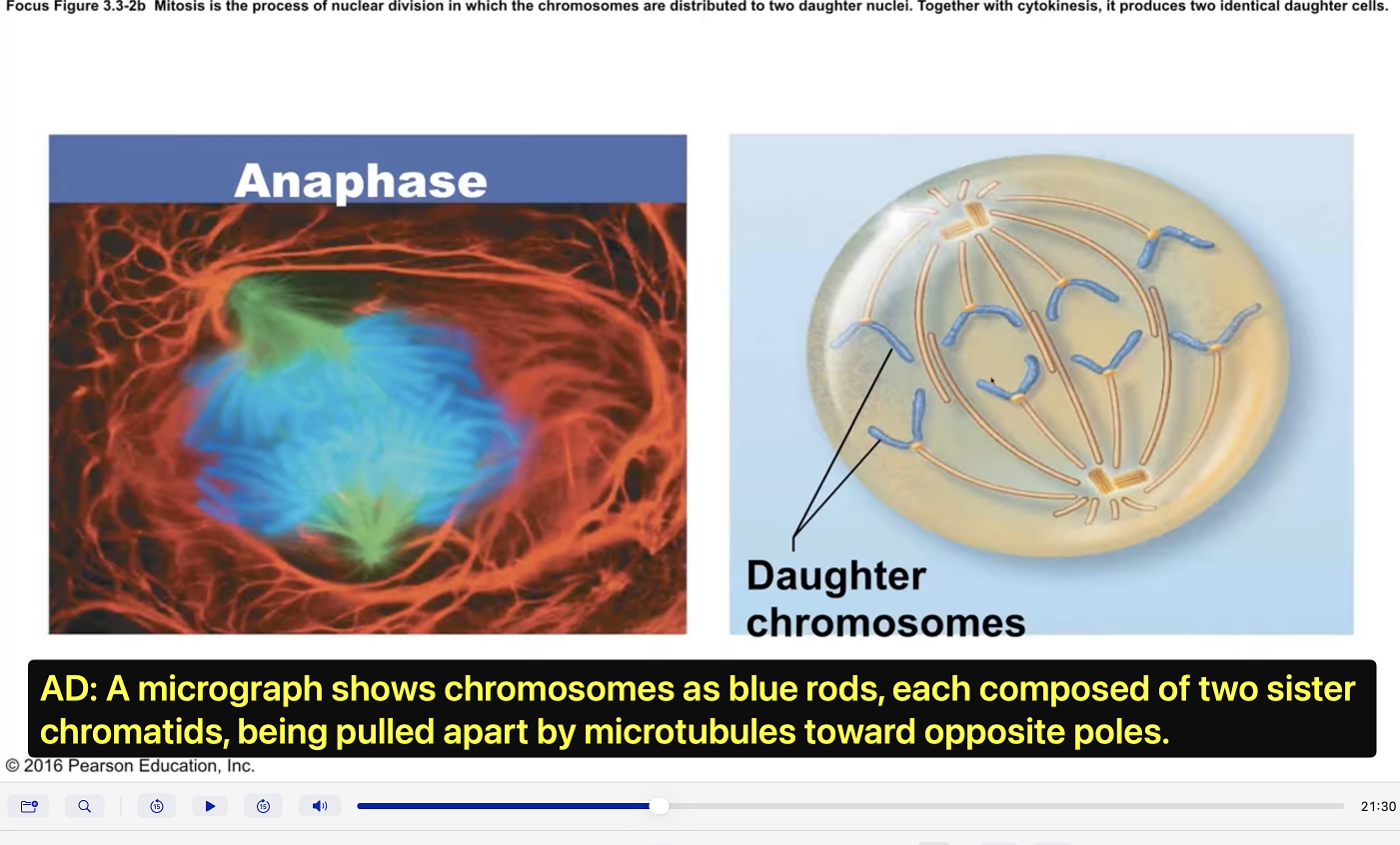}}{Screenshot of the player, twenty-one minutes into a lecture. The slide, headed Focus Figure 3.3-2b on mitosis, sets a fluorescence micrograph labeled Anaphase beside a drawing of a dividing cell in which spindle fibres draw two sets of daughter chromosomes towards opposite poles. A black bar across the foot of the slide carries the audio description in yellow: A micrograph shows chromosomes as blue rods, each composed of two sister chromatids, being pulled apart by microtubules toward opposite poles. Below are the transport controls and the elapsed time 21:30.}
\caption{\textbf{A description of a micrograph.} Twenty-one minutes in, a fluorescence micrograph of anaphase sits beside a drawing of the same stage. The description names what the micrograph shows and the direction the sister chromatids travel, rather than repeating the labels already printed on the slide. The slide is reproduced from the lecture recording, and its copyright remains with its publisher.}
\label{fig:player-anaphase}
\end{figure*}

Figures~\ref{fig:player-dna-replication}--\ref{fig:player-anaphase} show an MP4 player that presents ADs at their moments as synthesized speech and a high-contrast caption; the descriptions are shown as generated, not selected for compliance.

\newpage

\end{document}